\PassOptionsToPackage{natbib=true}{biblatex}
\RequirePackage{fix-cm}
\documentclass{druv_article}

\DeclareFontFamily{U}{bbold}{}
\DeclareFontShape{U}{bbold}{m}{n}{<-5.5> bbold5 <5.5-6.5> bbold6
    <6.5-7.5> bbold7
    <7.5-8.5> bbold8 <8.5-9.5> bbold9 <9.5-11> bbold10 <11-15> bbold12
<15-> bbold17}{}

\usepgfplotslibrary{groupplots}
\AtBeginDocument{\providecommand{\ours}{\textsc{Nestor}}}
\usetikzlibrary{calc}
\pgfdeclareplotmark{fstar}{%
  \pgfpathmoveto{\pgfqpointpolar{90}{\pgfplotmarksize}}%
  \pgfpathlineto{\pgfqpointpolar{126}{0.42\pgfplotmarksize}}%
  \pgfpathlineto{\pgfqpointpolar{162}{\pgfplotmarksize}}%
  \pgfpathlineto{\pgfqpointpolar{198}{0.42\pgfplotmarksize}}%
  \pgfpathlineto{\pgfqpointpolar{234}{\pgfplotmarksize}}%
  \pgfpathlineto{\pgfqpointpolar{270}{0.42\pgfplotmarksize}}%
  \pgfpathlineto{\pgfqpointpolar{306}{\pgfplotmarksize}}%
  \pgfpathlineto{\pgfqpointpolar{342}{0.42\pgfplotmarksize}}%
  \pgfpathlineto{\pgfqpointpolar{18}{\pgfplotmarksize}}%
  \pgfpathlineto{\pgfqpointpolar{54}{0.42\pgfplotmarksize}}%
  \pgfpathclose
  \pgfusepathqfillstroke
}
\definecolor{expCortex}{HTML}{884EA0}   %
\colorlet{expCortexMid}{expCortex!65!white}    %
\colorlet{expCortexLight}{expCortex!42!white}  %
\definecolor{expDense}{HTML}{808080}    %
\definecolor{expMoe}{HTML}{404040}      %
\definecolor{expSwaCtl}{HTML}{638ECA}   %
\definecolor{expSwaUni}{HTML}{C68A2C}   %
\definecolor{expRepeat}{HTML}{B85450}   %
\definecolor{expKimi}{HTML}{5F8940}     %
\definecolor{expNores}{HTML}{FF7E79}    %
\definecolor{expNoresEdge}{HTML}{C4534E} %
\definecolor{expBankShared}{HTML}{638ECA}  %
\definecolor{expBankPrivate}{HTML}{B85450} %
\definecolor{expSchedUniform}{HTML}{5F8940}  %
\definecolor{expSchedStep}{HTML}{638ECA}     %
\definecolor{expSchedSym}{HTML}{FF7E79}      %
\definecolor{expSchedSymEdge}{HTML}{C4534E}  %
\definecolor{expInk}{HTML}{1A1A1A}
\definecolor{expMuted}{HTML}{6F6F6F}
\definecolor{expGrid}{HTML}{E6E6E6}
\definecolor{houseRed}{HTML}{B85450}
\definecolor{houseBlue}{HTML}{6C8EBF}
\definecolor{houseBrown}{HTML}{80461B}
\definecolor{housePurple}{HTML}{884EA0}
\definecolor{houseGreen}{HTML}{5F8940}
\definecolor{houseOrange}{HTML}{FF7E79}
\newcommand{\expsnow}[1]{\edef\expstmp{#1}\expstmp}
\pgfplotsset{
  discard if not/.style 2 args={
    x filter/.append code={
      \edef\tempa{\thisrow{#1}}\edef\tempb{#2}%
      \ifx\tempa\tempb\else\def\pgfmathresult{inf}\fi
    }
  },
  discard if/.style 2 args={
    x filter/.append code={
      \edef\tempa{\thisrow{#1}}\edef\tempb{#2}%
      \ifx\tempa\tempb\def\pgfmathresult{inf}\fi
    }
  },
  exps cortex/.style={mark=fstar, mark size=4pt, mark options={fill=expCortex, draw=white, line width=0.4pt}},
  exps moe/.style={mark=diamond*, mark size=3pt, mark options={fill=white, draw=expMoe, line width=0.7pt}},
  exps dense/.style={mark=square*, mark size=2.5pt, mark options={fill=white, draw=expDense, line width=0.7pt}},
  exps cortex open/.style={mark=fstar, mark size=4pt, mark options={fill=white, draw=expCortex, line width=0.6pt}},
  exps moe open/.style={exps moe},   %
  exps moe filled/.style={mark=diamond*, mark size=3pt, mark options={fill=expMoe, draw=white, line width=0.4pt}},
  exps/.style={
    font=\sffamily\footnotesize,
    scaled ticks=false,
    tick label style={/pgf/number format/fixed, /pgf/number format/assume math mode=true, font=\sffamily\scriptsize, expMuted},
    axis lines*=left,
    axis line style={expMuted, line width=0.6pt},
    tick style={expMuted, line width=0.6pt},
    label style={expInk, font=\sffamily\footnotesize},
    ymajorgrids, grid style={expGrid, line width=0.4pt},
    unbounded coords=jump,
    legend style={draw=none, fill=none, font=\sffamily\scriptsize, /tikz/every even column/.append style={column sep=0.4em}},
    legend cell align=left,
    every axis title/.style={at={(0.5,1)}, above, font=\sffamily\footnotesize, color=expInk, text depth=0pt},  %
  },
  exps mark/.style={only marks, mark options={draw=white, line width=0.4pt}},
}
\tikzset{exps annot/.style={font=\sffamily\fontsize{6}{7}\selectfont}}

\usepgfplotslibrary{colormaps,fillbetween,statistics}
\usetikzlibrary{shapes.geometric}
\definecolor{rtMid}{HTML}{F3F3F1}       %
\definecolor{rtStruct}{HTML}{DADADA}    %
\colorlet{rtOver}{expCortex}            %
\colorlet{rtUnder}{expKimi}             %
\pgfplotsset{
  colormap={rt div}{color(0)=(rtUnder!72!black) color(0.2)=(rtUnder) color(0.35)=(rtUnder!45!rtMid)
                    color(0.5)=(rtMid)
                    color(0.65)=(rtOver!45!rtMid) color(0.8)=(rtOver) color(1)=(rtOver!72!black)},
  colormap={rt seq cortex}{color(0)=(rtMid) color(0.5)=(expCortex!55!rtMid) color(1)=(expCortex!85!black)},
  colormap={rt seq moe}{color(0)=(rtMid) color(0.5)=(expMoe!45!rtMid) color(1)=(expMoe!95!black)},
  colormap={rt seq cortex soft}{color(0)=(rtMid) color(1)=(expCortex!62!rtMid)},
  colormap={rt seq moe soft}{color(0)=(rtMid) color(1)=(expMoe!50!rtMid)},
  colormap={rt div soft}{color(0)=(rtUnder!70!rtMid) color(0.5)=(rtMid) color(1)=(rtOver!62!rtMid)},
  colormap={rt div soft rev}{color(0)=(rtOver!62!rtMid) color(0.5)=(rtMid) color(1)=(rtUnder!70!rtMid)},
  rt heat/.style={exps, enlargelimits=false, axis on top, axis lines=box, ymajorgrids=false,
    axis line style={expMuted, line width=0.4pt}, tick style={expMuted, line width=0.4pt},
    tick align=outside, y dir=reverse, tick label style={font=\sffamily\fontsize{6}{7}\selectfont, expMuted},
    label style={expInk, font=\sffamily\scriptsize},
    every axis title/.style={at={(0.5,1)}, above, font=\sffamily\scriptsize, color=expInk, text depth=0pt, yshift=-2pt}},
  rt colorbar/.style={colorbar style={exps, axis lines=box, ymajorgrids=false, width=0.12cm,
    axis line style={expMuted, line width=0.4pt}, tick style={expMuted, line width=0.4pt}, tick align=outside,
    tick label style={font=\sffamily\fontsize{6}{7}\selectfont, expMuted},
    label style={expInk, font=\sffamily\scriptsize}}},
  rt yjs/.style={ymin=0, ymax=0.32, ytick={0,0.1,0.2,0.3}}, rt yov/.style={ymin=0.2, ymax=1.02, ytick={0.2,0.4,0.6,0.8,1}},
  rt yfa/.style={ymin=0, ymax=0.5, ytick={0,0.2,0.4}}, rt yfb/.style={ymin=0, ymax=0.14, ytick={0,0.05,0.1}},
  rt yfc/.style={ymin=0, ymax=1.45, ytick={0,0.5,1}}, rt yfd/.style={ymin=0, ymax=0.42, ytick={0,0.2,0.4}},
  rt thin/.style={expMuted, line width=0.35pt, mark size=1.1pt, mark repeat=5, mark phase=3,
    mark options={fill=white, draw=expMuted, line width=0.35pt}},
  rt dclm/.style={mark=pentagon*}, rt pg19/.style={mark=*}, rt algstack/.style={mark=square*},
  rt owm/.style={mark=triangle*}, rt fw2de/.style={mark=triangle*, every mark/.append style={rotate=180}},
}

\usetikzlibrary{arrows.meta,calc,decorations.pathreplacing,fit,matrix,
positioning,shapes.geometric,shapes.misc,backgrounds,scopes}

\appto\appendix{%
}

\crefname{appendix}{Appendix}{Appendices}
\Crefname{appendix}{Appendix}{Appendices}
\appto\appendix{\crefalias{section}{appendix}\crefalias{subsection}{appendix}%
\crefalias{subsubsection}{appendix}}

\newcommand{\ours}{\textsc{Nestor}}
\newcommand{\TopK}{\operatorname{TopK}}
\newcommand{\vAttn}{\operatornamebf{Attn}}
\newcommand{\vAttnRes}{\operatornamebf{AttnRes}}
\newcommand{\vMLP}{\operatornamebf{MLP}}

\newcommand{\vnabla}{\boldsymbol{\nabla}}

\newcommand{\muP}{\ensuremath{\mu\mathrm{P}}}
\providecommand{\rms}{}
\renewcommand{\rms}{\mathrm{rms}}

\AtBeginBibliography{\sloppy}

\pretitle{
\begin{center}\LARGE}
    \posttitle{\par
\end{center}\vskip 0.5em}
\preauthor{
    \begin{center}\large\lineskip 0.5em
    \begin{tabular}[t]{c}}
        \postauthor{
        \end{tabular}\par
\end{center}}
\predate{
\begin{center}\large}
    \postdate{\par
\end{center}}

\title{Nested Sparsity for Scalable Pretraining\\ and Continual Learning}
\author{
    Tianzhe Chu\thanks{Equal contribution; ordered alphabetically.}\\
    University of Hong Kong
    \and
    Druv Pai\thanksmark{1}\\
    UC Berkeley
    \and
    Yi Ma\\
    University of Hong Kong\\
    UC Berkeley
}
\date{}

\begin{document}

\maketitle

\begin{abstract}
    Mixture-of-experts (MoE) transformers, the standard backbone for
    modern large-scale deep learning, employ conditional computation:
    only a sparse subset of model weights is used in each forward
    pass. In MoE transformers, this sparsity arises in the MoE
    operator, which generalizes the multi-layer perceptron in the
    classical transformer. In this work, we explore an alternative
    mechanism for conditional computation that analogously
    generalizes the attention operator. The sparsity induced by this
    mechanism naturally \textit{nests} (i.e., hierarchically
    composes) with the sparsity induced by the MoE operator. We show
    that a transformer employing this \textit{nested sparsity}
    mechanism, which we call \ours{},
    performs better than a conventional MoE transformer at equal
    compute and equal sparsity in language
    model pretraining. We also show that \ours{} is well-suited for
    continual pretraining: it can be adapted to longer
    contexts more effectively than conventional MoEs,
    as well as adapt more efficiently to new domains while forgetting
    less on old domains. Furthermore, we demonstrate that
    these benefits persist as model size, dataset size, and context
    length grow. Our results demonstrate the potential of alternative
    conditional computation mechanisms for deep neural
    network architectures. Code and checkpoints are available on
    \href{https://github.com/Ma-Lab-Berkeley/Nestor}{GitHub}.
\end{abstract}

\section{Introduction}\label{sec:introduction}

\paragraph{Conditional computation in LLMs.} Large language models
(LLMs) based on variants of the transformer architecture
\citep{vaswani2017attention} improve predictably as their parameters,
training data, and training compute grow \citep{kaplan2020scaling,
brown2020language, hoffmann2022training}. A central goal of
architecture design is to reach a lower loss with the same training
compute. A particularly successful approach is
\textit{conditional computation} \citep{bengio2013estimating,
shazeer2017outrageously}, in which each input uses only a small
(input-dependent) ``active'' subset of the model weights and
correspondingly less compute for training or inference. In this
setting, the number of total parameters grows much faster than the
active parameters and compute used on each token.

\paragraph{Mixtures of experts (MoEs).} In modern LLMs, this form of
conditional computation is
implemented almost exclusively via
the MoE operator \citep{jacobs1991adaptive,
shazeer2017outrageously, lepikhin2021gshard, fedus2022switch}, which replaces
the multi-layer perceptron (MLP) in each transformer layer by many expert MLPs
and routes each token to a few of them. MoE transformers need substantially
less compute than dense transformers to reach the same loss
\citep{fedus2022switch, clark2022unified}, and this advantage is retained at
scale when the experts are configured appropriately
\citep{krajewski2024scaling, abnar2025parameters}. As a result, many
modern LLMs are MoE transformers \citep{jiang2024mixtral,
liu2024deepseekv3, qwen2025qwen3, kimi2025k2, openai2025gptoss}. In all of
these models, however, this conditional computation is
confined to the MLP weights:
every token is processed by the same attention parameters in every layer. The
attention operator accounts for a considerable fraction of the
(active) parameters
used by each token, and at long context lengths it accounts for most of the
computation. One may then ask {\em whether conditional computation
    can be applied to the attention operator as well, and how it should interact
with the MoE.}

\paragraph{Sparsity in the attention operator.} Attention operators
traditionally implement two forms of conditional computation.
The more popular by far is the so-called ``sparse attention''
methods, which reduce the cost of attention by restricting the set of
tokens to which each
token attends, via fixed local patterns \citep{child2019generating,
beltagy2020longformer}, hashing or clustering \citep{kitaev2020reformer,
roy2021efficient}, or a top-\(K\) selection for each query
\citep{yuan2025native, lu2025moba, deepseek2025v32}. Relatedly, many LLMs
interleave a few global attention layers with cheaper local or linear
attention layers \citep{brown2020language, gemma2025gemma3, kimi2025linear}.
These methods sparsify the attention \textit{map}, with the purpose of
approximating full attention at a lower cost, while {\em all tokens still share
the same attention parameters}. Other methods restrict the set of
parameters used by each token, such as mixtures of
attention heads
\citep{zhang2022mixture, csordas2024switchhead}, which route each
token to a subset
of the attention \textit{parameters}, while every token still attends to the
whole context. Other methods make both the map and the parameters sparse
\citep{piekos2025mixture}, or pair mixtures of attention heads with an MoE
\citep{shen2023moduleformer, csordas2024switchhead}. In each of them, the
attention is routed separately from the MLP and anew in every layer.

\paragraph{Continual pretraining.} One of the circumstances in which
such attention sparsity is most effective at reducing both training
and inference costs is with long context documents. Pretrained models
are routinely continually (pre-)trained in a long-context extension stage
\citep{xiong2024effective, fu2024data, grattafiori2024llama3}, and
simultaneously further on new domains,
such as code and mathematics \citep{gururangan2020dont, roziere2023code,
azerbayev2023llemma, shao2024deepseekmath}. Since neural networks
trained on a new distribution tend to perform worse on the old
one \citep{mccloskey1989catastrophic, french1999catastrophic,
kirkpatrick2017overcoming}, and LLMs are no exception
\citep{luo2025empirical}, a central challenge is to maintain the
performance on old data while capably handling the new data.

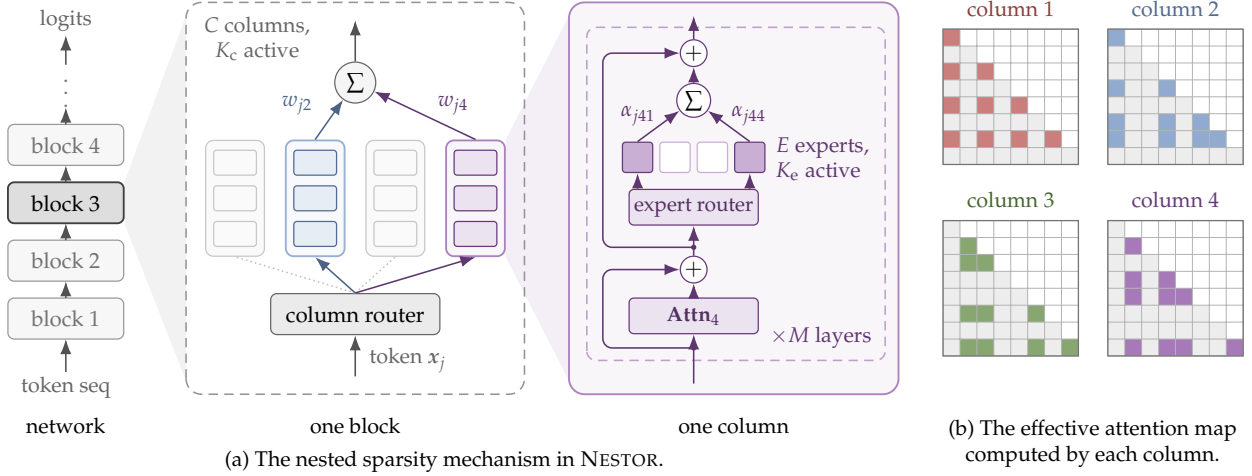
\begin{figure}[t!]
    \centering
    \resizebox{\textwidth}{!}{\begin{tikzpicture}[
  x=1cm, y=1cm, >=Latex, font=\scriptsize,
  line/.style={draw=black!68, -{Latex[length=1.7mm]}, line width=.5pt},
  io/.style={draw=black!58, fill=black!5, rounded corners=2pt,
             minimum width=.8cm, minimum height=.62cm, align=center},
  router/.style={draw=black!60, fill=black!8, rounded corners=2pt,
                 minimum width=1.15cm, minimum height=.74cm, align=center},
  merge/.style={draw=black!60, fill=black!4, circle, inner sep=0pt,
                minimum size=.46cm},
  cell/.style={rounded corners=1.5pt, minimum width=1.0cm, minimum height=.46cm,
               inner sep=0pt, align=center, font=\fontsize{6.6}{7.6}\selectfont},
  lbl/.style={font=\fontsize{6.8}{8}\selectfont, align=center, text=black!60}
]

\def\zc{housePurple}

\foreach \y/\t in {-1.35/1, -.7/2, .6/4}
  \node[draw=black!40, fill=black!3, rounded corners=2pt, minimum width=1.3cm,
        minimum height=.46cm, text=black!55] at (.65,\y) {block \t};
\node[draw=black!75, fill=black!13, rounded corners=2pt, minimum width=1.3cm,
      minimum height=.46cm, line width=.6pt] (blk) at (.65,-.05) {block 3};
\foreach \ya/\yb in {-1.95/-1.58, -1.12/-.93, -.47/-.28, .18/.37, .83/1.02,
                     1.5/1.83}
  \draw[line] (.65,\ya) -- (.65,\yb);
\node[text=black!55] at (.65,1.29) {$\vdots$};
\node[lbl] at (.65,-2.12) {token seq};
\node[lbl] at (.65,2.0) {logits};
\node[lbl, text=black] at (.65,-2.55) {network};

\draw[draw=black!42, densely dashed, rounded corners=5pt, line width=.5pt]
  (2.0,-2.2) rectangle (5.8,2.2);
\begin{scope}[on background layer]
  \fill[black!5] (blk.north east) -- (2.0,2.05) -- (2.0,-2.05)
    -- (blk.south east) -- cycle;
\end{scope}
\node[router, minimum height=.46cm, minimum width=1.9cm] (cr) at (3.9,-1.3)
  {column router};
\draw[line] (3.9,-2.0) -- (cr) node[pos=.4, right=1pt, lbl] {token $\vx_j$};
\node[merge] (mg) at (3.9,1.3) {$\textstyle\sum$};
\draw[line] (mg) -- (3.9,2.0);
\foreach \c/\x/\col/\on in {1/2.55/black/0, 2/3.45/houseBlue/1,
    3/4.35/black/0, 4/5.25/housePurple/1} {
  \ifnum\on=1
    \node[draw=\col!70, fill=\col!5, rounded corners=2.5pt, line width=.6pt,
          minimum width=.66cm, minimum height=1.3cm] (col\c) at (\x,0) {};
    \foreach \y in {-.4,0,.4}
      \node[draw=\col!75!black, fill=\col!16, rounded corners=1pt,
            minimum width=.48cm, minimum height=.28cm] at (\x,\y) {};
    \draw[line, draw=\col!70!black] (cr.north) -- (col\c.south);
    \draw[line, draw=\col!70!black] (col\c.north) -- (mg);
  \else
    \node[draw=black!22, fill=black!2, rounded corners=2.5pt,
          minimum width=.66cm, minimum height=1.3cm] (col\c) at (\x,0) {};
    \foreach \y in {-.4,0,.4}
      \node[draw=black!22, rounded corners=1pt, minimum width=.48cm,
            minimum height=.28cm] at (\x,\y) {};
    \draw[draw=black!25, densely dotted, line width=.5pt]
      (cr.north) -- (col\c.south);
  \fi
}
\node[lbl, text=houseBlue!70!black] at (3.25,1.05) {$w_{j2}$};
\node[lbl, text=housePurple!70!black] at (5.0,1.05) {$w_{j4}$};
\node[lbl, anchor=north west] at (2.08,2.17) {$C$ columns,\\$K_{\mathrm c}$ active};
\node[lbl, text=black] at (3.9,-2.55) {one block};

\draw[draw=\zc!70, fill=\zc!5, rounded corners=5pt, line width=.6pt]
  (6.3,-2.2) rectangle (10.0,2.2);
\begin{scope}[on background layer]
  \fill[\zc!9] (col4.north east) -- (6.3,2.05) -- (6.3,-2.05)
    -- (col4.south east) -- cycle;
\end{scope}
\draw[draw=\zc!55, densely dashed, rounded corners=3pt, line width=.5pt]
  (6.5,-1.82) rectangle (9.85,1.92);
\node[lbl, anchor=south east, text=\zc!70!black] at (9.82,-1.81)
  {$\times M$ layers};
\node[cell, draw=\zc!75!black, fill=\zc!16, text=\zc!60!black,
      minimum width=1.45cm, minimum height=.4cm] (attn) at (7.7,-1.32) {$\vAttn_{4}$};
\node[merge, draw=\zc!75!black, fill=white, minimum size=.3cm] (p1)
  at (7.7,-.8) {$+$};
\node[cell, draw=\zc!75!black, fill=\zc!16, text=\zc!60!black,
      minimum width=1.45cm, minimum height=.4cm] (er) at (7.7,-.1) {expert router};
\foreach \e/\on in {1/1, 2/0, 3/0, 4/1} {
  \pgfmathsetmacro{\ex}{7.7+.42*(\e-2.5)}
  \ifnum\on=1
    \node[draw=\zc!75!black, fill=\zc!45, rounded corners=1pt,
          minimum size=.34cm, inner sep=0pt] (e\e) at (\ex,.46) {};
  \else
    \node[draw=\zc!45, fill=white, rounded corners=1pt, minimum size=.34cm,
          inner sep=0pt] (e\e) at (\ex,.46) {};
  \fi
}
\node[merge, draw=\zc!75!black, fill=white, minimum size=.38cm] (ms)
  at (7.7,1.1) {$\textstyle\sum$};
\node[merge, draw=\zc!75!black, fill=white, minimum size=.3cm] (p2)
  at (7.7,1.63) {$+$};
\draw[line, draw=\zc!70!black] (7.7,-2.08) -- (attn);
\draw[line, draw=\zc!70!black] (attn) -- (p1);
\draw[line, draw=\zc!70!black] (p1) -- (er);
\foreach \e in {1,4} {
  \draw[line, draw=\zc!70!black] (e\e.south |- er.north) -- (e\e.south);
  \draw[line, draw=\zc!70!black] (e\e.north) -- (ms);
}
\draw[line, draw=\zc!70!black] (ms) -- (p2);
\draw[line, draw=\zc!70!black] (p2) -- (7.7,2.05);
\node[lbl, text=\zc!70!black] at (7.08,.93) {$\alpha_{j41}$};
\node[lbl, text=\zc!70!black] at (8.32,.93) {$\alpha_{j44}$};
\draw[line, draw=\zc!70!black, rounded corners=3pt]
  (7.7,-1.68) -- (6.68,-1.68) -- (6.68,-.8) -- (p1);
\draw[line, draw=\zc!70!black, rounded corners=3pt]
  (7.7,-.55) -- (6.68,-.55) -- (6.68,1.63) -- (p2);
\fill[\zc!70!black] (7.7,-1.68) circle (.03) (7.7,-.55) circle (.03);
\node[lbl, anchor=west, align=left, text=\zc!70!black] at (8.5,.44)
  {$E$ experts,\\$K_{\mathrm e}$ active};
\node[lbl, text=black] at (8.15,-2.55) {one column};

\node[lbl, text=black] at (4.9,-2.95)
  {(a) The nested sparsity mechanism in \ours{}.};

\def\s{0.19}
\foreach \c/\col/\ox/\oy/\set in {
    1/houseRed/10.5/1.9/{1,3,5,7}, 2/houseBlue/12.35/1.9/{1,4,6,7},
    3/houseGreen/10.5/-0.25/{2,3,6,8}, 4/housePurple/12.35/-0.25/{2,4,5,8}} {
  \begin{scope}[shift={(\ox,\oy)}]
    \foreach \j in {1,...,8} \foreach \i in {1,...,\j}
      \fill[black!7] (\i*\s-\s,-\j*\s) rectangle (\i*\s,-\j*\s+\s);
    \foreach \j in \set \foreach \i in \set {
      \ifnum\i>\j\else
        \fill[\col!75] (\i*\s-\s,-\j*\s) rectangle (\i*\s,-\j*\s+\s);
      \fi
    }
    \draw[black!30, line width=.2pt, step=\s] (0,-8*\s) grid (8*\s,0);
    \draw[black!55, line width=.4pt] (0,-8*\s) rectangle (8*\s,0);
    \node[lbl, text=\col!75!black, anchor=south] at (4*\s,0.02) {column \c};
  \end{scope}
}
\node[lbl, text=black] at (12.18,-2.75) {(b) The effective attention map \\ computed by each column.};

\end{tikzpicture}}
    \caption{\small \textbf{Nested sparsity in \ours{}.} \textit{(a)}
        The \ours{} architecture. In each block, the column
        router
        sends the token \(\vx_{j}\) to \(K_{\mathrm{c}} = 2\) of the \(C = 4\)
        columns. Each column has
        \(M\) layers, each with its own attention and its own
        \(E = 4\) experts, of which a per-layer expert
        router selects \(K_{\mathrm{e}} = 2\). The token uses only the
        colored parameters.
        \textit{(b)} The causal attention map computed by each column
        for \(S = 8\) tokens, shown in terms of the whole sequence: column
        \(c\) computes attention only among the tokens \(\cU_{c}\) routed to
    it.}
    \label{fig:architecture}
\end{figure}

\paragraph{Our contributions.} We propose \ours{}
(\Cref{fig:architecture}), a  transformer architecture
with \textit{nested sparsity}: we generalize the attention operator
in the same way that the MoE operator generalizes the MLP, and
hierarchically composes the two resultant operators' sparsity. This
yields a natural architecture that generalizes MoE transformers and
has three coupled but distinct forms of sparsity: attention weight
sparsity, attention map sparsity, and expert sparsity. We study it
empirically and theoretically in language model pretraining and
continual pretraining, showing that \ours{} architectures utilizing
all sparsity axes in tandem outperform strong MoE and dense
Transformer baselines at scales from 100M to 1B active parameters, at
the limit of our compute resources.
Namely, our contributions are:

\begin{itemize}
    \item \textbf{Compute-efficient pretraining} (\Cref{sec:pretraining}).
        We perform a scaling law study, finding that \ours{} is
        \(1.13\)--\(1.29\times\) as
        compute-efficient as the MoE transformer, and
        \(3.1\)--\(4.1\times\) as compute-efficient as a dense transformer;
        neither factor shrinks as the scale grows.
    \item \textbf{Effective continual pretraining}
        (\Cref{sec:continual-pretraining}). When adapted to various
        specialized domains, we show that \ours{}
        gains about as much as the MoE
        transformer on every new domain while forgetting less of the
        pretraining distribution, with fewer training FLOPs per
        token in the continual pretraining data. When extended
        from a context length of \(4\)k to lengths up to \(64\)k,
        \ours{} preserves
        its advantage over the MoE transformer when each model
        receives an equal number of continual pretraining tokens, and
        its advantage over the MoE transformer
        grows at longer context lengths when each model uses equal
        continual pretraining FLOPs.
    \item \textbf{Theoretical analysis} (\Cref{sec:theory}). We
        give a theoretical account of these observations. Namely, we
        give a statistical model of data with hierarchical latent
        structure, in which a stylized version of \ours{}
        outperforms a stylized version of an
        MoE transformer. We show that this analytically tractable
        model reproduces the pretraining compute efficiency, its growth with the
        context length, and the reduced forgetting that we observe
        empirically.
\end{itemize}
Overall, our results suggest that conditional computation need not be
confined to the MLP, and that adding and composing new sparsity
mechanisms is a promising direction for architecture design. We
postpone a detailed discussion of related work to \Cref{app:related-work}.

\section{\ours{}: Nested Sparse Transformer}\label{sec:architecture}

This section describes the \ours{} architecture. Here we give its
overall structure, intuitions, and relations to other architectures.
A complete mathematical description of the architecture is in
\Cref{app:architecture}.

\paragraph{Conditional computation in MoE transformers.} Each layer
of a transformer \citep{vaswani2017attention} updates the token
features \(\vx_{1}, \dots, \vx_{S} \in \R^{D}\) using an attention
operator, which mixes information across tokens, followed by an MLP, which
acts on each token separately. The MoE transformer
\citep{shazeer2017outrageously, fedus2022switch, dai2024deepseekmoe}
replaces the MLP by \(E\) \textit{expert} MLPs, of which each token uses
only \(K_{\mathrm{e}} \ll E\). A learned router \(\vr\) assigns the token
\(\vx_{j}\) a vector of scores \(\vr(\vx_{j}) \in \R^{E}\), and the MoE
operator computes (omitting normalization for simplicity, here and after):
\begin{equation}\label{eq:moe}
    \vx_{j} + \sum_{e \in \cE_{j}} \alpha_{je}\, \vMLP_{e}(\vx_{j}),
    \qquad \text{where} \quad
    \cE_{j} \doteq \TopK(\vr(\vx_{j}), K_{\mathrm{e}}) \subseteq [E],
\end{equation}
and the gating weights \((\alpha_{je})_{e \in \cE_{j}}\) are the softmax of
the routing scores for the selected experts (see
\Cref{fig:architecture}a). The number of
parameters in the MoE operator
grows with \(E\), while the computation spent on each token grows only
with \(K_{\mathrm{e}}\). This conditional computation is confined to the
MLP.

\paragraph{Column-routed attention.} We now apply the same principle to
the attention operator. We replace the attention
operator by \(C\) attention operators \(\vAttn_{1}, \dots, \vAttn_{C}\)
with separate parameters, which we call \textit{columns}. A \textit{column
router} with scores \(\vs(\vx_{j}) \in \R^{C}\) selects for each token a set
\(\cC_{j} \subseteq [C]\) of \(K_{\mathrm{c}}\) columns, with gating weights
\((w_{jc})_{c \in \cC_{j}}\), as in \eqref{eq:moe}. Namely, letting
\(\cU_{c} \doteq \{i \in [S] \colon c \in \cC_{i}\}\) be the set of
tokens that were
routed to the same column, the routed attention computes:
\begin{equation}\label{eq:col-attn}
    \va_{j} \doteq \vx_{j} + \sum_{c \in \cC_{j}} w_{jc}\,
    \vAttn_{c}(\vx_{j},\, (\vx_{i})_{i \in \cU_{c},\, i \leq j}),
    \qquad \text{where} \quad \cC_j \doteq \TopK(\vs(\vx_j),
    K_{\mathrm{c}})\subseteq [C],
\end{equation}
where \(\vAttn_{c}(\vx, \cdot)\) denotes multi-head attention with the
query computed from \(\vx\) and the keys and values computed from the given
context. Each column is a sequence-to-sequence operator applied to a
subsequence chosen by the router. Routing
therefore sparsifies both the parameters used by a token, i.e.,
\(K_{\mathrm{c}}\) of the \(C\) attention operators, and the
attention pattern (\Cref{fig:architecture}b).

\paragraph{Nested sparsity.} The two routing mechanisms compose
hierarchically, as in a hierarchical mixture of experts
\citep{jordan1994hierarchical, shazeer2017outrageously}, except that the
first level of routing also selects the attention operator and its context.
We give each column \(c\) its own \(E\) experts
\(\vMLP_{c, 1}, \dots, \vMLP_{c, E}\) and its own expert router
\(\vr_{c}\). In each active column \(c \in \cC_{j}\), this router selects
the set \(\cE_{jc} \doteq \TopK(\vr_{c}(\va_{j}), K_{\mathrm{e}}) \subseteq
[E]\) of experts, with gating weights \((\alpha_{jce})_{e \in \cE_{jc}}\):
\begin{equation}\label{eq:nested-moe}
    \vx_{j}^{+} \doteq \va_{j} + \sum_{c \in \cC_{j}} w_{jc}
    \sum_{e \in \cE_{jc}} \alpha_{jce}\, \vMLP_{c, e}(\va_{j}).
\end{equation}
An expert is indexed by the pair \((c, e)\) (see
\Cref{fig:architecture}a). One layer
stores \(C\) attention operators and \(C E\) experts, while a token uses
\(K_{\mathrm{c}}\) and \(K_{\mathrm{c}} K_{\mathrm{e}}\) of them
respectively. Each layer also has a small
number \(CE_{\mathrm{s}}\) of experts whose weights are tied across
columns (cf.\ the shared experts of \citet{dai2024deepseekmoe}).

\paragraph{\ours{} as a meta-architecture.}
We group the layers of the network into \textit{blocks} of \(M\) layers
each, and compute the column routing once per block, from the input of the
block, so that a token stays in the same columns for all \(M\) layers
within the block (see
\Cref{fig:architecture}a left). Within a block, each
column is a small MoE transformer of depth \(M\) whose attention operates
on its own subsequence; the columns of a token share the input to each
layer (\Cref{app:architecture}). Apart from the shared experts, the routers, and
the residual connections, a token uses only the parameters of its
\(K_{\mathrm{c}}\) columns.
Setting
\(C = 1\) in \eqref{eq:col-attn} and \eqref{eq:nested-moe} recovers
the regular MoE
transformer exactly, and further setting \(E = 1\) recovers the dense
transformer, so conventional
transformers are special cases of \ours{}. Actually, at
the expense of a slightly more inefficient implementation, we can use
any sequence-to-sequence deep architecture for each column. Thus, one can
view \ours{} as a meta-architecture for sparsely and
hierarchically composing parallel deep networks, such that the presented
version is a special case.

\paragraph{Relation to sparse attention.} Sparse attention methods
restrict the set of tokens to which each token attends (the attention
\textit{pattern}) to reduce the cost of long contexts. The
pattern may be fixed in advance, as in sliding window attention and its
variants \citep{child2019generating, beltagy2020longformer}; computed by
clustering the tokens \citep{roy2021efficient}; or selected for each query
by a top-\(K\) rule over blocks or tokens \citep{yuan2025native,
lu2025moba, deepseek2025v32}. In all of these methods every token is
processed by the same attention parameters, so only the pattern is
conditional, and the purpose is to approximate full
attention at a lower cost. Mixtures of attention heads
\citep{zhang2022mixture, csordas2024switchhead} route each token to a
subset of the attention \textit{parameters}, and each token still
attends to the full context. Column-routed attention
\eqref{eq:col-attn} is sparse in both senses: one routing decision
determines both the parameters that the token uses and the tokens to which
it attends. The closest mechanism is that of \citet{piekos2025mixture}, in
which each attention head selects a fixed number of tokens and attends only
among them. There, the selection is made by the heads, separately in every
layer, and is independent of the MLP; in \ours{}, the selection is made by
the tokens, once per block, and the selected module contains the attention
operators and the experts of all \(M\) layers. This makes the
sparsity nested. The two approaches can be combined, because any
sparse attention pattern (or, again, any backbone network) can be
applied within a column. In this work
each column uses full causal attention over its tokens; the
baselines in \Cref{sec:pretraining,sec:continual-pretraining} also
use full causal attention. We discuss the relation of \ours{} to
such related work further and in detail in \Cref{app:related-work}.

\paragraph{Further design choices.} We summarize the remaining
empirical design choices here, with more details in \Cref{app:experiments}.
\textit{Sparsity schedule.} The number of active columns
\(K_{\mathrm{c}}\) can differ between blocks. In all experiments we use
\(C = 8\) and increase \(K_{\mathrm{c}}\) with depth, from
\(K_{\mathrm{c}} = 3\) in the first block to \(K_{\mathrm{c}} = C\) in the
last block, which computes full attention over the whole sequence.
Hence, this architecture relates to the interleaved globalized and
local (or linear) attention layers popular in modern LLMs
\citep{brown2020language, gemma2025gemma3, kimi2025linear}; here, one
can interpret \ours{} as using progressively more globalized
attention layers. Our experimental baselines use global attention at
each layer, therefore using more attention compute than \ours{}.
\textit{Residual connections.} Equations \eqref{eq:col-attn} and
\eqref{eq:nested-moe} show plain residual connections. In \ours{},
each active column accumulates its outputs in its own residual stream. The
streams are combined, within and across
blocks, by an attention residual \citep{kimi2026attnres}: the input of each
layer is a learned softmax-weighted mixture of the token embedding, the
outputs of the earlier blocks, and the mean of the streams of the current
block. The input of the output head also includes the deviation of each
stream from the mean. We ablate this choice in \Cref{app:ablations}; note
that our baselines also use and benefit from attention residuals.
\textit{Routing.} Both routers use top-\(K\) selection by
each token, and are balanced without an auxiliary loss by a bias that
is updated in a backpropagation-free manner \citep{wang2024auxiliary,
liu2024deepseekv3}. We ablate our routing and sparsity setup in
\Cref{app:ablations}. We fully describe the final architecture in
\Cref{app:architecture}.

\paragraph{Aspirational connection to the cerebral cortex.} To some
extent, one may view \ours{} as an architecture that
tries to emulate the neural architecture of the cerebral cortex of a
brain \citep{Hawkins-2021}: the blocks correspond to different lobes
or functional regions in the cortex, from low-level to high-level,
and the many columns in each block emulate the cortical columns
within each region.

\section{Compute-Efficient Pretraining with \ours{}}\label{sec:pretraining}

In this section we show that \ours{} is more compute-efficient than
strong iso-sparsity MoE transformer and dense transformer baselines in language
model pretraining, and that the advantage does not shrink as the
scale grows to the limits of our compute. Following prior scaling law
studies of MoE transformers (see \Cref{app:related-work}), we
compare architectures at the same sparsity, fit a scaling law to each,
and report the projected compute that each baseline needs to reach the
held-out loss of \ours{}.

\paragraph{Setup.} We compare three architectures at four scales, which we
call 100M, 200M, 500M, and 1B after the \textit{active} parameter
count. The \textit{dense transformer} has width \(D \in \{480, 640,
960, 1280\}\), depth \(\{12, 16, 24, 32\}\), \(16\) heads, and MLP width
\(4D\). The \textit{MoE transformer} has the same shape, with each MLP
replaced by \(E = 24\) experts of which \(K_{\mathrm{e}} = 3\) are active, so
that its active MLP width is also \(4D\). \ours{} uses \(C = 8\)
columns, increasing with depth from \(K_{\mathrm{c}} = 3\) in the first
block to \(K_{\mathrm{c}} = 8\) (all columns active) in the
final block. Both baselines use full causal attention with the
attention residual
\citep{kimi2026attnres}. At each
scale, \ours{} uses no more training FLOPs than either
baseline,\footnote{Our FLOP count includes attention FLOPs
    (\Cref{app:configurations}). Under the common approximation that
    counts only linear layers' FLOPs, \ours{} uses \(4\)--\(8\%\)
    more FLOPs than the
baselines at each scale, at length \(S=4096\).} and stores
approximately the same number of parameters as the MoE
transformer. We consider further baselines in \Cref{app:ablations}.
All models are trained
on DCLM-baseline \citep{li2024datacomp} at context length \(S = 4096\) for
\(50\) tokens per active parameter (beyond the roughly \(20\) that are
    compute optimal for dense transformers;
\citealp{hoffmann2022training}).The parameterization and
optimizer settings used in these experiments are specified in
\Cref{app:experimental-parameterization}. We report
the cross-entropy on held-out
DCLM-baseline tokens at the end of training.

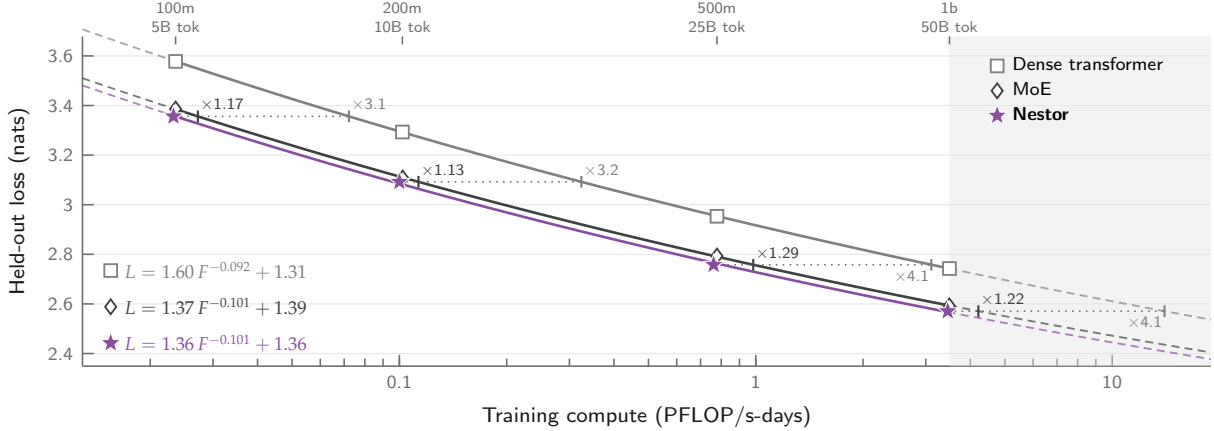
\begin{figure}[t]
    \centering
    {\def\scalingwidth{\textwidth}\def\scalingheight{0.36\textwidth}%
        \def\scalingwide{1}\def\scalingeqstep{0.11}%
    \providecommand{\scalingprefix}{scaling}
\providecommand{\scalingdense}{dense}
\providecommand{\scalingdenselabel}{Dense transformer}
\providecommand{\scalingwidth}{0.62\textwidth}
\providecommand{\scalingheight}{0.5\textwidth}
\providecommand{\scalingwide}{0}
\providecommand{\scalingeqstep}{0.075}
\providecommand{\scalingextra}{}
\providecommand{\scalingeqbase}{2}
\providecommand{\scalingline}{solid}
\providecommand{\scalingdenseoff}{0}
\providecommand{\scalingresid}{0}
\providecommand{\scalingresidheight}{0.34\textwidth}
\expsnow{\noexpand\pgfplotstableread[col sep=comma]{figures/exps/raw_data/derived/\scalingprefix_fits.csv}\noexpand\scalingfits}
\expsnow{\noexpand\pgfplotstableread[col sep=comma]{figures/exps/raw_data/derived/\scalingprefix_multiples.csv}\noexpand\scalingmult}
\expsnow{\noexpand\pgfplotstableread[col sep=comma]{figures/exps/raw_data/derived/\scalingprefix_rungs.csv}\noexpand\scalingrungs}
\expsnow{\noexpand\pgfplotstableread[col sep=comma]{figures/exps/raw_data/derived/\scalingprefix_axis.csv}\noexpand\scalingaxis}
\pgfplotstablegetelem{0}{xlo}\of\scalingaxis\let\xlo\pgfplotsretval
\pgfplotstablegetelem{0}{xhi}\of\scalingaxis\let\xhi\pgfplotsretval
\pgfplotstablegetelem{0}{ylo}\of\scalingaxis\let\ylo\pgfplotsretval
\pgfplotstablegetelem{0}{yhi}\of\scalingaxis\let\yhi\pgfplotsretval
\pgfplotstablegetelem{0}{xmin}\of\scalingfits\let\xdatamin\pgfplotsretval
\pgfplotstablegetelem{0}{xmax}\of\scalingfits\let\xdatamax\pgfplotsretval
\edef\scalingpoints{figures/exps/raw_data/derived/\scalingprefix_points.csv}
\def\scalingpanel{0}
\newcommand{\scalingseries}[7]{%
  \pgfplotstablegetelem{#2}{A}\of\scalingfits\let\A\pgfplotsretval
  \pgfplotstablegetelem{#2}{alpha}\of\scalingfits\let\al\pgfplotsretval
  \pgfplotstablegetelem{#2}{B}\of\scalingfits\let\B\pgfplotsretval
  \pgfplotstablegetelem{#2}{xmin}\of\scalingfits\let\xminv\pgfplotsretval
  \pgfplotstablegetelem{#2}{xmax}\of\scalingfits\let\xmaxv\pgfplotsretval
  \pgfplotstablegetelem{#2}{eq}\of\scalingfits\let\eq\pgfplotsretval
  \ifnum\scalingpanel=0
  \expsnow{\noexpand\addplot[#3, line width=1pt, \scalingline, domain=\xminv:\xmaxv, samples=60, forget plot] {\A*x^(-\al)+\B};}
  \expsnow{\noexpand\addplot[#3, line width=0.7pt, densely dashed, opacity=0.7, domain=\xlo:\xminv, samples=30, forget plot] {\A*x^(-\al)+\B};}
  \expsnow{\noexpand\addplot[#3, line width=0.7pt, densely dashed, opacity=0.7, domain=\xmaxv:\xhi, samples=30, forget plot] {\A*x^(-\al)+\B};}
  \addplot[#3, forget plot, mark=none, draw=none,
           error bars/.cd, y dir=plus, y explicit, error mark=none, error bar style={#3, opacity=0.6, line width=0.7pt}]
    table[col sep=comma, x=x, y=ce, y error=twin, discard if not={series}{#1}] {\scalingpoints};
  \addplot[exps mark, mark=#4, mark size=#5, mark options={#7}]
    table[col sep=comma, x=x, y=ce, discard if not={series}{#1}] {\scalingpoints};
  \addlegendentry{#6}
  \ifnum\scalingwide=1
    \pgfmathsetmacro\eqx{\xlo*exp(0.025*ln(\xhi/\xlo))}%
    \pgfmathsetmacro\eqy{\ylo+(\yhi-\ylo)*(0.07+\scalingeqstep*(\scalingeqbase-#2))}%
    \expsnow{\noexpand\addplot[exps mark, mark=#4, mark size=#5, mark options={#7}, forget plot] coordinates {(\eqx,\eqy)};}
    \expsnow{\noexpand\node[anchor=west, font=\noexpand\sffamily\noexpand\scriptsize, text=#3, inner sep=1pt, xshift=4pt]
      at (axis cs:\eqx,\eqy) {\eq};}
  \else
    \expsnow{\noexpand\node[anchor=south west, font=\noexpand\sffamily\noexpand\scriptsize, text=#3, inner sep=1pt]
      at (rel axis cs:0.015, {0.01 + \scalingeqstep*(\scalingeqbase-#2)}) {\eq};}
  \fi
  \else
  \expsnow{\noexpand\addplot[#3, line width=1pt, \scalingline, domain=\xminv:\xmaxv, samples=60, forget plot] {\A*x^(-\al)+\B-(\Am*x^(-\alm)+\Bm)};}
  \addplot[#3, forget plot, mark=none, draw=none,
           error bars/.cd, y dir=plus, y explicit, error mark=none, error bar style={#3, opacity=0.6, line width=0.7pt}]
    table[col sep=comma, x=x, y=rel, y error=twin, discard if not={series}{#1}] {\scalingpoints};
  \addplot[exps mark, forget plot, mark=#4, mark size=#5, mark options={#7}]
    table[col sep=comma, x=x, y=rel, discard if not={series}{#1}] {\scalingpoints};
  \fi
}
\newcommand{\scalingallseries}{%
  \ifnum\scalingdenseoff=0
    \scalingseries{\scalingdense}{0}{expDense}{square*}{2.4pt}{\scalingdenselabel}{fill=white, draw=expDense, line width=0.7pt}
  \fi
  \scalingseries{moe}{1}{expMoe}{diamond*}{2.8pt}{MoE}{fill=white, draw=expMoe, line width=0.7pt}
  \scalingseries{cortex}{2}{expCortex}{fstar}{4pt}{\bfseries\ours{}}{fill=expCortex, draw=white, line width=0.4pt}
  \scalingextra}
\ifnum\scalingresid=1
  \pgfplotsset{scaling x axis/.style={xticklabel=\empty}}
\else
  \pgfplotsset{scaling x axis/.style={xlabel={Training compute (PFLOP/s-days)}}}
\fi
\begin{tikzpicture}
\begin{axis}[exps, name=scalingmain,
  width=\scalingwidth, height=\scalingheight,
  xmode=log, xmin=\xlo, xmax=\xhi, ymin=\ylo, ymax=\yhi, ytick distance=0.2,
  xtick={0.01,0.1,1,10}, xticklabels={0.01,0.1,1,10}, log ticks with fixed point,
  scaling x axis, ylabel={Held-out loss (nats)},
  legend pos=north east, legend style={row sep=-1pt},
  clip=false,
]
  \expsnow{\noexpand\addplot[draw=none, fill=expGrid, fill opacity=0.45, forget plot]
    coordinates {(\xdatamax,\ylo) (\xhi,\ylo) (\xhi,\yhi) (\xdatamax,\yhi)} \noexpand\closedcycle;}
  \scalingallseries
  \pgfplotsinvokeforeach{0,1,2,3}{
    \pgfplotstablegetelem{#1}{xstar}\of\scalingmult\let\xs\pgfplotsretval
    \pgfplotstablegetelem{#1}{y}\of\scalingmult\let\ys\pgfplotsretval
    \pgfplotstablegetelem{#1}{xmoe}\of\scalingmult\let\xm\pgfplotsretval
    \pgfplotstablegetelem{#1}{moe}\of\scalingmult\let\lm\pgfplotsretval
    \ifnum\scalingdenseoff=0
      \pgfplotstablegetelem{#1}{xdense}\of\scalingmult\let\xd\pgfplotsretval
      \pgfplotstablegetelem{#1}{dense}\of\scalingmult\let\ld\pgfplotsretval
    \else
      \let\xd\xm
    \fi
    \expsnow{\noexpand\draw[expMuted, dotted, line width=0.6pt] (axis cs:\xs,\ys) -- (axis cs:\xd,\ys);}
    \expsnow{\noexpand\draw[expMoe, line width=0.8pt] ([yshift=-2.2pt]axis cs:\xm,\ys) -- ([yshift=2.2pt]axis cs:\xm,\ys)
      node[above right, exps annot, text=expMoe, inner sep=0.5pt, yshift=-1pt] {\lm};}
    \ifnum\scalingdenseoff=0
      \ifnum#1>1 \def\dplace{below left, yshift=-3.4pt}\else\def\dplace{above right, yshift=-1pt}\fi
      \expsnow{\noexpand\draw[expDense, line width=0.8pt] ([yshift=-2.2pt]axis cs:\xd,\ys) -- ([yshift=2.2pt]axis cs:\xd,\ys)
        node[\dplace, exps annot, text=expDense, inner sep=0.5pt] {\ld};}
    \fi
  }
  \pgfplotsinvokeforeach{0,1,2,3}{
    \pgfplotstablegetelem{#1}{x}\of\scalingrungs\let\xr\pgfplotsretval
    \pgfplotstablegetelem{#1}{label}\of\scalingrungs\let\lr\pgfplotsretval
    \ifnum\scalingwide=1
      \expsnow{\noexpand\draw[expMuted, line width=0.6pt] (axis cs:\xr,\yhi) -- ([yshift=-3pt]axis cs:\xr,\yhi)
        (axis cs:\xr,\yhi) node[above, exps annot, text=expMuted, align=center, inner sep=1.5pt] {\lr};}
    \else
      \expsnow{\noexpand\draw[expGrid, dashed, line width=0.6pt] ({axis cs:\xr,0}|-{rel axis cs:0,0}) -- ({axis cs:\xr,0}|-{rel axis cs:0,1})
        node[above, exps annot, text=expMuted, align=center, inner sep=1.5pt] {\lr};}
    \fi
  }
\end{axis}
\ifnum\scalingresid=1
  \def\scalingpanel{1}
  \pgfplotstablegetelem{1}{A}\of\scalingfits\let\Am\pgfplotsretval
  \pgfplotstablegetelem{1}{alpha}\of\scalingfits\let\alm\pgfplotsretval
  \pgfplotstablegetelem{1}{B}\of\scalingfits\let\Bm\pgfplotsretval
  \pgfplotstablegetelem{0}{rello}\of\scalingaxis\let\rlo\pgfplotsretval
  \pgfplotstablegetelem{0}{relhi}\of\scalingaxis\let\rhi\pgfplotsretval
  \begin{axis}[exps,
    at={(scalingmain.south west)}, anchor=north west, yshift=-10pt,
    width=\scalingwidth, height=\scalingresidheight,
    xmode=log, xmin=\xlo, xmax=\xhi, ymin=\rlo, ymax=\rhi, ytick distance=0.01,
    yticklabel={\pgfmathparse{abs(\tick)}\ifdim\tick pt<0pt \textminus\fi\pgfmathprintnumber{\pgfmathresult}},
    xtick={0.01,0.1,1,10}, xticklabels={0.01,0.1,1,10}, log ticks with fixed point,
    xlabel={Training compute (PFLOP/s-days)}, ylabel={Loss relative to\\the MoE law (nats)}, ylabel style={align=center},
    clip=false,
  ]
    \expsnow{\noexpand\addplot[draw=none, fill=expGrid, fill opacity=0.45, forget plot]
      coordinates {(\xdatamax,\rlo) (\xhi,\rlo) (\xhi,\rhi) (\xdatamax,\rhi)} \noexpand\closedcycle;}
    \pgfplotsinvokeforeach{0,1,2,3}{
      \pgfplotstablegetelem{#1}{x}\of\scalingrungs\let\xr\pgfplotsretval
      \expsnow{\noexpand\draw[expGrid, dashed, line width=0.6pt] ({axis cs:\xr,0}|-{rel axis cs:0,0}) -- ({axis cs:\xr,0}|-{rel axis cs:0,1});}
    }
    \scalingallseries
  \end{axis}
\fi
\end{tikzpicture}}
    \caption{\small \textbf{Held-out loss scaling laws} for \ours{}, MoE,
        and dense models. Each law is
        fitted as $L(F) = A\,F^{-\alpha} + L_{\infty}$, where $F$ is
        the training FLOPs. \ours{} reaches the
        lowest hold-out loss at every scale. The horizontal
        ticks mark the projected compute that the MoE and dense
        baselines need to match
    \ours{}.}
    \label{fig:pretraining}
\end{figure}

\paragraph{\ours{} is more compute-efficient at every scale.}
\ours{} achieves lower held-out loss than both baselines at all four
model scales, as shown in~\Cref{fig:pretraining}; the losses,
parameter counts, and training compute at each scale are in
\Cref{tab:exp-rungs} of \Cref{app:pretraining-results}. To quantify
the compute savings at equal
loss, we fit each architecture's loss-compute curve with $L(F) =
A F^{-\alpha} + L_{\infty}$ \citep{kaplan2020scaling, hoffmann2022training}.
For each measured \ours{} result, we trace a horizontal line to each
baseline's fitted curve. The intersection gives an estimate of the
baseline compute needed to reach the same loss. We report this
estimate, normalized by the training compute of \ours{}, as the
horizontal ticks in \Cref{fig:pretraining} and the last column of
\Cref{tab:exp-rungs}. At
the largest scale, matching the loss of \ours{}  is estimated to need
$1.22\times$ as much compute for the MoE transformer and $4.1\times$
for the dense transformer. The fitted exponents of \ours{} and the
MoE transformer agree to three digits, and \ours{} has a smaller
estimated coefficient and irreducible loss.

\section{\ours{} Enables Efficient Continual
Learning}\label{sec:continual-pretraining}

A pretrained model is often customized via continual training to
learn new capabilities while retaining old ones. Concretely,
pretrained models are
often continually pre-trained on new domains \citep{gururangan2020dont,
roziere2023code, azerbayev2023llemma, shao2024deepseekmath} and longer contexts
\citep{xiong2024effective, fu2024data, grattafiori2024llama3}; in
this setting, it is important to
retain its original capabilities on old domains and shorter contexts,
which continual training tends to erode \citep{french1999catastrophic,
luo2025empirical}.
In this section we
show that \ours{} is well suited to both settings: it adapts to new domains
as well as the MoE transformer while forgetting less
(\Cref{sec:domain-adaptation}), and it extends to long contexts more
efficiently and effectively (\Cref{sec:long-context-extension}).

\subsection{Domain Adaptation}\label{sec:domain-adaptation}

\paragraph{Setup.} We compare the 500M \ours{}, 500M MoE transformer,
and 1B dense transformer, selected for their similar pretraining
losses. All models first undergo context extension to
\(32\)k using equal token budgets on DCLM. Details of the starting
checkpoints and their initial performance are in
\Cref{app:domain-adaptation-settings}. Each model is then trained
separately on \(0.5\)B tokens from each of four target domains,
without replay of pretraining data (a common way to reduce
forgetting; \citealp{ibrahim2024simple}): books (PG-19;
\citealp{rae2019compressive}), mathematics (OpenWebMath;
\citealp{paster2023openwebmath}), code (AlgebraicStack;
\citealp{azerbayev2023llemma}), and German (FineWeb-2;
\citealp{penedo2025fineweb2}). We freeze the router parameters of
both sparse models (cf.\ \citealp{li2025theory}) and update all
remaining trainable parameters; the dense model is fully fine-tuned.
We measure adaptation by the reduction in held-out loss on the target
corpus and forgetting by the increase in held-out loss on the
pretraining corpus, both relative to the model before adaptation. We
also repeat the experiment with \(25\%\) of the training tokens drawn
from the pretraining corpus, keeping the total budget fixed at
\(0.5\)B tokens; details and results are in
\Cref{app:domain-adaptation-settings}.
\Cref{tab:domain-adaptation-results} reports the held-out losses
before and after adaptation.

\begin{figure}[t]
    \centering
    \providecommand{\dadense}{dense1b}
\providecommand{\dadenselabel}{Dense 1B}
\def\darelpoints{figures/exps/raw_data/derived/domain_adaptation_rel.csv}
\colorlet{daChrome}{black!35}
\pgfplotsset{
  da ours/.style={only marks, mark=fstar, mark size=3.3pt, mark options={fill=expCortex, draw=expCortex!80!black, line width=0.3pt}},
  da moe/.style={only marks, mark=diamond*, mark size=2.6pt, mark options={fill=white, draw=expMoe, line width=0.6pt}},
  da dense/.style={only marks, mark=square*, mark size=2.1pt, mark options={fill=white, draw=expDense, line width=0.6pt}},
  da faded/.style={opacity=0.4},
}
\newcommand{\darelpoint}[6]{%
  \addplot[#6, forget plot]
    table[col sep=comma, x expr=#4, y=#2, discard if not={corpus}{#1}, discard if not={arm}{#3},
          discard if not={replay_pct}{#5}] {\darelpoints};
}
\newcommand{\darelpanel}[2]{%
  \draw[expMuted, line width=0.5pt] ({rel axis cs:0,0}|-{axis cs:0,0}) -- ({rel axis cs:1,0}|-{axis cs:0,0});
  \addplot[da moe, forget plot] coordinates {(1,0)};
  \darelpoint{#1}{#2}{\dadense}{2.18}{25}{da dense, da faded}
  \darelpoint{#1}{#2}{\dadense}{1.82}{0}{da dense}
  \darelpoint{#1}{#2}{flatq12}{0.18}{25}{da ours, da faded}
  \darelpoint{#1}{#2}{flatq12}{-0.18}{0}{da ours}
}
\begin{tikzpicture}
\begin{groupplot}[exps,
  group style={group size=5 by 2, horizontal sep=0.3cm, vertical sep=0.62cm, yticklabels at=edge left},
  scale only axis, width=0.126\textwidth, height=0.11\textwidth,
  xmin=-0.7, xmax=2.7, xtick={0,1,2}, xticklabels={}, x tick style={draw=none},
  enlarge y limits=false,
  yticklabel={\pgfmathprintnumber{\tick}\%},
  axis line style={daChrome, line width=0.4pt}, tick style={daChrome, line width=0.4pt},
  ytick align=outside, tick label style={font=\sffamily\fontsize{6}{7}\selectfont, black!55},
  yticklabel style={/pgf/number format/precision=0, /pgf/number format/fixed, text width=2.2em, align=right},
  grid style={black!9, line width=0.3pt},
  every axis title/.append style={font=\sffamily\scriptsize, yshift=-3pt},
]
\nextgroupplot[title={Books}, ymin=-6, ymax=6, ytick={-4,0,4}]              \darelpanel{pg19}{gain_rel}
\nextgroupplot[title={Math}, ymin=-6, ymax=6, ytick={-4,0,4}]               \darelpanel{owm}{gain_rel}
\nextgroupplot[title={Code}, ymin=-6, ymax=6, ytick={-4,0,4}]               \darelpanel{algstack}{gain_rel}
\nextgroupplot[title={German}, ymin=-6, ymax=6, ytick={-4,0,4}]             \darelpanel{fw2de}{gain_rel}
\nextgroupplot[title={\bfseries Average}, ymin=-6, ymax=6, ytick={-4,0,4}]  \darelpanel{avg}{gain_rel}
\nextgroupplot[ymin=-12, ymax=36, ytick={-10,0,10,20,30}]                   \darelpanel{pg19}{forget_len_rel}
\nextgroupplot[ymin=-12, ymax=36, ytick={-10,0,10,20,30}]                   \darelpanel{owm}{forget_len_rel}
\nextgroupplot[ymin=-12, ymax=36, ytick={-10,0,10,20,30}]                   \darelpanel{algstack}{forget_len_rel}
\nextgroupplot[ymin=-12, ymax=36, ytick={-10,0,10,20,30}]                   \darelpanel{fw2de}{forget_len_rel}
\nextgroupplot[ymin=-12, ymax=36, ytick={-10,0,10,20,30}, legend to name=exps:domainadaptationrel\dadense, legend columns=-1,
               legend style={font=\sffamily\scriptsize, /tikz/every even column/.append style={column sep=0.9em}}]
  \darelpanel{avg}{forget_len_rel}
  \addlegendimage{da ours}\addlegendentry{\bfseries \ours{} 500M}
  \addlegendimage{da moe}\addlegendentry{MoE 500M (reference)}
  \addlegendimage{da dense}\addlegendentry{\dadenselabel}
  \addlegendimage{only marks, mark=*, mark size=2pt, expMuted}\addlegendentry{no replay}
  \addlegendimage{only marks, mark=*, mark size=2pt, expMuted, opacity=0.4}\addlegendentry{25\thinspace\% replay}
\end{groupplot}
\node[below=0.04cm, font=\sffamily\scriptsize, text=expInk] at ($(group c1r1.south)!0.5!(group c5r1.south)$) {Gain $\uparrow$ (\% relative to the MoE)};
\node[below=0.04cm, font=\sffamily\scriptsize, text=expInk] at ($(group c1r2.south)!0.5!(group c5r2.south)$) {Forgetting $\downarrow$ (\% relative to the MoE)};
\node[below=0.38cm] at ($(group c1r2.south)!0.5!(group c5r2.south)$) {\pgfplotslegendfromname{exps:domainadaptationrel\dadense}};
\end{tikzpicture}
    \caption{\small \textbf{Domain adaptation and forgetting relative to the MoE
        transformer.} \textbf{Top:} reduction in held-out loss on the target
        corpus. \textbf{Bottom:} increase in held-out loss on the pretraining
        corpus. Both are measured relative to each model before adaptation
    and shown as a percentage of the MoE transformer's value. }
    \label{fig:domain-adaptation}
\end{figure}
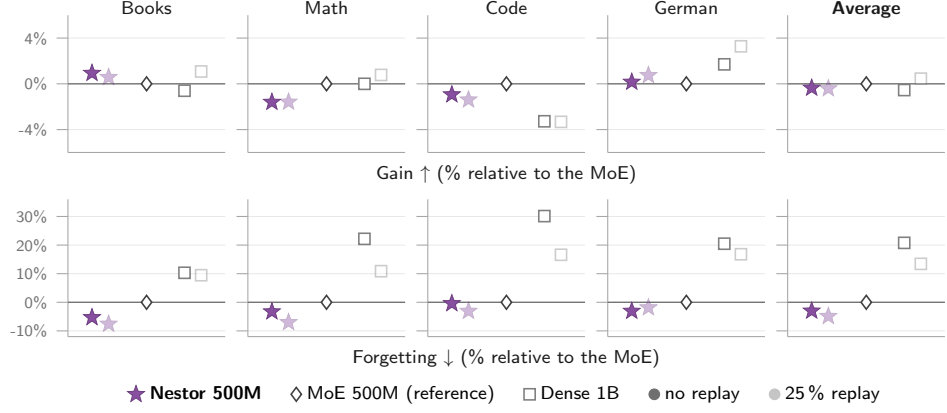

\paragraph{\ours{} adapts as well as the baselines with less forgetting.}
All three models achieve similar adaptation gains: averaged across
the four target domains, the gains of \ours{} and the 1B dense
transformer are both within \(1\%\) of the 500M MoE transformer's
(\Cref{fig:domain-adaptation}). The models differ more in how much
they forget. \ours{} forgets \(3\%\) less than the MoE on average,
whereas the dense model forgets \(21\%\).
Thus, \ours{} has a modest retention advantage over the MoE,
and both sparse models forget substantially less than the dense model.
Replaying \(25\%\) pretraining data reduces forgetting by about
\(70\%\) across all three models at a cost of about \(6\%\) of
the adaptation gain and preserves this ordering: \ours{} forgets
\(5\%\) less than the MoE on average and the dense model \(13\%\)
more, while the two sparse models' gains remain within \(2\%\)
of each other (\Cref{fig:domain-adaptation}, faded marks).

\subsection{Long Context Extension}\label{sec:long-context-extension}

\begin{figure}[t]
    \centering
    \providecommand{\cptprefix}{cpt}
\providecommand{\cptdense}{dense}
\providecommand{\cptdenselabel}{dense transformer}
\providecommand{\cptextra}[1]{}
\providecommand{\cptyticks}{0,-0.02,-0.04,-0.06}
\providecommand{\cptlegend}{\cptlegenddefault}
\providecommand{\cptlegendcols}{4}
\providecommand{\cptbottomheight}{0.2\textwidth}  %
\expsnow{\noexpand\pgfplotstableread[col sep=comma]{figures/exps/raw_data/derived/\cptprefix_axis.csv}\noexpand\cptaxis}
\edef\cptdeltafile{figures/exps/raw_data/derived/\cptprefix_delta.csv}
\edef\cpttwinsfile{figures/exps/raw_data/derived/\cptprefix_twins.csv}
\newcommand{\cptlim}[3]{\pgfplotstablegetelem{#1}{#2}\of\cptaxis\let#3\pgfplotsretval}
\cptlim{0}{ylo}{\cptyloA}\cptlim{0}{yhi}{\cptyhiA}\cptlim{0}{tlo}{\cpttloA}\cptlim{0}{thi}{\cptthiA}\cptlim{0}{xmax}{\cptxmaxA}
\cptlim{1}{ylo}{\cptyloB}\cptlim{1}{yhi}{\cptyhiB}\cptlim{1}{tlo}{\cpttloB}\cptlim{1}{thi}{\cptthiB}\cptlim{1}{xmax}{\cptxmaxB}
\cptlim{2}{ylo}{\cptyloC}\cptlim{2}{yhi}{\cptyhiC}\cptlim{2}{tlo}{\cpttloC}\cptlim{2}{thi}{\cptthiC}\cptlim{2}{xmax}{\cptxmaxC}
\cptlim{3}{ylo}{\cptyloD}\cptlim{3}{yhi}{\cptyhiD}\cptlim{3}{tlo}{\cpttloD}\cptlim{3}{thi}{\cptthiD}\cptlim{3}{xmax}{\cptxmaxD}
\newcommand{\cptseries}[5]{%
  \addplot[#3, mark=none, forget plot]
    table[col sep=comma, x=x, y=delta, discard if not={scale}{#1}, discard if not={series}{#2}]
    {\cptdeltafile};
  \addplot[only marks, #4, forget plot]
    table[col sep=comma, x=x, y=delta, discard if not={scale}{#1}, discard if not={series}{#2}, discard if={x}{4}]
    {\cptdeltafile};
  \addplot[only marks, #5, forget plot]
    table[col sep=comma, x=x, y=delta, discard if not={scale}{#1}, discard if not={series}{#2}, discard if not={x}{4}]
    {\cptdeltafile};
}
\newcommand{\cpttwin}[3]{%
  \addplot[#3, opacity=0.5, line width=0.6pt, mark=none, forget plot]
    table[col sep=comma, x=x, y=y, discard if not={scale}{#1}, discard if not={series}{#2}]
    {\cpttwinsfile};
}
\colorlet{cptChrome}{black!35}
\colorlet{cptOursA}{expCortex}
\colorlet{cptOursB}{expCortexMid}    %
\colorlet{cptOursC}{expCortexLight}
\providecolor{cptSwaA}{named}{expMoe}
\colorlet{cptSwaB}{cptSwaA!65!white}
\colorlet{cptSwaC}{cptSwaA!42!white}
\pgfplotsset{
  cpt fresh line/.style={cptOursA, line width=0.8pt},
  cpt fresh mark/.style={mark=fstar, mark size=3pt, mark options={fill=cptOursA, draw=cptOursA!80!black, line width=0.3pt}},
  cpt fresh open/.style={mark=fstar, mark size=3pt, mark options={fill=white, draw=cptOursA, line width=0.5pt}},
  cpt repeat line/.style={cptOursB, dash dot, line width=0.6pt},
  cpt repeat mark/.style={mark=fstar, mark size=2.8pt, mark options={fill=cptOursB, draw=white, line width=0.3pt}},
  cpt repeat open/.style={mark=fstar, mark size=2.8pt, mark options={fill=white, draw=cptOursB, line width=0.5pt}},
  cpt isotok line/.style={cptOursC, line width=0.55pt},
  cpt isotok mark/.style={mark=fstar, mark size=2.8pt, mark options={fill=cptOursC, draw=white, line width=0.3pt}},
  cpt isotok open/.style={mark=fstar, mark size=2.8pt, mark options={fill=white, draw=cptOursC, line width=0.5pt}},
  cpt moe line/.style={black!45, dashed, line width=0.4pt},
  cpt moe mark/.style={mark=diamond*, mark size=1.8pt, mark options={fill=expMoe, draw=expMoe, line width=0.3pt, solid}},
  cpt moe open/.style={mark=diamond*, mark size=1.8pt, mark options={fill=white, draw=expMoe, line width=0.45pt, solid}},
  cpt dense line/.style={black!40, line width=0.55pt},
  cpt dense mark/.style={mark=square*, mark size=1.7pt, mark options={fill=black!40, draw=white, line width=0.3pt}},
  cpt dense open/.style={mark=square*, mark size=1.7pt, mark options={fill=white, draw=black!40, line width=0.45pt}},
  cpt swa fresh line/.style={cptSwaA, line width=0.7pt},
  cpt swa fresh mark/.style={mark=triangle*, mark size=2.4pt, mark options={fill=cptSwaA, draw=cptSwaA!80!black, line width=0.3pt, solid}},
  cpt swa fresh open/.style={mark=triangle*, mark size=2.4pt, mark options={fill=white, draw=cptSwaA, line width=0.5pt, solid}},
  cpt swa repeat line/.style={cptSwaB, dash dot, line width=0.55pt},
  cpt swa repeat mark/.style={mark=triangle*, mark size=2.2pt, mark options={fill=cptSwaB, draw=white, line width=0.3pt, solid}},
  cpt swa repeat open/.style={mark=triangle*, mark size=2.2pt, mark options={fill=white, draw=cptSwaB, line width=0.5pt, solid}},
  cpt swa isotok line/.style={cptSwaC, line width=0.5pt},
  cpt swa isotok mark/.style={mark=triangle*, mark size=2.2pt, mark options={fill=cptSwaC, draw=white, line width=0.3pt, solid}},
  cpt swa isotok open/.style={mark=triangle*, mark size=2.2pt, mark options={fill=white, draw=cptSwaC, line width=0.5pt, solid}},
  cpt fresh legend/.style={cpt fresh line, cpt fresh mark},
  cpt repeat legend/.style={cpt repeat line, cpt repeat mark},
  cpt isotok legend/.style={cpt isotok line, cpt isotok mark},
  cpt moe legend/.style={cpt moe line, cpt moe mark},
  cpt dense legend/.style={cpt dense line, cpt dense mark},
  cpt swa fresh legend/.style={cpt swa fresh line, cpt swa fresh mark},
  cpt swa repeat legend/.style={cpt swa repeat line, cpt swa repeat mark},
  cpt swa isotok legend/.style={cpt swa isotok line, cpt swa isotok mark},
  cpt open legend/.style={only marks, mark=fstar, mark size=2.8pt, mark options={fill=white, draw=black!45, line width=0.5pt}},
}
\newcommand{\cpttop}[1]{\cptseries{#1}{\cptdense}{cpt dense line}{cpt dense mark}{cpt dense open}}
\newcommand{\cpttwins}[1]{\cpttwin{#1}{isotok}{cptOursC}\cpttwin{#1}{isoflop_lr}{cptOursA}\cpttwin{#1}{isoflop_ep_lr2}{cptOursB}}
\newcommand{\cptbottom}[1]{%
  \cptextra{#1}%
  \cptseries{#1}{moe}{cpt moe line}{cpt moe mark}{cpt moe open}
  \cptseries{#1}{isotok}{cpt isotok line}{cpt isotok mark}{cpt isotok open}
  \cptseries{#1}{isoflop_ep_lr2}{cpt repeat line}{cpt repeat mark}{cpt repeat open}
  \cptseries{#1}{isoflop_lr}{cpt fresh line}{cpt fresh mark}{cpt fresh open}
}
\def\cptlegenddefault{%
  \addlegendimage{empty legend}\addlegendentry{\bfseries\ours{}}
  \addlegendimage{cpt fresh legend}\addlegendentry{equal FLOPs, fresh}
  \addlegendimage{cpt repeat legend}\addlegendentry{equal FLOPs, repeated}
  \addlegendimage{cpt isotok legend}\addlegendentry{equal tokens}
  \addlegendimage{empty legend}\addlegendentry{baselines}
  \addlegendimage{cpt dense legend}\addlegendentry{\cptdenselabel}
  \addlegendimage{cpt moe legend}\addlegendentry{MoE (reference, $1.00\times$)}
  \addlegendimage{cpt open legend}\addlegendentry{open: pretrained, at 4k}}
\newcommand{\cptbreak}[1]{%
  \draw[cptChrome, line width=0.5pt] ([xshift=-2.2pt, yshift=-1pt]group c#1r1.south west) -- ([xshift=2.2pt, yshift=1.6pt]group c#1r1.south west);
  \draw[cptChrome, line width=0.5pt] ([xshift=-2.2pt, yshift=-1.6pt]group c#1r2.north west) -- ([xshift=2.2pt, yshift=1pt]group c#1r2.north west);
}
\begin{tikzpicture}
\begin{groupplot}[exps,
  group style={group size=4 by 2, horizontal sep=0.26cm, vertical sep=0.2cm,
               yticklabels at=edge left, xticklabels at=edge bottom},
  scale only axis, width=0.182\textwidth,
  xmode=log, log basis x=2, xmin=3.3,
  xtick={4,8,16,32,64}, xticklabels={4k,8k,16k,32k,64k}, minor xtick={},
  tick label style={font=\sffamily\fontsize{6}{7}\selectfont, black!55},
  axis line style={cptChrome, line width=0.4pt}, tick style={cptChrome, line width=0.4pt},
  grid style={black!7, line width=0.3pt},
  every axis title/.append style={font=\sffamily\scriptsize, yshift=-3pt},
]
\nextgroupplot[title={100M}, height=0.09\textwidth, xmax=\cptxmaxA, ymin=\cpttloA, ymax=\cptthiA, ytick={0.12,0.16,0.20}]  \cpttop{100m}
\nextgroupplot[title={200M}, height=0.09\textwidth, xmax=\cptxmaxB, ymin=\cpttloB, ymax=\cptthiB, ytick={0.12,0.16,0.20}]  \cpttop{200m}
\nextgroupplot[title={500M}, height=0.09\textwidth, xmax=\cptxmaxC, ymin=\cpttloC, ymax=\cptthiC, ytick={0.12,0.16,0.20}]  \cpttop{500m}
\nextgroupplot[title={1B},   height=0.09\textwidth, xmax=\cptxmaxD, ymin=\cpttloD, ymax=\cptthiD, ytick={0.12,0.16,0.20}]  \cpttop{1b}
\nextgroupplot[height=\cptbottomheight, xmax=\cptxmaxA, ymin=\cptyloA, ymax=\cptyhiA, ytick={\cptyticks}]   \cptbottom{100m}
\nextgroupplot[height=\cptbottomheight, xmax=\cptxmaxB, ymin=\cptyloB, ymax=\cptyhiB, ytick={\cptyticks}]   \cpttwins{200m}\cptbottom{200m}
\nextgroupplot[height=\cptbottomheight, xmax=\cptxmaxC, ymin=\cptyloC, ymax=\cptyhiC, ytick={\cptyticks}]   \cptbottom{500m}
\nextgroupplot[height=\cptbottomheight, xmax=\cptxmaxD, ymin=\cptyloD, ymax=\cptyhiD, ytick={\cptyticks},
               legend to name=exps:cpt\cptprefix, legend columns=\cptlegendcols,
               legend style={font=\sffamily\fontsize{6.5}{8}\selectfont, /tikz/every even column/.append style={column sep=0.5em},
                             /tikz/every odd column/.append style={column sep=0.15em}}]
  \cptbottom{1b}
  \cptlegend
\end{groupplot}
\cptbreak{1}\cptbreak{2}\cptbreak{3}\cptbreak{4}
\node[rotate=90, anchor=south, font=\sffamily\scriptsize, text=expInk] at ([xshift=-1pt]$(group c1r1.outer west)!0.5!(group c1r2.outer west)$) {$\Delta$ held-out loss vs the MoE (nats)};
\node[below=0.22cm, font=\sffamily\scriptsize, text=expInk] at ($(group c1r2.south)!0.5!(group c4r2.south)$) {Sequence length $S$};
\node[below=0.5cm] at ($(group c1r2.south)!0.5!(group c4r2.south)$) {\pgfplotslegendfromname{exps:cpt\cptprefix}};
\end{tikzpicture}
    \caption{\small \textbf{Long context extension by continued
        pretraining.} We show the
        held-out loss at each context length from $4$k to $64$k at all
        scales, relative to the MoE transformer of the same scale.
        \textbf{Top:} the dense transformer. \textbf{Bottom:} \ours{} at
        \textcolor{expCortex}{equal FLOPs with new documents}, at
        \textcolor{expCortexMid}{equal FLOPs with the documents of the
        equal-token run repeated over the run}, and at
        \textcolor{expCortexLight}{equal tokens}. The open markers at $4$k
    are the pretrained checkpoints.}
    \label{fig:long-context-extension}
\end{figure}
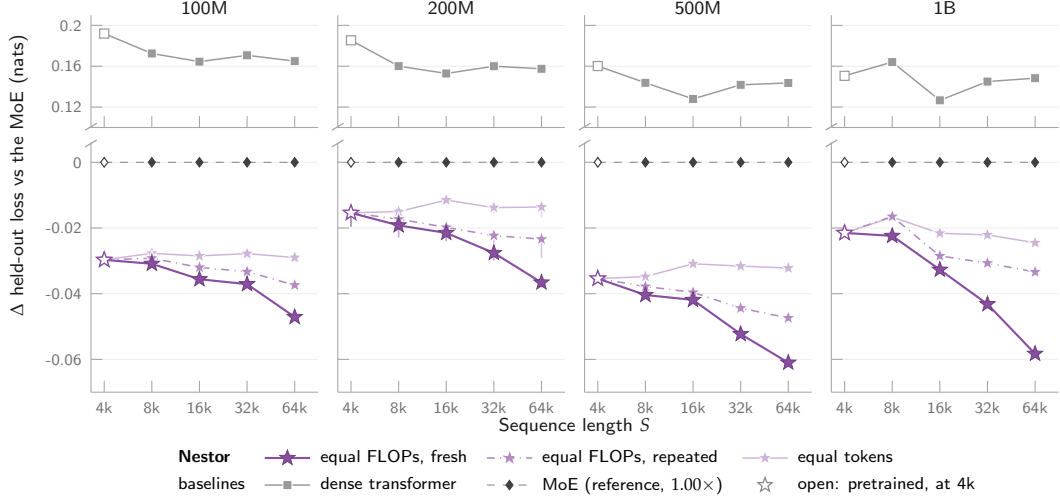

\paragraph{Setup.} Starting from the pretrained models of
\Cref{sec:pretraining}, which have context length \(4\)k, we extend each
model to a target context length \(S \in \{8\mathrm{k}, 16\mathrm{k},
32\mathrm{k}, 64\mathrm{k}\}\) by continued pretraining on long documents
\citep{xiong2024effective, fu2024data,gao2025train} of at least
\(S\) tokens from DCLM-baseline. Each target length is a
separate run from the \(4\)k checkpoint. The baselines are trained on
\(0.5\)B tokens. Since the FLOPs per token of \ours{} decrease relative to the
baselines as \(S\) grows (\Cref{fig:flops-vs-length}), we train
\ours{} in two settings: at \textit{equal
tokens} (\(0.5\)B), and at \textit{equal FLOPs}, where the saved compute is
spent on more tokens of either repeated or new documents. We report
the held-out loss at length \(S\), relative
to the MoE transformer of the same scale. Details of the data, the
schedules, and the losses behind
\Cref{fig:long-context-extension} are provided in
\Cref{app:long-context-extension-settings}.

\paragraph{The advantage of \ours{} grows with context length at equal compute.}
At equal tokens, \ours{} achieves gains from context extension
comparable to those of the MoE transformer, broadly preserving its
advantage before extension (\Cref{fig:long-context-extension}). As
the context length increases, its lower FLOPs per token allow more
training within the same compute budget (\Cref{fig:flops-vs-length}).
Spending these savings on repeated data produces a downward trend in
held-out loss relative to MoE, even without additional training data.
Spending them on fresh data, as one might expect, yields the largest
improvements. In
\Cref{app:scaling-laws} we quantify the saving: for equal shapes and
balanced routing, at equal FLOPs \ours{} trains on \(R(S) = (S_{0} + S) /
(S_{0} + \rho S)\) times as many tokens as the MoE, where \(\rho\) is
the mean fraction
\(K_{\mathrm{c}} / C\) of active columns and \(S_{0}\) the context
length at which the MoE transformer's attention FLOPs equal its other
FLOPs, and \(R(S)\) increases with \(S\) towards \(1 / \rho\).

\section{Theoretical Analysis}\label{sec:theory}

In this section we compare stylized versions of \ours{} and of the MoE
transformer on a statistical model with hierarchical latent
structure. In this analytically tractable model, we reproduce the
empirical pretraining compute efficiency and
continual training properties of the two architectures. We state the
results informally here, and give the formal statements and the proofs in
\Cref{app:theory}.

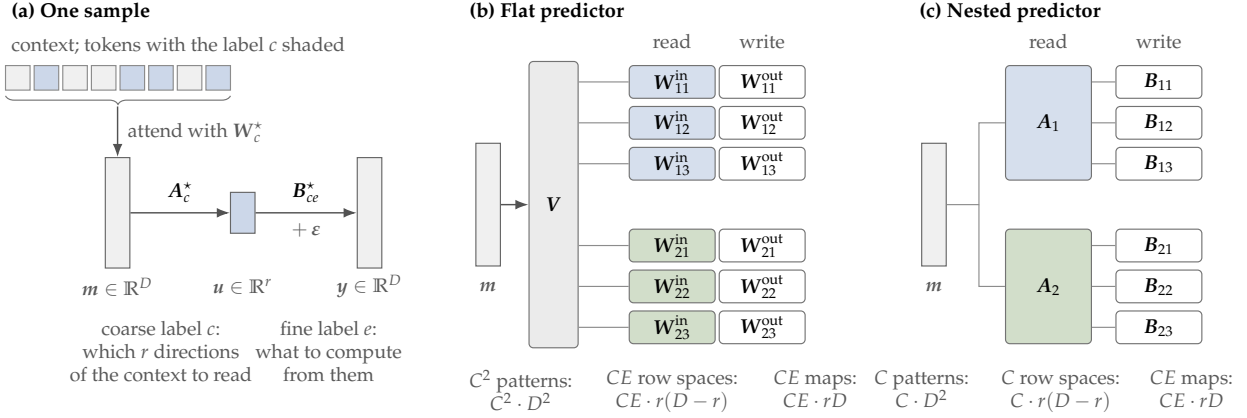
\begin{figure}[t]
    \centering
    \resizebox{\textwidth}{!}{\begin{tikzpicture}[
  x=1cm, y=1cm, >=Latex, font=\scriptsize,
  line/.style={draw=black!68, -{Latex[length=1.6mm]}, line width=.5pt},
  wire/.style={draw=black!45, line width=.4pt},
  tok/.style={draw=black!45, minimum size=.3cm, inner sep=0pt},
  vec/.style={draw=black!60, fill=black!6, minimum width=.3cm, inner sep=0pt},
  cell/.style={draw=black!55, rounded corners=1.5pt, minimum width=1.05cm,
               minimum height=.4cm, inner sep=0pt, align=center},
  lbl/.style={font=\fontsize{6.8}{8}\selectfont, align=center, text=black!65},
  ttl/.style={font=\scriptsize\bfseries, anchor=west}
]
\def\ca{houseBlue}
\def\cb{houseGreen}

\node[ttl] at (-.2,2.35) {(a) One sample};
\foreach \i/\f in {0/black!6, 1/\ca!35, 2/black!6, 3/black!6, 4/\ca!35,
                   5/\ca!35, 6/black!6, 7/\ca!35}
  \node[tok, fill=\f] (t\i) at (.35*\i,1.55) {};
\node[lbl, anchor=west] at (-.2,1.95) {context; tokens with the
  label $c$ shaded};
\draw[decorate, decoration={brace, mirror, amplitude=3pt}, draw=black!55]
  (-.15,1.33) -- (2.6,1.33);
\node[vec, minimum height=1.35cm] (m) at (1.22,-.1) {};
\draw[line] (1.22,1.2) -- node[right, lbl] {attend with $\vW_{c}^{\star}$} (m.north);
\node[lbl] at (1.22,-1.0) {$\vm \in \R^{D}$};
\node[vec, minimum height=.5cm, fill=\ca!35] (u) at (2.75,-.1) {};
\node[lbl] at (2.75,-1.0) {$\vu \in \R^{r}$};
\node[vec, minimum height=1.35cm] (y) at (4.3,-.1) {};
\node[lbl] at (4.3,-1.0) {$\vy \in \R^{D}$};
\draw[line] (m) -- node[above, font=\scriptsize] {$\vA_{c}^{\star}$} (u);
\draw[line] (u) -- node[above, font=\scriptsize] {$\vB_{ce}^{\star}$}
  node[below, lbl] {$+\,\veps$} (y);
\node[lbl, anchor=north] at (1.75,-1.3) {coarse label $c$:\\which $r$
  directions\\of the context to read};
\node[lbl, anchor=north] at (3.8,-1.3) {fine label $e$:\\what to compute\\from
  them};

\node[ttl] at (5.4,2.35) {(b) Flat predictor};
\node[vec, minimum height=1.5cm] (mb) at (5.75,0) {};
\node[lbl] at (5.75,-1.0) {$\vm$};
\node[cell, minimum width=.6cm, minimum height=3.5cm, fill=black!8]
  (V) at (6.55,0) {$\vV$};
\draw[line] (mb) -- (V);
\foreach \e/\yy in {1/1.5, 2/1.0, 3/.5} {
  \node[cell, fill=\ca!28] (bi\e) at (8.0,\yy)
    {$\vW^{\mathrm{in}}_{1\e}$};
  \node[cell] (bo\e) at (9.1,\yy) {$\vW^{\mathrm{out}}_{1\e}$};
  \draw[wire] (V.east |- bi\e) -- (bi\e.west);
}
\foreach \e/\yy in {1/-.5, 2/-1.0, 3/-1.5} {
  \node[cell, fill=\cb!28] (gi\e) at (8.0,\yy)
    {$\vW^{\mathrm{in}}_{2\e}$};
  \node[cell] (go\e) at (9.1,\yy) {$\vW^{\mathrm{out}}_{2\e}$};
  \draw[wire] (V.east |- gi\e) -- (gi\e.west);
}
\node[lbl] at (8.0,2.0) {read};
\node[lbl] at (9.1,2.0) {write};
\node[lbl, anchor=north] at (6.15,-1.9) {$C^{2}$ patterns:\\$C^{2} \cdot
  D^{2}$};
\node[lbl, anchor=north] at (8.0,-1.9) {$C E$ row spaces:\\$C E \cdot r (D
  - r)$};
\node[lbl, anchor=north] at (9.75,-1.9) {$C E$ maps:\\$C E \cdot r D$};

\node[ttl] at (10.9,2.35) {(c) Nested predictor};
\node[vec, minimum height=1.5cm] (mc) at (11.2,0) {};
\node[lbl] at (11.2,-1.0) {$\vm$};
\node[cell, minimum height=1.4cm, fill=\ca!28] (A1) at (12.6,1.0)
  {$\vA_{1}$};
\node[cell, minimum height=1.4cm, fill=\cb!28] (A2) at (12.6,-1.0)
  {$\vA_{2}$};
\draw[wire] (mc.east) -- (11.75,0) |- (A1.west);
\draw[wire] (11.75,0) |- (A2.west);
\foreach \e/\yy in {1/1.5, 2/1.0, 3/.5} {
  \node[cell] (nb\e) at (13.95,\yy) {$\vB_{1\e}$};
  \draw[wire] (A1.east |- nb\e) -- (nb\e.west);
}
\foreach \e/\yy in {1/-.5, 2/-1.0, 3/-1.5} {
  \node[cell] (ng\e) at (13.95,\yy) {$\vB_{2\e}$};
  \draw[wire] (A2.east |- ng\e) -- (ng\e.west);
}
\node[lbl] at (12.6,2.0) {read};
\node[lbl] at (13.95,2.0) {write};
\node[lbl, anchor=north] at (11.05,-1.9) {$C$ patterns:\\$C \cdot
  D^{2}$};
\node[lbl, anchor=north] at (12.75,-1.9) {$C$ row spaces:\\$C \cdot r (D -
  r)$};
\node[lbl, anchor=north] at (14.35,-1.9) {$C E$ maps:\\$C E \cdot r D$};
\end{tikzpicture}}
    \caption{\small \textbf{The linear model and the two predictors}, for
        \(C = 2\) coarse labels (blue, green) and \(E = 3\) fine labels.
        \textit{(a)} One sample: the coarse label determines the relevant
        tokens, the pattern \(\vW_{c}^{\star}\), and the row space of
        \(\vA_{c}^{\star}\); the fine label determines the map
        \(\vB_{ce}^{\star}\). \textit{(b)} The flat predictor (MoE
        transformer): the shared attention scores every token, and each
        expert has its own row space, fit from its own samples. \textit{(c)}
        The nested predictor (\ours{}): the attention of a column scores only
        the tokens with its label, and its row space is fit once from the
        samples of all of its experts.
    The free parameters are counted below each panel.}
    \label{fig:theory-model}
\end{figure}

\paragraph{Nested structure in the data.} We assume that the next token is
predicted in two stages. A coarse latent variable \(c \in [C]\) determines
which tokens of the context are relevant, how they are weighted, and which
statistic of them is computed. A fine latent variable \(e \in [E]\)
determines how this statistic is used. For example, in a corpus of code
and mathematics, \(c\) may be the kind of text, the relevant tokens may be
the declarations in scope or the hypotheses in effect, and \(e\) may
determine whether the next token of code is a variable or an operator. In
\ours{}, \(c\) corresponds to the output of the column router, the
statistic to the output of the attention operator of column \(c\), with its
own pattern and value map, and \(e\) to the output of the expert router of
the column. In the MoE transformer, one shared pattern weights the tokens
and one shared value map computes the statistic. We study the simplest
case: one layer with one head, linear
attention and maps, and the squared loss (\Cref{fig:theory-model}a).

\begin{model}{}{linear}
    Let \(r < D\) and \(S \geq 2\). For each \(c \in [C]\) and \(e \in
    [E]\), fix an invertible pattern \(\vW_{c}^{\star} \in \R^{D \times
    D}\) and matrices \(\vA_{c}^{\star} \in \R^{r \times D}\) and
    \(\vB_{ce}^{\star} \in \R^{D \times r}\) of rank \(r\). Each sample
    \((c, e, \vX, \vy)\) is generated in four steps.
    \begin{enumerate}[1., nosep, leftmargin=2em]
        \item The labels \((c, e)\) are drawn with \(\Pr[c, e] > 0\) for all
            \((c, e) \in [C] \times [E]\).
        \item The context \(\vX = (\vx_{1}, \dots, \vx_{S})\) consists
            of tokens \(\vx_{i} = (c_{i}, \vz_{i})\) for \(i < S\) and the
            query \(\vx_{S} = (c, \vz_{S})\). The labels \(c_{i}\) are
            uniform on \([C]\), and the contents \(\vz_{i} \in \R^{D}\) are
            drawn from a distribution with a density and finite moments,
            all independently of each other.
        \item The summary is \(\vm \doteq \sum_{i \in \cI}
            (\vz_{S}^{\top} \vW_{c}^{\star} \vz_{i})\, \vz_{i}\) over the
            relevant tokens \(\cI \doteq \{i < S \colon c_{i} = c\}\).
        \item The target is \(\vy \doteq \vB_{ce}^{\star} \vA_{c}^{\star}
            \vm + \veps\), with independent noise \(\veps\) of mean zero
            and covariance \(\sigma^{2} \vI\).
    \end{enumerate}
    We observe \(T\) independent samples \((c_{i}, e_{i}, \vX_{i},
    \vy_{i})_{i \in [T]}\).
\end{model}

The summary \(\vm\) is the output of one head of linear attention at the
query, over the \((S - 1) / C\) tokens with the label \(c\) on average. The
target \(\vy\) stands in for the next token, for example the update that
the layer adds to the residual stream. The regression function is
\(\vf^{\star}(c, e, \vX) = \vG_{ce}^{\star} \vm\), where
\begin{equation}\label{eq:target-structure}
    \vG_{ce}^{\star} \doteq \vB_{ce}^{\star} \vA_{c}^{\star},
    \qquad
    \rank \vG_{ce}^{\star} = r,
    \qquad
    \operatorname{row}(\vG_{ce}^{\star}) =
    \operatorname{row}(\vA_{c}^{\star}) \ \text{ for all } e \in [E].
\end{equation}
The last property is the nested structure: for a fixed coarse label \(c\),
the \(E\) maps \(\vG_{ce}^{\star}\) share the pattern \(\vW_{c}^{\star}\)
and the row space of \(\vA_{c}^{\star}\), while both of these may differ from
one coarse label to another.

\paragraph{The two predictors and their training.} Both predictors take the
labels as input along with the context, so no router is learned. The
\textit{nested predictor} is one layer of \ours{} with \(K_{\mathrm{c}} =
1\): the attention of column \(c\) sums over the tokens with the label
\(c\), with its own pattern \(\vW_{c}\) and a value map \(\vA_{c}\)
of width \(r\), and expert
\((c, e)\) applies a linear map \(\vB_{ce}\). The \textit{flat predictor} is
one layer of an MoE transformer: a shared attention operator sums over all
tokens, with a value map \(\vV\) of full width \(D\), as in our baseline, and
each expert is a linear MLP with \(r\) hidden units. Since a score
bilinear in the raw tokens cannot depend on both labels
(\Cref{thm:nested-informal}a), the shared score is bilinear in the product
features \(\ve_{c_{i}} \otimes \vz_{i}\) as in \Cref{rmk:simulation}, with
one block \(\vW_{cc'}\) per pair of labels. The two predictors are
\begin{equation}\label{eq:nested-predictor}
    \vf^{\mathrm{N}}(c, e, \vX) \doteq \vB_{ce} \vA_{c} \sum_{i \in \cI}
    (\vz_{S}^{\top} \vW_{c} \vz_{i})\, \vz_{i},
    \quad
    \vf^{\mathrm{F}}(c, e, \vX) \doteq \vW_{ce}^{\mathrm{out}}
    \vW_{ce}^{\mathrm{in}} \vV \sum_{i < S} (\vz_{S}^{\top} \vW_{c
    c_{i}} \vz_{i})\, \vz_{i}.
\end{equation}
We fit each predictor by least squares over its function class \(\cM\), and
measure its excess risk:
\begin{equation}\label{eq:least-squares}
    \hat{\vf} \in \argmin_{\vf \in \cM} \sum_{i \in [T]} \norm{\vy_{i} -
    \vf(c_{i}, e_{i}, \vX_{i})}_{2}^{2},
    \qquad
    \cE(\hat{\vf}) \doteq \Ex\big[\norm{\hat{\vf}(c, e, \vX) -
    \vf^{\star}(c, e, \vX)}_{2}^{2} \,\big|\, \hat{\vf}\big].
\end{equation}

\begin{theorem}{Informal version of \Cref{thm:nested}}{nested-informal}
    Let \(C \geq 2\).
    \begin{enumerate}[(a), nosep, leftmargin=2em]
        \item Both predictors can represent \(\vf^{\star}\), unlike a flat
            predictor with a bilinear score on raw tokens.
        \item As \(T \to \infty\), the excess risk of either predictor is
            asymptotically \(\sigma^{2} \chi^{2}_{d} / T\), where \(d\) is
            the number of directions in weight space that change the
            prediction. For \(r \ll D\) and \(E \gg 1\),
            \begin{equation}\label{eq:dims-informal}
                d_{\mathrm{N}} \approx C (D^{2} + E r D),
                \qquad
                d_{\mathrm{F}} \approx C (C D^{2} + 2 E r D),
                \qquad
                \frac{d_{\mathrm{F}}}{d_{\mathrm{N}}} \approx \frac{C D + 2
                E r}{D + E r} \in [2, C].
            \end{equation}
        \item If a sample costs \(F_{0}\) FLOPs plus \(F_{1}\) FLOPs per
            scored token, the flat predictor requires
            \begin{equation}\label{eq:compute-multiplier-informal}
                \frac{d_{\mathrm{F}}}{d_{\mathrm{N}}} \cdot \frac{S_{0} +
                S - 1}{S_{0} + (S - 1) / C}, \qquad S_{0} \doteq F_{0} /
                F_{1},
            \end{equation}
            times as many FLOPs for the same expected
            excess risk (in the large-data limit \(T \to \infty\)), a
            factor that grows in \(S\) from \(d_{\mathrm{F}} /
            d_{\mathrm{N}}\) to \(C
            d_{\mathrm{F}} / d_{\mathrm{N}}\).
    \end{enumerate}
\end{theorem}

The count \(d\) has three parts: the patterns, the row spaces, i.e., the
\(r\)-dimensional subspaces of \(\vm\) on which the maps depend, and the
maps from the row spaces to the output (\Cref{fig:theory-model}b,c). Both
predictors have \(C E\) maps to fit. The flat predictor also has one row
space per expert, fit from the samples of that expert, and \(C^{2}\)
pattern blocks, of which the \(C (C - 1)\) blocks with \(c' \neq c\) are
zero at the target and have to be fit as zero from data. The nested
predictor has one row space and one pattern per column, fit from the
samples of all experts of the column, and its routing fixes the other
blocks to zero. The ratio is near \(C\) when the
patterns dominate the parameter count and near \(2\) when the experts do.
With the routing, the nested predictor scores only \(1 / C\) of the
context, so part (c) is the multiplier \(R(S)\) of
\Cref{sec:long-context-extension} with \(\rho = 1 / C\), which grows with
\(S\) as in \Cref{fig:long-context-extension}. As in
\Cref{tab:attribution}, both parts of nesting are needed: with \(C = 1\)
the predictors coincide, and with column-routed attention and experts
\(\vW_{e}\) shared across columns, the column space of \(\vW_{e} \vA_{c}\)
lies in that of \(\vW_{e}\) for every \(c\), so this predictor cannot
represent a target where the column space of \(\vB_{ce}^{\star}\)
changes with \(c\).

\paragraph{Continual training and forgetting.} Each predictor starts from
weights that represent \(\vf^{\star}\) and is trained by gradient flow on
the population squared loss of a new domain \(Q\). Every sample of \(Q\)
has the coarse label \(c_{Q}\), e.g., a new corpus of code, and new
maps \(\vG'_{e} \neq \vG^{\star}_{c_{Q} e}\), and otherwise follows
\Cref{model:linear}. The forgetting on the old labels \((c, e)\) at time
\(t\) is the increase of their risk since time \(0\).

\begin{proposition}{Informal version of
    \Cref{prop:forgetting}}{forgetting-informal}
    In the nested predictor, the columns \(c \neq c_{Q}\) never change, so
    there is no forgetting on any other coarse label. In the flat predictor,
    \(\vf(c, e, \vX; t) = \vG_{ce}^{\star} \vV(t) \vm\) for \(c \neq
    c_{Q}\), and the forgetting on \((c, e)\) is
    \begin{equation}\label{eq:forgetting-informal}
        \frac{t^{2}}{2} \norm{\vB_{ce}^{\star} \vA_{c}^{\star}
        \vA_{c_{Q}}^{\star\top} \vXi\, \vSigma_{ce}^{1/2}}_{F}^{2} +
        O(t^{3}),
    \end{equation}
    where \(\vSigma_{ce} \doteq \Ex[\vm \vm^{\top} \mid c, e]\) and \(\vXi
    \neq \vzero\) whenever \(\dot{\vV}(0) \neq \vzero\). For orthonormal
    rows, the coefficient lies between multiples of \(\cos^{2} \theta_{r}\)
    and \(\cos^{2} \theta_{1}\), where \(\theta_{1} \leq \dots \leq
    \theta_{r}\) are principal angles between the row spaces of
    \(\vA_{c}^{\star}\) and \(\vA_{c_{Q}}^{\star}\).
\end{proposition}

In the flat predictor, the shared value map is the only source of
forgetting on other labels, and the forgetting grows with the overlap
between the row spaces of \(\vA_{c}^{\star}\) and
\(\vA_{c_{Q}}^{\star}\). This is consistent with
\Cref{fig:domain-adaptation}.\footnote{
    Here, the map \(\vV\) is in some sense
    redundant: freezing it would
    remove this forgetting
    without changing the function class.
    \Cref{prop:forgetting-informal} compares the
    predictors with attention and expert weights trained, as in
\Cref{sec:domain-adaptation}.} Both results depend on how the
relevant tokens, patterns, and row spaces vary across coarse labels.
Nesting improves sample and compute efficiency when these differ
between labels, and, with all weights trained, it removes forgetting
through the shared value map when the row spaces overlap,
which is the generic case.

\section{Conclusion}\label{sec:conclusion}

In this work, we introduced nested sparsity, a conditional computation
mechanism in which a router sends each token to several columns, and each
column contains its own attention operators and its own experts. The
resulting architecture, \ours{}, generalizes both the MoE transformer and the
dense transformer. In language model pretraining at scales from 100M to 1B
active parameters, the MoE transformer needs \(1.13\)--\(1.29\times\) the
compute of \ours{} to reach the same loss at an equal number of stored
parameters, and this factor shows no downward trend as the scale
grows (\Cref{sec:pretraining}). In
continual pretraining, \ours{} adapts to new domains as well as the MoE
transformer while forgetting less (\Cref{sec:domain-adaptation}), and
its advantage in long context
extension grows with the context length at equal FLOPs
(\Cref{sec:long-context-extension}). Our
theoretical analysis (\Cref{sec:theory}) provides
conceptual reasoning for why \ours{} can obtain such empirical
results in the context of highly structured data.

Our study has several limitations, which suggest directions for future work.
First, our largest model has 1B active parameters, and we evaluate models by
their held-out loss; an important next step is to test whether the
compute multiplier persists at larger scales and carries over to downstream
tasks, such as standard benchmarks after post-training. Second,
\ours{} is in some sense a meta-architecture, and we studied only the
case in which each column uses full causal attention. Columns with sparse or
linear attention would compound the savings at long contexts. More
broadly, columns comprised of subnetworks with interesting properties
could preserve them through the meta-architecture.

\printbibliography

\clearpage
\appendix
\begin{center}
    \noindent\rule{\textwidth}{4pt}\vspace{-0.2cm}

    \LARGE\textbf{Appendix}

    \noindent\rule{\textwidth}{1.2pt}
\end{center}

\section{Related Work}\label{app:related-work}

\paragraph{Scaling laws and sparsity.} The loss of a language model is well
described by a power law in the number of parameters, the number of training
tokens, and the training compute \citep{kaplan2020scaling}, which determines
the compute-optimal allocation between model size and data
\citep{hoffmann2022training}. Several works extend these laws to MoE
transformers. \citet{clark2022unified} fit a law that is jointly a function
of the model size and the number of experts. \citet{krajewski2024scaling}
add the granularity of the experts as a variable, and find that MoE
transformers with appropriately sized experts outperform dense transformers
at every compute budget. \citet{abnar2025parameters} study the sparsity
level, i.e., the fraction of inactive parameters, and find that the optimal
sparsity increases with the model size, and \citet{ludziejewski2025joint}
fit a joint law for dense and MoE models under memory constraints. In all of
these works only the MLP is sparse. Since our architectures are sparse in
more than one component, we compare them at the \textit{same} sparsity to
avoid ambiguity in the scaling laws: we fit a scaling
law for each architecture and report
the projected compute efficiency compared to a strong MoE transformer
baseline to reach the same held-out loss.

\paragraph{Conditional computation and mixtures of experts.} Mixtures of
experts were introduced by \citet{jacobs1991adaptive}, and
\citet{jordan1994hierarchical} extended them to a tree of gating networks.
\citet{shazeer2017outrageously} used them as a sparsely gated layer in
deep networks, following the proposal of conditional computation
\citep{bengio2013estimating}, and later work used them in the
transformer and changed the routing \citep{lepikhin2021gshard,
fedus2022switch, zhou2022mixture}. Recent MoE transformers use many small
experts, some of which are shared by all tokens \citep{dai2024deepseekmoe},
and balance the load of the experts with a bias in the router
\citep{wang2024auxiliary, liu2024deepseekv3}; we do the same. Nested
sparsity is closest to the hierarchical mixture of experts, since in \ours{}
the experts of a token are selected from the experts of its columns, which
are in turn selected by the column router. The difference is that in a
hierarchical mixture of experts both levels of routing select MLPs, whereas
in \ours{} the first level selects an attention operator together with its
context, and this selection is held fixed over several layers. Other works
route tokens around whole layers: in \citet{raposo2024mixture} only the
top-\(K\) tokens of each layer go through its attention operator and MLP,
and \citet{ainslie2023colt5} apply a heavy attention operator and MLP only
to the routed tokens and a light one to all tokens. In these methods there
is a single routed module per layer, and a token that is not selected does
not go through it; in \ours{}, every layer applies to every token, with
modules that depend on the token.

\paragraph{Sparse attention and mixtures of attention heads.} A large body
of work reduces the quadratic cost of attention by making the attention
map sparse. The pattern can be fixed in advance, such as local or strided
windows \citep{child2019generating, beltagy2020longformer}; it can depend on
the content through hashing or clustering of the queries and keys
\citep{kitaev2020reformer, roy2021efficient}; or it can be selected for each
query by a top-\(K\) rule over blocks or tokens \citep{yuan2025native,
lu2025moba, deepseek2025v32}. A complementary line of work makes the
attention parameters sparse, by routing each token to a subset of the
attention heads or of their projections \citep{zhang2022mixture,
csordas2024switchhead, jin2025moh}, possibly together with an MoE operator
for the MLP \citep{shen2023moduleformer}. The method of
\citet{piekos2025mixture}, in which a fixed number of tokens is selected for
each head and attention is computed only among them, is the closest to
column-routed attention. We compare \ours{} with these methods in detail in
\Cref{sec:architecture}. In column-routed attention, both the pattern and
the parameters are sparse, and both are given by one routing decision, which
is computed for each token, held fixed over a block of layers, and shared
with the experts. Also, in our experiments \ours{} has lower loss than
baselines with full attention, which sparse attention methods approximate.

\paragraph{Interleaved attention.} In many LLMs, the cost of attention is
reduced by using full attention in only some of the layers. GPT-3 has dense
and locally banded sparse attention layers in alternation
\citep{brown2020language}, and more recent models interleave sliding window
and global attention layers at ratios from one to one up to five to one
\citep{gemma2024gemma2, gemma2025gemma3, cohere2025commanda, openai2025gptoss}.
In hybrid models, the local layers are state space or linear attention
layers \citep{lieber2024jamba, minimax2025minimax01, kimi2025linear}. \ours{}
uses the same principle with a different local operator: the number of
active columns increases with depth, so attention in the early blocks is
computed within routed subsequences, and attention in the last block is full
attention over the whole sequence (\Cref{sec:architecture}). Unlike a sliding
window, the context of a column depends on the input and can contain tokens
from anywhere in the sequence.

\paragraph{Continual pretraining and forgetting.} Continued pretraining on
text from a target domain improves performance in that domain
\citep{gururangan2020dont}, and it is the standard way to obtain specialized
models for code and mathematics \citep{roziere2023code, azerbayev2023llemma,
shao2024deepseekmath}. Its main difficulty is catastrophic forgetting
\citep{mccloskey1989catastrophic, french1999catastrophic,
kirkpatrick2017overcoming, luo2025empirical}, which is usually reduced by
replaying pretraining data \citep{ibrahim2024simple}. \citet{french1991using}
observed that forgetting comes from the overlap of distributed
representations, and that sparser representations reduce it. For this
reason, many continual learning methods use separate parameters for each
task, obtained by adding, freezing, or masking parts of the network
\citep{rusu2016progressive,
    fernando2017pathnet, aljundi2017expert, mallya2018packnet,
wortsman2020supermasks}. For language models, modularity is typically obtained
with experts: \citet{gururangan2022demix} condition the experts on the domain
and adapt to a new domain by training a new expert with the rest of the
model frozen, \citet{chen2023lifelong} add and freeze experts as the
distribution changes, \citet{sukhbaatar2024branch} train domain models
separately and merge their MLPs into the experts of an MoE transformer, and
\citet{lin2025continual} update a sparse set of memory slots.
\citet{shen2023moduleformer} route tokens among both attention and MLP
modules, and find that their model has less forgetting than a dense
model and that
new modules can be added to it. \citet{therien2025continual} find that the
routing of standard MoE transformers changes little under the distribution
shifts of continual pretraining, and \citet{li2025theory} analyze MoE models
in continual learning theoretically. When an MoE transformer is continually
pretrained with trainable attention parameters, these parameters are
shared by all tokens, so the gradient of every new token changes them. In
\ours{}, each layer applies to a token only the attention operators and
experts of its columns, in addition to the shared experts, with no task
labels and no changes to training. In the one-layer model of
\Cref{sec:theory}, where
each token uses one column, there is then no forgetting on the other coarse
labels. Our analysis of forgetting in \Cref{sec:theory} uses
the overlap between the row spaces of the value maps of the old and the new
data, as in \citet{doan2021theoretical}.

\section{Mathematical Description of the
Architecture}\label{app:architecture}

This section specifies the forward pass of \ours{} as trained in all
experiments. It expands \eqref{eq:col-attn} and \eqref{eq:nested-moe}
with the normalization, the residual connections, and the routing details
that \Cref{sec:architecture} omits. The initialization and the learning
rate of each parameter are in
\Cref{app:experimental-parameterization}, and the sizes of
each model are in \Cref{app:configurations}.

\paragraph{Notation and overall structure.} The input is a sequence of \(S\)
tokens, and all features have width \(D\). Let \(\rms(\vv) \doteq
    (\norm{\vv}_{2}^{2}
/ D + \epsilon)^{1/2}\), and let \(\operatorname{Norm}(\vv) \doteq \vgamma
\odot \vv / \rms(\vv)\) be the RMS normalization with a learned gain
\(\vgamma \in \R^{D}\); each use of \(\operatorname{Norm}\) below has its own
gain. No linear map has a bias. The embedding of the token \(j \in [S]\) is
\(\ve_{j} \in \R^{D}\), with no learned position embedding, and the network
then applies \(B\) blocks. Block \(\ell \in [B]\) has \(M_{\ell}\)
layers, \(C\) columns,
and \(K_{\mathrm{c}, \ell}\) active columns per token. Below we describe one
block and omit the index \(\ell\) from its parameters and its routing
variables. Each block has one output \(\bar{\vu}_{j}^{\ell} \in \R^{D}\)
for each token, which we define in \eqref{eq:app-export}. The input of block
\(\ell\) is \(\vx_{j} \doteq \bar{\vu}_{j}^{\ell - 1}\), with
\(\bar{\vu}_{j}^{0} \doteq \ve_{j}\).

\paragraph{Column router.} The router is computed once per block from the
block input. Its scores are \(\vs_{j} \in \R^{C}\) with \(\vs_{j}
    \doteq \vW_{2}\,
\operatorname{GELU}(\vW_{1} \vx_{j})\), with a hidden layer of
width \(128\). During training we add input-dependent noise
\citep{shazeer2017outrageously}, i.e., we replace \(\vs_{j}\) by \(\vs_{j} +
    \vxi_{j} \odot \operatorname{softplus}(\vW_{2}'\,
\operatorname{GELU}(\vW_{1}' \vx_{j}))\) with \(\vxi_{j} \sim \dNorm(\vzero,
\vI)\); there is no noise at evaluation. The selected columns and their gating
weights are
\begin{equation}\label{eq:app-col-router}
    \cC_{j} \doteq \TopK(\vs_{j} + \vb,\, K_{\mathrm{c}}) \subseteq [C],
    \qquad
    w_{jc} \doteq \frac{\exp(s_{jc})}{\sum_{c' \in \cC_{j}} \exp(s_{jc'})}
    \quad \text{for } c \in \cC_{j},
\end{equation}
and \(\cU_{c} \doteq \{i \in [S] \colon c \in \cC_{i}\}\) is the set of tokens
routed to column \(c\). The bias \(\vb \in \R^{C}\) is used only for the
selection: the gating weights do not contain it, and it has no gradient.
After each optimizer step we update the bias by \citep{wang2024auxiliary}
\begin{equation}\label{eq:app-bias}
    b_{c} \leftarrow b_{c} + \gamma \operatorname{sign}\Big(\frac{1}{C}
    \sum_{c' \in [C]} n_{c'} - n_{c}\Big),
\end{equation}
where \(n_{c}\) is the number of tokens in the batch which selected the column
\(c\), and \(\gamma = 10^{-3}\). By this update, the bias increases for the
columns with less than the average load. We use no auxiliary
balancing loss. The sets \(\cC_{j}\)
and the weights \(w_{jc}\) are used by all \(M\) layers of the block. In the
last block \(K_{\mathrm{c}} = C\), so that \(\cC_{j} = [C]\), \(\cU_{c} = [S]\),
and the router gives only the weights \(w_{jc}\).

\paragraph{Attention residual.} We replace all residual connections by the
following operator, which is applied to each token separately. For a list of
\textit{sources} \(\vz_{1}, \dots, \vz_{N} \in \R^{D}\), its output is
the mixture
\begin{equation}\label{eq:app-attnres}
    \vAttnRes(\vz_{1}, \dots, \vz_{N}) \doteq \sum_{i = 1}^{N}
    \frac{\exp(\ip{\vq}{\vz_{i}} / (\sqrt{D}\,
    \rms(\vz_{i})))}{\sum_{i' = 1}^{N} \exp(\ip{\vq}{\vz_{i'}} /
    (\sqrt{D}\, \rms(\vz_{i'})))}\,
    g_{i}\, \frac{\vz_{i}}{\rms(\vz_{i})},
\end{equation}
with a learned query \(\vq \in \R^{D}\) and learned gains \(g_{1}, \dots, g_{N}
\in \R\). Each use of \(\vAttnRes\) below has its own query and gains. This is
the attention residual of \citet{kimi2026attnres} with two changes,
which make the
scales of the mixture and of its logits independent of the width: in the
mixture, the sources are normalized to unit RMS, as in the logits, and
multiplied by the gains \(g_{i}\); and the logits are divided by
\(\sqrt{D}\). The query does not depend on the token, so the mixture weights
depend on the input only through the sources. We initialize \(\vq = \vzero\), so
that the mixture is initially uniform.

\paragraph{One layer.} Within a block, each token \(j\) has one residual
stream \(\vu_{jc}^{m} \in \R^{D}\) for each active column \(c \in \cC_{j}\),
where \(m = 0, \tfrac{1}{2}, 1, \dots, M\) counts the attention and MLP
operators applied so far, and \(\vu_{jc}^{0} \doteq \vzero\). Let
\(\bar{\vu}_{j}^{m} \doteq \frac{1}{K_{\mathrm{c}}} \sum_{c \in \cC_{j}}
\vu_{jc}^{m}\) be the mean of the streams. For each \(c \in
\cC_{j}\), layer \(m \in [M]\) is
\begin{align}
    \vh_{j} &\doteq \operatorname{Norm}(\vAttnRes(\ve_{j},
            \bar{\vu}_{j}^{1}, \dots, \bar{\vu}_{j}^{\ell - 1},
    \bar{\vu}_{j}^{m - 1})),
    \label{eq:app-read-attn} \\
    \vu_{jc}^{m - 1/2} &\doteq \vu_{jc}^{m - 1} + w_{jc}\,
    \vAttn_{c}(\vh_{j},\, (\vh_{i})_{i \in \cU_{c},\, i \leq j}),
    \label{eq:app-attn} \\
    \vh_{j}' &\doteq \operatorname{Norm}(\vAttnRes(\ve_{j},
            \bar{\vu}_{j}^{1}, \dots, \bar{\vu}_{j}^{\ell - 1},
    \bar{\vu}_{j}^{m - 1/2})),
    \label{eq:app-read-mlp} \\
    \vu_{jc}^{m} &\doteq \vu_{jc}^{m - 1/2}
    + \lambda_{j}^{\mathrm{s}} \sum_{e \in \cE_{j}^{\mathrm{s}}}
    \alpha_{je}^{\mathrm{s}}\, \vMLP_{e}^{\mathrm{s}}(\vh_{j}')
    + \lambda_{j}^{\mathrm{p}}\, w_{jc} \sum_{e \in \cE_{jc}} \alpha_{jce}\,
    \vMLP_{c, e}(\vh_{j}').
    \label{eq:app-mlp}
\end{align}
In \eqref{eq:app-read-attn} and \eqref{eq:app-read-mlp}, the sources are the
embedding, the outputs of the earlier blocks (here the superscripts \(1,
\dots, \ell - 1\) are block indices), and the current mean of the streams of
this block. We omit the last source in \eqref{eq:app-read-attn} when \(m =
1\), since it is zero. The outputs of the attention operator of column \(c\)
and of the experts \(\vMLP_{c, e}\) of column \(c\) are added only to the
stream of column \(c\), and the outputs of the shared experts
\(\vMLP_{e}^{\mathrm{s}}\) are added to every stream. All columns have the
same inputs \(\vh_{j}\) and \(\vh_{j}'\), which depend on the streams only
through their mean. The output of the block is the
final mean,
\begin{equation}\label{eq:app-export}
    \bar{\vu}_{j}^{\ell} \doteq \bar{\vu}_{j}^{M_{\ell}} =
    \frac{1}{K_{\mathrm{c}}}
    \sum_{c \in \cC_{j}} \vu_{jc}^{M_{\ell}}.
\end{equation}
By \eqref{eq:app-attn} and \eqref{eq:app-mlp}, this mean is the sum of the
updates in \eqref{eq:col-attn} and \eqref{eq:nested-moe} over the layers of
the block, with a factor \(1 / K_{\mathrm{c}}\) on the terms of the columns
and a factor \(\lambda_{j}^{\mathrm{s}}\) or \(\lambda_{j}^{\mathrm{p}}\) on
the experts.

\paragraph{Attention in a column.} Each column has its own projections
\(\vW_{Q}, \vW_{K}, \vW_{V} \in \R^{H_{\mathrm{c}} D_{\mathrm{h}} \times D}\)
and \(\vW_{O} \in \R^{D \times H_{\mathrm{c}} D_{\mathrm{h}}}\) in each layer,
where \(H_{\mathrm{c}}\) is the number of heads of one column and
\(D_{\mathrm{h}}\) is the width of a head. In the head \(\eta \in
[H_{\mathrm{c}}]\), the token \(j \in \cU_{c}\) has the query, key, and value
\(\vq_{j}^{\eta}, \vk_{j}^{\eta}, \vv_{j}^{\eta} \in
\R^{D_{\mathrm{h}}}\), which are the
corresponding slices of \(\vW_{Q} \vh_{j}\), \(\vW_{K} \vh_{j}\), and \(\vW_{V}
\vh_{j}\); we apply rotary position embeddings \citep{su2024roformer} to the
queries and the keys, using the position \(j\) of the token in the whole
sequence and not its position within \(\cU_{c}\). The output of the head is
\begin{equation}\label{eq:app-head}
    \vo_{j}^{\eta} \doteq \sum_{i \in \cU_{c},\, i \leq j}
    \frac{\exp(\ip{\vq_{j}^{\eta}}{\vk_{i}^{\eta}} /
    D_{\mathrm{h}})}{\sum_{i' \in \cU_{c},\, i' \leq j}
    \exp(\ip{\vq_{j}^{\eta}}{\vk_{i'}^{\eta}} / D_{\mathrm{h}})}\,
    \vv_{i}^{\eta},
\end{equation}
and \(\vAttn_{c}(\vh_{j}, (\vh_{i})_{i \in \cU_{c}, i \leq j}) \doteq \vW_{O}
(\vo_{j}^{1}, \dots, \vo_{j}^{H_{\mathrm{c}}})\). The division by
\(D_{\mathrm{h}}\) instead of \(\sqrt{D_{\mathrm{h}}}\) follows \muP{}
\citep{yang2022tensorprogramsv}. In the implementation, we gather the
\(\abs{\cU_{c}}\) tokens routed to column \(c\), apply a standard causal
attention kernel to this shorter sequence, and scatter the result back to
the positions of these tokens. To compute the
saving, give each column \(H / K_{\mathrm{c}}\) heads, so that all
projections cost the same FLOPs as in a transformer with \(H\) heads. If the
routing is balanced, i.e., \(\abs{\cU_{c}} \approx S K_{\mathrm{c}} / C\),
then the quadratic term in the attention FLOPs decreases from \(S^{2} D\) to
\begin{equation}\label{eq:attn-cost}
    C \cdot \Big(\frac{S K_{\mathrm{c}}}{C}\Big)^{2} \cdot
    \frac{D}{K_{\mathrm{c}}}
    = \frac{K_{\mathrm{c}}}{C} \cdot S^{2} D.
\end{equation}
The saving \((1 - K_{\mathrm{c}} / C) S^{2} D\) grows with the context
length \(S\).

\paragraph{Experts.} Every expert is an MLP with gated linear units,
\(\vMLP(\vh) \doteq \vW_{\mathrm{down}}
    (\operatorname{SiLU}(\vW_{\mathrm{gate}} \vh) \odot \vW_{\mathrm{up}}
\vh)\) \citep{shazeer2020glu}. Each expert router is a linear map of
\(\vh_{j}'\) with its own balancing bias, updated by
\eqref{eq:app-bias}. Each layer has
\(E_{\mathrm{s}}\) shared experts with one router \(\vr^{\mathrm{s}}\), and
each column \(c\) has \(E\) experts with its own router \(\vr_{c}\):
\begin{align}
    \cE_{j}^{\mathrm{s}} &\doteq \TopK(\vr^{\mathrm{s}}(\vh_{j}') +
    \vb^{\mathrm{s}},\, K_{\mathrm{s}}) \subseteq [E_{\mathrm{s}}],
    &
    \alpha_{je}^{\mathrm{s}} &\doteq
    \frac{\exp(r_{e}^{\mathrm{s}}(\vh_{j}'))}{\sum_{e' \in
    \cE_{j}^{\mathrm{s}}} \exp(r_{e'}^{\mathrm{s}}(\vh_{j}'))},
    \label{eq:app-shared-router} \\
    \cE_{jc} &\doteq \TopK(\vr_{c}(\vh_{j}') + \vb_{c},\, K_{\mathrm{e}})
    \subseteq [E],
    &
    \alpha_{jce} &\doteq \frac{\exp(r_{c, e}(\vh_{j}'))}{\sum_{e' \in [E]}
    \exp(r_{c, e'}(\vh_{j}'))}.
    \label{eq:app-private-router}
\end{align}
The weights of the shared experts are normalized over the selected experts, as
in \eqref{eq:app-col-router}. The weights \(\alpha_{jce}\) are normalized over
all \(E\) experts of the column \citep{fedus2022switch}, so that they
depend on the router also when \(K_{\mathrm{e}} = 1\). The
load \(n_{e}\) in the update \eqref{eq:app-bias} of \(\vb_{c}\) counts only
the tokens in \(\cU_{c}\). The two families of experts are weighted by a
gate with the same input,
\begin{equation}\label{eq:app-path-gate}
    (\lambda_{j}^{\mathrm{s}}, \lambda_{j}^{\mathrm{p}}) \doteq
    \operatorname{softmax}(\vW_{\lambda} \vh_{j}'),
    \qquad \vW_{\lambda} \in \R^{2 \times D}.
\end{equation}
In all experiments \(E_{\mathrm{s}} = 8\) and \(K_{\mathrm{s}} = 2\). The
hidden widths of the experts are \(2 D / K_{\mathrm{s}}\) for a shared expert
and \(2 D / (K_{\mathrm{c}} K_{\mathrm{e}})\) for an expert of a column,
rounded to a multiple of \(8\), so that each token uses hidden units of total
width \(K_{\mathrm{s}} \cdot 2 D / K_{\mathrm{s}} + K_{\mathrm{c}}
K_{\mathrm{e}} \cdot 2 D / (K_{\mathrm{c}} K_{\mathrm{e}}) = 4 D\) in every
block. We use \(K_{\mathrm{e}} = 2\) in the blocks where the resulting width
is at least \(160\), and \(K_{\mathrm{e}} = 1\) in the other blocks.

\paragraph{Output.} The input of the output head is an attention residual
with sources the embedding, the output of every block, and the deviation of
every stream from the mean of its block:
\begin{equation}\label{eq:app-head-read}
    \vx_{j}^{\mathrm{out}} \doteq
    \operatorname{Norm}(\vAttnRes(\ve_{j},\,
            (\bar{\vu}_{j}^{\ell})_{\ell \in [B]},\,
            (\vu_{jc}^{\ell} -
    \bar{\vu}_{j}^{\ell})_{\ell \in [B],\, c \in \cC_{j}^{\ell}})),
\end{equation}
where \(\vu_{jc}^{\ell}\) and \(\cC_{j}^{\ell}\) are the final stream
\(\vu_{jc}^{M_{\ell}}\) and the selected columns of block \(\ell\). The logits
of the next token are a linear map of \(\vx_{j}^{\mathrm{out}}\), whose
weights are not tied to the embedding, and the training loss is the
cross-entropy of the next token, with no auxiliary term. This is the
only place where the
individual streams are used, since the layers in
\eqref{eq:app-read-attn} and \eqref{eq:app-read-mlp} use only the
means. We use the deviations
because after the normalization in \eqref{eq:app-attnres} the streams of a
block are close to their mean, so using the streams themselves would add
nearly the same source several times. The deviations sum to zero over
\(c \in \cC_{j}^{\ell}\), so they contain only the part of each stream that
is specific to its column. Their gains in \eqref{eq:app-attnres} are
initialized at \(0\), and
all other gains at \(1\), so that at initialization the output does not
depend on the deviations.

\paragraph{Special cases.} Let \(C = K_{\mathrm{c}} = 1\). Then \(w_{j1} = 1\)
and \(\cU_{1} = [S]\), each block has one stream, the deviations in
\eqref{eq:app-head-read} are zero, and
\eqref{eq:app-read-attn}--\eqref{eq:app-mlp}
are \(M\) layers of an MoE transformer with full causal attention and the
block attention residual of \citet{kimi2026attnres}, with the two changes
stated after \eqref{eq:app-attnres}. Our MoE baseline is this case with a
single family of \(E = 24\) experts, \(K_{\mathrm{e}} = 3\), the
gating weights of
\eqref{eq:app-shared-router}, and no gate \eqref{eq:app-path-gate}. The dense
baseline further sets \(E = K_{\mathrm{e}} = 1\), with a single SwiGLU of
hidden width \(4D\) per layer and gating weight \(1\). Both baselines have
the same block attention residual as above.

\section{Further Experiments and Ablations}\label{app:experiments}

\subsection{Architecture Configurations and Hyperparameters}
\label{app:configurations}

This subsection gives the configuration of \ours{} and of the two baselines
at the four scales of \Cref{sec:pretraining}, the rules that set every
quantity other than the number of active columns, the experiment by which
we selected the schedule of active columns over depth, and the FLOP accounting
used throughout the paper.

\paragraph{Sizes.} The four scales are listed in
\Cref{tab:config-sizes}. The dense
transformer and the MoE transformer share the width \(D\), the depth,
\(16\) heads of width \(D / 16\), and the training budget; in the MoE
transformer, each MLP is replaced by \(E = 24\) experts of which
\(K_{\mathrm{e}} = 3\) are active. \ours{} has \(C = 8\) columns in every
block and the same schedule of active columns at every scale,
\(K_{\mathrm{c}} = 3, 4, 5, 6, 6, 7\) in the first six blocks and
\(K_{\mathrm{c}} = C\) in the last. The number of layers per block grows
with the scale, and the number of experts per column \(E\) is raised at
500M and 1B so that the stored parameters match those of the MoE
transformer. The batch grows with the scale so that every model is trained for
about the same number of steps.

\begin{table}[ht]
    \centering
    \small
    \setlength{\tabcolsep}{3.5pt}
    \begin{tabular}{lrrrrrlrcc}
        \toprule
        Scale & \(D\) & Depth & \(D / 16\) & Tokens per step & Steps &
        \ours{}: layers per block & \(E\) & FLOPs & Stored \\
        \midrule
        100M & 480 & 12 & 30 & 262{,}144 & 19{,}080 & 1, 1, 2, 2, 2,
        3, 2 (13) & 8 & \(0.986\times\) & 0.38B \\
        200M & 640 & 16 & 40 & 524{,}288 & 19{,}080 & 2, 2, 2, 3, 3,
        3, 2 (17) & 8 & \(0.981\times\) & 0.76B \\
        500M & 960 & 24 & 60 & 1{,}048{,}576 & 23{,}844 & 3, 3, 3, 4,
        4, 5, 3 (25) & 12 & \(0.979\times\) & 2.44B \\
        1B & 1280 & 32 & 80 & 2{,}097{,}152 & 23{,}842 & 3, 4, 4, 6,
        6, 6, 4 (33) & 15 & \(0.989\times\) & 5.53B \\
        \bottomrule
    \end{tabular}
    \caption{The four scales. Width, depth, head width, and training budget
        are those of the dense transformer and of the MoE transformer; the
        sequence length is \(4096\). For \ours{}, the layers of the seven blocks
        (total in parentheses) at the schedule \(K_{\mathrm{c}} = 3, 4, 5, 6,
        6, 7, 8\), the number of experts per column \(E\), its forward FLOPs per
        token at \(S = 4096\) relative to the dense transformer, and its stored
    parameters (the MoE transformer has 0.37B, 0.79B, 2.41B, and 5.49B).}
    \label{tab:config-sizes}
\end{table}

\paragraph{Derived quantities.} Only \(C\) and the schedule of
\(K_{\mathrm{c}}\) are design choices; every other quantity that affects
the FLOPs is set by a rule, so two configurations with different
schedules differ in nothing else. Each column has \(H_{\mathrm{c}} =
\operatorname{round}(16 / K_{\mathrm{c}})\) heads of the baselines' head
width, i.e., \(5, 4, 3, 3, 3, 2, 2\) heads in the seven blocks. The
FLOPs of the attention projections of a token are then within \(10\%\) of
those of the baselines, and a sparse block uses fewer FLOPs only in the
products of queries with keys and of attention weights with values
\eqref{eq:attn-cost}. The expert widths follow \Cref{app:architecture}:
the shared experts and the experts of the columns of a token have
total hidden width \(4D\) in every block, the active MLP width of the
baselines, and \(E\) determines the number of stored parameters without
changing the active width. With heads and widths fixed, the FLOPs of a
layer depend only on the
\(K_{\mathrm{c}}\) of its block, and we choose the number of layers in
each block to meet the budget: for a given schedule we enumerate the
splits of layers over blocks and keep those whose forward FLOPs at
\(S = 4096\), with the accounting below, do not exceed the dense
transformer's. Among the splits within budget at 200M, we selected
the one in \Cref{tab:config-sizes}, which has most of its layers in the
later blocks where \(K_{\mathrm{c}}\) is large. The other
configurations of the search with the same budget (a sparser or
    denser first block, a
    lower endpoint of the schedule, more layers in the sparse blocks, wider
attention or wider experts instead of more layers) have losses higher by
\(0.004\) to \(0.033\) nats. At the other scales we used splits of
the same form without further search. The resulting models use
\(1\)--\(2\%\) fewer FLOPs than the budget.

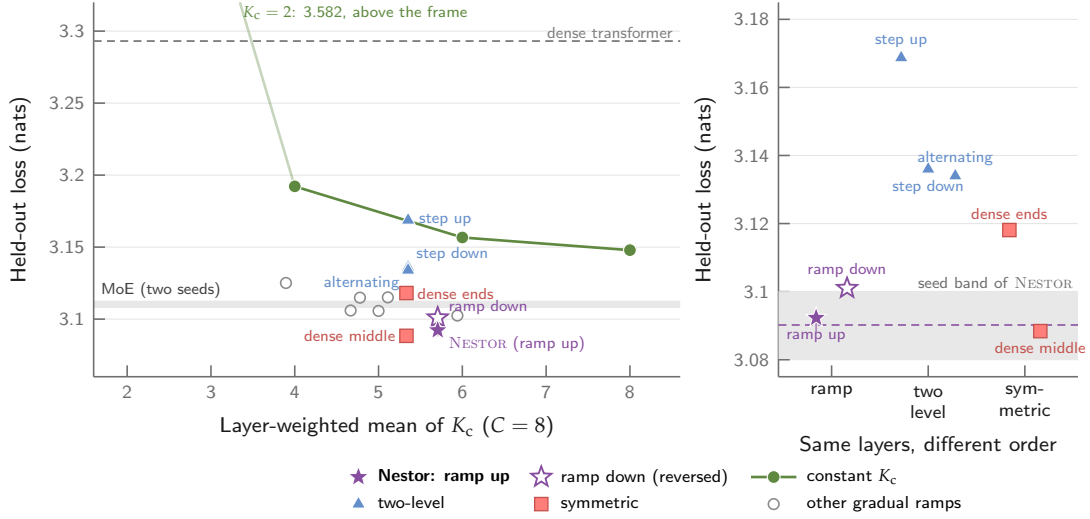
\begin{figure}[ht]
    \centering
    \pgfplotstableread[col sep=comma]{figures/exps/raw_data/derived/ksched_refs.csv}\ksrefs
\pgfplotstableread[col sep=comma]{figures/exps/raw_data/derived/ksched_points.csv}\kstable
\def\kspoints{figures/exps/raw_data/derived/ksched_points.csv}
\pgfplotstablegetelem{0}{dense}\of\ksrefs\let\ksdense\pgfplotsretval
\pgfplotstablegetelem{0}{moelo}\of\ksrefs\let\ksmoelo\pgfplotsretval
\pgfplotstablegetelem{0}{moehi}\of\ksrefs\let\ksmoehi\pgfplotsretval
\pgfplotstablegetelem{0}{band}\of\ksrefs\let\ksband\pgfplotsretval
\pgfplotstablegetelem{0}{halfband}\of\ksrefs\let\kshalf\pgfplotsretval
\pgfplotstablegetelem{0}{unitwo}\of\ksrefs\let\ksunitwo\pgfplotsretval
\pgfplotstablegetelem{0}{unifour}\of\ksrefs\let\ksunifour\pgfplotsretval
\pgfplotstablegetelem{0}{clipx}\of\ksrefs\let\ksclipx\pgfplotsretval
\pgfplotstablegetelem{0}{ylo}\of\ksrefs\let\ksylo\pgfplotsretval
\pgfplotstablegetelem{0}{yhi}\of\ksrefs\let\ksyhi\pgfplotsretval
\pgfplotstablegetelem{0}{oylo}\of\ksrefs\let\ksoylo\pgfplotsretval
\pgfplotstablegetelem{0}{oyhi}\of\ksrefs\let\ksoyhi\pgfplotsretval
\pgfplotstablegetelem{0}{mean_k}\of\kstable\let\kspaperx\pgfplotsretval
\pgfplotstablegetelem{0}{ce}\of\kstable\let\kspapery\pgfplotsretval
\pgfplotstablegetelem{0}{order_x}\of\kstable\let\kspaperox\pgfplotsretval
\pgfplotsset{
  ks paper mark/.style={mark=fstar, mark size=4pt, mark options={fill=expCortex, draw=white, line width=0.4pt}},
  ks mirror mark/.style={mark=fstar, mark size=4pt, mark options={fill=white, draw=expCortex, line width=0.7pt}},
  ks uniform line/.style={expSchedUniform, line width=1pt},
  ks uniform mark/.style={mark=*, mark size=2.4pt, mark options={fill=expSchedUniform, draw=white, line width=0.4pt}},
  ks step mark/.style={mark=triangle*, mark size=3pt, mark options={fill=expSchedStep, draw=white, line width=0.4pt}},
  ks sym mark/.style={mark=square*, mark size=2.5pt, mark options={fill=expSchedSym, draw=expSchedSymEdge, line width=0.5pt}},
  ks context mark/.style={mark=o, mark size=2pt, mark options={draw=black!45, line width=0.6pt}},
}
\tikzset{ks label/.style={font=\sffamily\fontsize{6}{7}\selectfont, inner sep=1pt}}
\newcommand{\kspts}[3]{\addplot[only marks, #2, forget plot] table[col sep=comma, x=#3, y=ce, discard if not={family}{#1}] {\kspoints};}
\newcommand{\kswhisk}[3]{\addplot[#2, forget plot, mark=none, draw=none,
    error bars/.cd, y dir=both, y explicit, error mark=none, error bar style={#2, solid, opacity=0.6, line width=0.7pt}]
  table[col sep=comma, x=#3, y=ce, y error plus=eplus, y error minus=eminus, discard if not={family}{#1}] {\kspoints};}
\newcommand{\kslab}[5]{\addplot[only marks, mark=none, forget plot, nodes near coords, point meta=explicit symbolic,
    every node near coord/.style={ks label, #5}]
  table[col sep=comma, x=#4, y=ce, meta=label, discard if not={family}{#1}, discard if not={#2}{#3}] {\kspoints};}
\begin{tikzpicture}
\begin{groupplot}[exps,
  group style={group size=2 by 1, horizontal sep=1.3cm},
  scale only axis, height=0.3\textwidth,
  ylabel={Held-out loss (nats)}, ytick distance=0.05,
]
\nextgroupplot[width=0.47\textwidth, xmin=1.6, xmax=8.6, ymin=\ksylo, ymax=\ksyhi, xtick={2,3,4,5,6,7,8},
  xlabel={Layer-weighted mean of $K_{\mathrm{c}}$ ($C = 8$)}, clip=false,
  legend to name=exps:ksched, legend columns=3,
  legend style={font=\sffamily\fontsize{6.5}{8}\selectfont, /tikz/every even column/.append style={column sep=0.6em},
                /tikz/every odd column/.append style={column sep=0.15em}}]
  \expsnow{\noexpand\addplot[expDense, densely dashed, line width=0.7pt, forget plot] coordinates {(1.6,\ksdense) (8.6,\ksdense)};}
  \expsnow{\noexpand\node[ks label, text=expDense, anchor=south east] at (axis cs:8.55,\ksdense) {dense transformer};}
  \expsnow{\noexpand\addplot[draw=none, fill=expMoe, fill opacity=0.12, forget plot] coordinates {(1.6,\ksmoelo) (8.6,\ksmoelo) (8.6,\ksmoehi) (1.6,\ksmoehi)} \noexpand\closedcycle;}
  \expsnow{\noexpand\node[ks label, text=expMoe, anchor=south west] at (axis cs:1.65,\ksmoehi) {MoE (two seeds)};}
  \addplot[ks uniform line, forget plot] table[col sep=comma, x=mean_k, y=ce, discard if not={family}{uniform}, discard if={arm}{hyb_uni2}] {\kspoints};
  \expsnow{\noexpand\addplot[ks uniform line, opacity=0.35, forget plot] coordinates {(4,\ksunifour) (\ksclipx,\ksyhi)};}
  \expsnow{\noexpand\node[ks label, text=expSchedUniform, anchor=north west] at (axis cs:\ksclipx,\ksyhi) {$K_{\noexpand\mathrm{c}} = 2$: \ksunitwo, above the frame};}
  \kspts{context}{ks context mark}{mean_k}
  \addplot[only marks, ks uniform mark, forget plot] table[col sep=comma, x=mean_k, y=ce, discard if not={family}{uniform}, discard if={arm}{hyb_uni2}] {\kspoints};
  \kspts{twolevel}{ks step mark}{mean_k}
  \kswhisk{symmetric}{expSchedSymEdge}{mean_k}
  \kspts{symmetric}{ks sym mark}{mean_k}
  \kspts{mirror}{ks mirror mark}{mean_k}
  \kswhisk{paper}{expCortex}{mean_k}
  \kspts{paper}{ks paper mark}{mean_k}
  \kslab{twolevel}{lab_pos}{e}{mean_k}{anchor=west, xshift=3pt, text=expSchedStep}
  \kslab{twolevel}{lab_pos}{ne}{mean_k}{anchor=south west, xshift=2pt, yshift=1pt, text=expSchedStep}
  \kslab{twolevel}{lab_pos}{sw}{mean_k}{anchor=north east, xshift=-2pt, yshift=-1pt, text=expSchedStep}
  \kslab{symmetric}{lab_pos}{e}{mean_k}{anchor=west, xshift=3pt, text=expSchedSymEdge}
  \kslab{symmetric}{lab_pos}{w}{mean_k}{anchor=east, xshift=-3pt, text=expSchedSymEdge}
  \kslab{mirror}{lab_pos}{e}{mean_k}{anchor=south west, xshift=3pt, text=expCortex}
  \expsnow{\noexpand\node[ks label, text=expCortex, anchor=north west, xshift=3pt, yshift=-1pt] at (axis cs:\kspaperx,\kspapery) {\noexpand\ours{} (ramp up)};}
  \addlegendimage{ks paper mark, only marks}\addlegendentry{\bfseries\ours{}: ramp up}
  \addlegendimage{ks mirror mark, only marks}\addlegendentry{ramp down (reversed)}
  \addlegendimage{ks uniform line, ks uniform mark}\addlegendentry{constant $K_{\mathrm{c}}$}
  \addlegendimage{ks step mark, only marks}\addlegendentry{two-level}
  \addlegendimage{ks sym mark, only marks}\addlegendentry{symmetric}
  \addlegendimage{ks context mark, only marks}\addlegendentry{other gradual ramps}
\nextgroupplot[width=0.24\textwidth, xmin=-0.55, xmax=2.55, ymin=\ksoylo, ymax=\ksoyhi, ytick distance=0.02,
  xtick={0,1,2}, xticklabels={ramp, {two\\level}, {sym-\\metric}}, x tick label style={font=\sffamily\scriptsize, expInk, align=center},
  xlabel={Same layers, different order}, clip=false]
  \expsnow{\noexpand\addplot[draw=none, fill=expGrid, fill opacity=0.9, forget plot]
    coordinates {(-0.55,\ksband-\kshalf) (2.55,\ksband-\kshalf) (2.55,\ksband+\kshalf) (-0.55,\ksband+\kshalf)} \noexpand\closedcycle;}
  \expsnow{\noexpand\addplot[expCortex, densely dashed, line width=0.6pt, forget plot] coordinates {(-0.55,\ksband) (2.55,\ksband)};}
  \expsnow{\noexpand\node[ks label, text=expMuted, anchor=south east] at (axis cs:2.53,\ksband+\kshalf) {seed band of \noexpand\ours{}};}
  \kspts{twolevel}{ks step mark}{order_x}
  \kswhisk{symmetric}{expSchedSymEdge}{order_x}
  \kspts{symmetric}{ks sym mark}{order_x}
  \kspts{mirror}{ks mirror mark}{order_x}
  \kswhisk{paper}{expCortex}{order_x}
  \kspts{paper}{ks paper mark}{order_x}
  \kslab{twolevel}{lab2_pos}{above}{order_x}{anchor=south, yshift=3pt, text=expSchedStep}
  \kslab{twolevel}{lab2_pos}{below}{order_x}{anchor=north, yshift=-3pt, text=expSchedStep}
  \kslab{symmetric}{lab2_pos}{above}{order_x}{anchor=south, yshift=3pt, text=expSchedSymEdge}
  \kslab{symmetric}{lab2_pos}{below}{order_x}{anchor=north, yshift=-3pt, text=expSchedSymEdge}
  \kslab{mirror}{lab2_pos}{above}{order_x}{anchor=south, yshift=3pt, text=expCortex}
  \kslab{paper}{lab2_pos}{below}{order_x}{anchor=north, yshift=-4pt, text=expCortex}
\end{groupplot}
\node[below=0.95cm] at ($(group c1r1.south)!0.5!(group c2r1.south)$) {\pgfplotslegendfromname{exps:ksched}};
\end{tikzpicture}
    \caption{\textbf{The schedule of active columns.} Left: held-out loss
        of the ten schedules of \Cref{tab:k-schedules} against the
        layer-weighted mean of \(K_{\mathrm{c}}\). The constant schedules
        are joined by a line; the run at \(K_{\mathrm{c}} = 2\) is above the
        plotted range. The grey circles are the gradual ramps of the
        configuration search, the dashed line is the dense transformer, and
        the band is the range of the two seeds of the MoE
        transformer. Right: groups
        of schedules that differ only in the order of their layers. Within
        a group, the schedules have the same multiset of layers and hence
        the same FLOPs and stored parameters, so the differences within a
        group are due to the order alone. The shaded band is the mean of the
        two seeds of \ours{} plus or minus \(0.010\) nats, the difference
        between seeds at this scale. Where two seeds were run, they are
    joined by a whisker.}
    \label{fig:k-schedule}
\end{figure}

\paragraph{The schedule of active columns.} We compare ten schedules of
\(K_{\mathrm{c}}\) over depth on the 200M model (\Cref{tab:k-schedules}
and \Cref{fig:k-schedule}). Everything else is set as above, by the same
rules, so that the FLOPs of every schedule are within \(0.97\) to
\(1.02\times\) those of \ours{}. The schedules are: a constant
\(K_{\mathrm{c}} \in \{2, 4, 6, 8\}\) with the depth set by the budget;
the schedule of \ours{} reversed, which has the same layers, FLOPs, and
stored parameters; two symmetric schedules, dense
at the ends and sparse in the middle, and the reverse; and three two-level
schedules on one multiset of layers, sparse then dense, dense then sparse,
and the same layers interleaved. Because an expert of a column has width
\(2D / (K_{\mathrm{c}} K_{\mathrm{e}})\), the stored parameters vary with
the schedule, from \(0.82\times\) those of \ours{} at constant
\(K_{\mathrm{c}} = 8\) to \(1.63\times\) at \(K_{\mathrm{c}} = 2\). The
number of stored parameters does not explain the ordering below, since the
model with the most stored parameters has the highest loss.

Varying the number of active columns with depth generally improves
held-out loss (\Cref{tab:k-schedules}; \Cref{fig:k-schedule}, left).
Among constant schedules, loss decreases as \(K_{\mathrm{c}}\) increases,
but even full attention has a higher loss than \ours{} and the MoE
transformer.
Every varying schedule except the sparse-to-dense step has a lower loss
than all constant schedules. The sparsest constant schedule has a much
higher loss than the others, although its training loss curve is smooth.

The order of layers also affects the loss (\Cref{fig:k-schedule}, right).
Gradual schedules have lower losses than two-level schedules at
similar mean \(K_{\mathrm{c}}\). With the layer composition, FLOPs,
and stored parameters fixed, the dense-to-sparse step has a lower
loss than its reverse, and the symmetric schedule with a dense middle has
a lower loss than the one with dense ends. Alternating sparse and dense
layers gives the same loss as the better step schedule, while a long
run of sparse layers at the start gives the highest loss of the
varying schedules.
Reversing the ramp of \ours{} has a smaller effect, close to the variation
across seeds.

At 200M, the ramp of \ours{} and the symmetric schedule with a dense
middle have the lowest losses, with no clear difference across the two
seeds. Both start with sparse layers and have their largest
\(K_{\mathrm{c}}\) at or after the middle. At 500M, with the stored
parameters of both schedules matched to approximately those of the MoE
transformer, the ramp has the lower loss in a single-seed comparison, so
we use it in \ours{}. The symmetric schedule uses fewer attention FLOPs at
long contexts, and is an alternative there.

\providecommand{\kshape}[1]{\raisebox{-0.32\height}{\tikz[x=1.0cm,
        y=0.34cm]{\draw[expGrid, line width=0.4pt] (0,0) rectangle (1,1);
\draw[expInk, line width=0.7pt] plot coordinates {#1};}}}
\begin{table}[ht]
    \centering
    \footnotesize
    \setlength{\tabcolsep}{2.5pt}
    \begin{tabular}{lcllrcrr}
        \toprule
        Schedule & Shape & \(K_{\mathrm{c}}\) per block & Layers per block &
        Mean \(K_{\mathrm{c}}\) & FLOPs & Stored & Loss \\
        \midrule
        \ours{} (ramp up) & \kshape{(0.000,0.167) (0.118,0.167)
            (0.118,0.333) (0.235,0.333) (0.235,0.500) (0.353,0.500)
            (0.353,0.667) (0.529,0.667) (0.529,0.667) (0.706,0.667)
        (0.706,0.833) (0.882,0.833) (0.882,1.000) (1.000,1.000)} & 3,
        4, 5, 6, 6, 7, 8 & 2, 2, 2, 3, 3, 3, 2 & 5.71 &
        \(1.00\times\) & 762M & 3.092 / 3.088 \\
        ramp down & \kshape{(0.000,1.000) (0.118,1.000) (0.118,0.833)
            (0.294,0.833) (0.294,0.667) (0.471,0.667) (0.471,0.667)
            (0.647,0.667) (0.647,0.500) (0.765,0.500) (0.765,0.333)
        (0.882,0.333) (0.882,0.167) (1.000,0.167)} & 8, 7, 6, 6, 5,
        4, 3 & 2, 3, 3, 3, 2, 2, 2 & 5.71 & \(1.00\times\) & 762M & 3.101 \\
        constant \(K_{\mathrm{c}} = 2\) & \kshape{(0.000,0.000)
        (1.000,0.000)} & 2 & 20 & 2.00 & \(1.00\times\) & 1244M & 3.582 \\
        constant \(K_{\mathrm{c}} = 4\) & \kshape{(0.000,0.333)
        (1.000,0.333)} & 4 & 18 & 4.00 & \(0.98\times\) & 719M & 3.192 \\
        constant \(K_{\mathrm{c}} = 6\) & \kshape{(0.000,0.667)
        (1.000,0.667)} & 6 & 16 & 6.00 & \(1.00\times\) & 751M & 3.157 \\
        constant \(K_{\mathrm{c}} = 8\) & \kshape{(0.000,1.000)
        (1.000,1.000)} & 8 & 16 & 8.00 & \(1.02\times\) & 627M & 3.148 \\
        dense ends & \kshape{(0.000,1.000) (0.167,1.000)
            (0.167,0.500) (0.333,0.500) (0.333,0.167) (0.500,0.167)
            (0.500,0.167) (0.667,0.167) (0.667,0.500) (0.833,0.500)
        (0.833,1.000) (1.000,1.000)} & 8, 5, 3, 3, 5, 8 & 3 each &
        5.33 & \(1.02\times\) & 822M & 3.118 \\
        dense middle & \kshape{(0.000,0.167) (0.167,0.167)
            (0.167,0.500) (0.333,0.500) (0.333,1.000) (0.500,1.000)
            (0.500,1.000) (0.667,1.000) (0.667,0.500) (0.833,0.500)
        (0.833,0.167) (1.000,0.167)} & 3, 5, 8, 8, 5, 3 & 3 each &
        5.33 & \(1.02\times\) & 822M & 3.088 / 3.087 \\
        step up & \kshape{(0.000,0.167) (0.529,0.167) (0.529,1.000)
        (1.000,1.000)} & 3, 8 & 9, 8 & 5.35 & \(0.97\times\) & 743M & 3.169 \\
        step down & \kshape{(0.000,1.000) (0.471,1.000) (0.471,0.167)
        (1.000,0.167)} & 8, 3 & 8, 9 & 5.35 & \(0.97\times\) & 743M & 3.136 \\
        alternating & \kshape{(0.000,0.167) (0.059,0.167)
            (0.059,1.000) (0.118,1.000) (0.118,0.167) (0.176,0.167)
            (0.176,1.000) (0.235,1.000) (0.235,0.167) (0.294,0.167)
            (0.294,1.000) (0.353,1.000) (0.353,0.167) (0.412,0.167)
            (0.412,1.000) (0.471,1.000) (0.471,0.167) (0.529,0.167)
            (0.529,1.000) (0.588,1.000) (0.588,0.167) (0.647,0.167)
            (0.647,1.000) (0.706,1.000) (0.706,0.167) (0.765,0.167)
            (0.765,1.000) (0.824,1.000) (0.824,0.167) (0.882,0.167)
        (0.882,1.000) (0.941,1.000) (0.941,0.167) (1.000,0.167)} & 3,
        8, 3, 8, \dots, 3 & 17 of 1 & 5.35 & \(0.97\times\) & 745M & 3.134 \\
        \bottomrule
    \end{tabular}
    \caption{Ten schedules of active columns on the 200M model,
        \(C = 8\), everything else as in \ours{}; constant
        \(K_{\mathrm{c}} = 8\) is full attention. In the column ``Shape'',
        \(K_{\mathrm{c}}\), from \(2\) to \(8\), is plotted against
        depth. Mean \(K_{\mathrm{c}}\) is weighted by the layers per
        block; FLOPs are
        forward FLOPs per token at \(S = 4096\) relative to \ours{}; stored
        parameters are totals. Held-out loss at the matched final step,
    seeds 42 / 43 if two seeds were run.}
    \label{tab:k-schedules}
\end{table}

\paragraph{FLOP accounting.} In the forward pass of one layer of width
\(D\), the four attention projections cost \(8 D^{2}\) FLOPs per token,
and the three matrices of a gated MLP of width \(4 D\) cost \(2 \cdot 3
\cdot 4 D^{2} = 24 D^{2}\) FLOPs per token. With a causal mask, the
attention of a token is over \(S / 2\)
tokens on average, and the products of queries with keys and of attention
weights with values cost \(2 \cdot 2 D \cdot S / 2 = 2 S D\) FLOPs per
token. The ratio of the two is \(S / (16 D)\), which is between \(0.20\) and
\(0.53\) for our models at \(S = 4096\). In a layer of \ours{} with \(C\)
columns of width \(D / K_{\mathrm{c}}\), each token is routed to
\(K_{\mathrm{c}}\) columns, and each column contains \(S K_{\mathrm{c}} /
C\) tokens, so the
quadratic term is \(2 S D K_{\mathrm{c}} / C\) FLOPs per token, which is the
fraction \(K_{\mathrm{c}} / C\) of the \(2 S D\) FLOPs of full attention, as
in \eqref{eq:attn-cost}. The routers, the
attention residual, and the output head are not counted.

\subsection{Pretraining Results}\label{app:pretraining-results}

The pretraining runs of \Cref{sec:pretraining}, plotted in
\Cref{fig:pretraining}, are listed in \Cref{tab:exp-rungs}: the number of
training tokens and the training compute at each scale, and the
stored parameters and the held-out loss of each model.

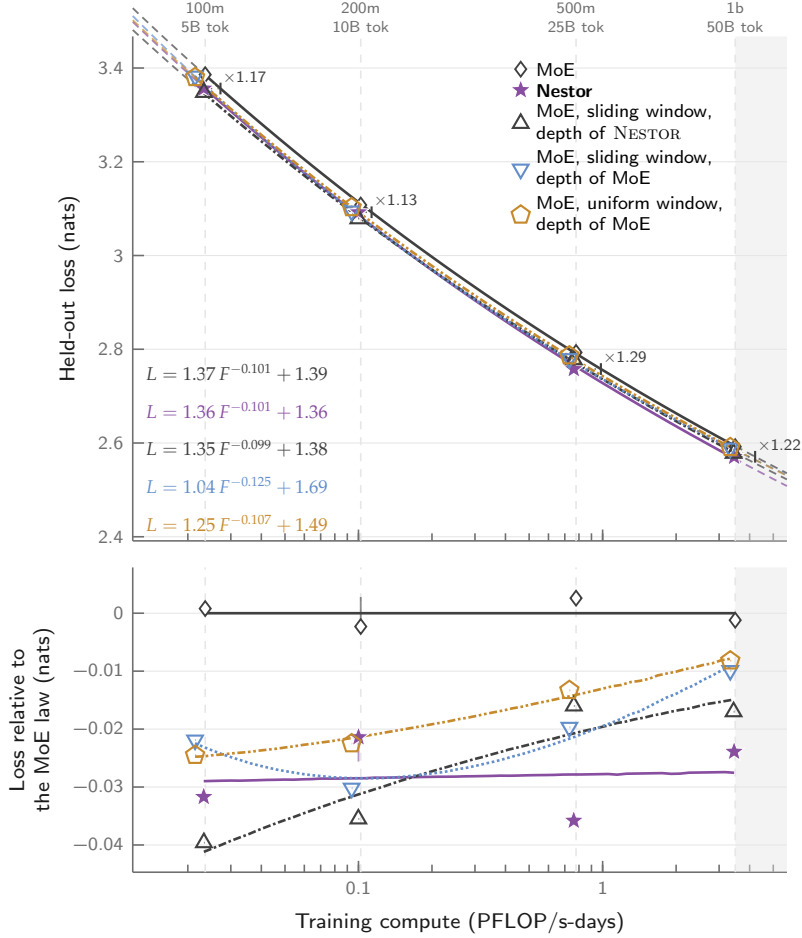
\begin{figure}[ht]
    \centering
    \begingroup
\pgfplotsset{every axis legend/.append style={nodes={text depth=}}}
\def\scalingprefix{scalingswa}
\def\scalingdenseoff{1}
\def\scalingresid{1}
\def\scalingeqbase{5}
\def\scalingextra{%
  \def\scalingline{densely dashdotted}%
  \scalingseries{swa}{3}{expMoe}{triangle}{3.4pt}{\parbox[c]{11em}{\raggedright MoE, sliding window,\\depth of \ours{}}}{fill=none, draw=expMoe, line width=0.8pt}%
  \def\scalingline{densely dotted}%
  \scalingseries{swactl}{4}{expSwaCtl}{triangle}{3.4pt}{\parbox[c]{11em}{\raggedright MoE, sliding window,\\depth of MoE}}{fill=none, draw=expSwaCtl, line width=0.8pt, rotate=180}%
  \def\scalingline{densely dashdotdotted}%
  \scalingseries{swauni}{5}{expSwaUni}{pentagon}{3.6pt}{\parbox[c]{11em}{\raggedright MoE, uniform window,\\depth of MoE}}{fill=none, draw=expSwaUni, line width=0.8pt}%
  \def\scalingline{solid}}
\providecommand{\scalingprefix}{scaling}
\providecommand{\scalingdense}{dense}
\providecommand{\scalingdenselabel}{Dense transformer}
\providecommand{\scalingwidth}{0.62\textwidth}
\providecommand{\scalingheight}{0.5\textwidth}
\providecommand{\scalingwide}{0}
\providecommand{\scalingeqstep}{0.075}
\providecommand{\scalingextra}{}
\providecommand{\scalingeqbase}{2}
\providecommand{\scalingline}{solid}
\providecommand{\scalingdenseoff}{0}
\providecommand{\scalingresid}{0}
\providecommand{\scalingresidheight}{0.34\textwidth}
\expsnow{\noexpand\pgfplotstableread[col sep=comma]{figures/exps/raw_data/derived/\scalingprefix_fits.csv}\noexpand\scalingfits}
\expsnow{\noexpand\pgfplotstableread[col sep=comma]{figures/exps/raw_data/derived/\scalingprefix_multiples.csv}\noexpand\scalingmult}
\expsnow{\noexpand\pgfplotstableread[col sep=comma]{figures/exps/raw_data/derived/\scalingprefix_rungs.csv}\noexpand\scalingrungs}
\expsnow{\noexpand\pgfplotstableread[col sep=comma]{figures/exps/raw_data/derived/\scalingprefix_axis.csv}\noexpand\scalingaxis}
\pgfplotstablegetelem{0}{xlo}\of\scalingaxis\let\xlo\pgfplotsretval
\pgfplotstablegetelem{0}{xhi}\of\scalingaxis\let\xhi\pgfplotsretval
\pgfplotstablegetelem{0}{ylo}\of\scalingaxis\let\ylo\pgfplotsretval
\pgfplotstablegetelem{0}{yhi}\of\scalingaxis\let\yhi\pgfplotsretval
\pgfplotstablegetelem{0}{xmin}\of\scalingfits\let\xdatamin\pgfplotsretval
\pgfplotstablegetelem{0}{xmax}\of\scalingfits\let\xdatamax\pgfplotsretval
\edef\scalingpoints{figures/exps/raw_data/derived/\scalingprefix_points.csv}
\def\scalingpanel{0}
\newcommand{\scalingseries}[7]{%
  \pgfplotstablegetelem{#2}{A}\of\scalingfits\let\A\pgfplotsretval
  \pgfplotstablegetelem{#2}{alpha}\of\scalingfits\let\al\pgfplotsretval
  \pgfplotstablegetelem{#2}{B}\of\scalingfits\let\B\pgfplotsretval
  \pgfplotstablegetelem{#2}{xmin}\of\scalingfits\let\xminv\pgfplotsretval
  \pgfplotstablegetelem{#2}{xmax}\of\scalingfits\let\xmaxv\pgfplotsretval
  \pgfplotstablegetelem{#2}{eq}\of\scalingfits\let\eq\pgfplotsretval
  \ifnum\scalingpanel=0
  \expsnow{\noexpand\addplot[#3, line width=1pt, \scalingline, domain=\xminv:\xmaxv, samples=60, forget plot] {\A*x^(-\al)+\B};}
  \expsnow{\noexpand\addplot[#3, line width=0.7pt, densely dashed, opacity=0.7, domain=\xlo:\xminv, samples=30, forget plot] {\A*x^(-\al)+\B};}
  \expsnow{\noexpand\addplot[#3, line width=0.7pt, densely dashed, opacity=0.7, domain=\xmaxv:\xhi, samples=30, forget plot] {\A*x^(-\al)+\B};}
  \addplot[#3, forget plot, mark=none, draw=none,
           error bars/.cd, y dir=plus, y explicit, error mark=none, error bar style={#3, opacity=0.6, line width=0.7pt}]
    table[col sep=comma, x=x, y=ce, y error=twin, discard if not={series}{#1}] {\scalingpoints};
  \addplot[exps mark, mark=#4, mark size=#5, mark options={#7}]
    table[col sep=comma, x=x, y=ce, discard if not={series}{#1}] {\scalingpoints};
  \addlegendentry{#6}
  \ifnum\scalingwide=1
    \pgfmathsetmacro\eqx{\xlo*exp(0.025*ln(\xhi/\xlo))}%
    \pgfmathsetmacro\eqy{\ylo+(\yhi-\ylo)*(0.07+\scalingeqstep*(\scalingeqbase-#2))}%
    \expsnow{\noexpand\addplot[exps mark, mark=#4, mark size=#5, mark options={#7}, forget plot] coordinates {(\eqx,\eqy)};}
    \expsnow{\noexpand\node[anchor=west, font=\noexpand\sffamily\noexpand\scriptsize, text=#3, inner sep=1pt, xshift=4pt]
      at (axis cs:\eqx,\eqy) {\eq};}
  \else
    \expsnow{\noexpand\node[anchor=south west, font=\noexpand\sffamily\noexpand\scriptsize, text=#3, inner sep=1pt]
      at (rel axis cs:0.015, {0.01 + \scalingeqstep*(\scalingeqbase-#2)}) {\eq};}
  \fi
  \else
  \expsnow{\noexpand\addplot[#3, line width=1pt, \scalingline, domain=\xminv:\xmaxv, samples=60, forget plot] {\A*x^(-\al)+\B-(\Am*x^(-\alm)+\Bm)};}
  \addplot[#3, forget plot, mark=none, draw=none,
           error bars/.cd, y dir=plus, y explicit, error mark=none, error bar style={#3, opacity=0.6, line width=0.7pt}]
    table[col sep=comma, x=x, y=rel, y error=twin, discard if not={series}{#1}] {\scalingpoints};
  \addplot[exps mark, forget plot, mark=#4, mark size=#5, mark options={#7}]
    table[col sep=comma, x=x, y=rel, discard if not={series}{#1}] {\scalingpoints};
  \fi
}
\newcommand{\scalingallseries}{%
  \ifnum\scalingdenseoff=0
    \scalingseries{\scalingdense}{0}{expDense}{square*}{2.4pt}{\scalingdenselabel}{fill=white, draw=expDense, line width=0.7pt}
  \fi
  \scalingseries{moe}{1}{expMoe}{diamond*}{2.8pt}{MoE}{fill=white, draw=expMoe, line width=0.7pt}
  \scalingseries{cortex}{2}{expCortex}{fstar}{4pt}{\bfseries\ours{}}{fill=expCortex, draw=white, line width=0.4pt}
  \scalingextra}
\ifnum\scalingresid=1
  \pgfplotsset{scaling x axis/.style={xticklabel=\empty}}
\else
  \pgfplotsset{scaling x axis/.style={xlabel={Training compute (PFLOP/s-days)}}}
\fi
\begin{tikzpicture}
\begin{axis}[exps, name=scalingmain,
  width=\scalingwidth, height=\scalingheight,
  xmode=log, xmin=\xlo, xmax=\xhi, ymin=\ylo, ymax=\yhi, ytick distance=0.2,
  xtick={0.01,0.1,1,10}, xticklabels={0.01,0.1,1,10}, log ticks with fixed point,
  scaling x axis, ylabel={Held-out loss (nats)},
  legend pos=north east, legend style={row sep=-1pt},
  clip=false,
]
  \expsnow{\noexpand\addplot[draw=none, fill=expGrid, fill opacity=0.45, forget plot]
    coordinates {(\xdatamax,\ylo) (\xhi,\ylo) (\xhi,\yhi) (\xdatamax,\yhi)} \noexpand\closedcycle;}
  \scalingallseries
  \pgfplotsinvokeforeach{0,1,2,3}{
    \pgfplotstablegetelem{#1}{xstar}\of\scalingmult\let\xs\pgfplotsretval
    \pgfplotstablegetelem{#1}{y}\of\scalingmult\let\ys\pgfplotsretval
    \pgfplotstablegetelem{#1}{xmoe}\of\scalingmult\let\xm\pgfplotsretval
    \pgfplotstablegetelem{#1}{moe}\of\scalingmult\let\lm\pgfplotsretval
    \ifnum\scalingdenseoff=0
      \pgfplotstablegetelem{#1}{xdense}\of\scalingmult\let\xd\pgfplotsretval
      \pgfplotstablegetelem{#1}{dense}\of\scalingmult\let\ld\pgfplotsretval
    \else
      \let\xd\xm
    \fi
    \expsnow{\noexpand\draw[expMuted, dotted, line width=0.6pt] (axis cs:\xs,\ys) -- (axis cs:\xd,\ys);}
    \expsnow{\noexpand\draw[expMoe, line width=0.8pt] ([yshift=-2.2pt]axis cs:\xm,\ys) -- ([yshift=2.2pt]axis cs:\xm,\ys)
      node[above right, exps annot, text=expMoe, inner sep=0.5pt, yshift=-1pt] {\lm};}
    \ifnum\scalingdenseoff=0
      \ifnum#1>1 \def\dplace{below left, yshift=-3.4pt}\else\def\dplace{above right, yshift=-1pt}\fi
      \expsnow{\noexpand\draw[expDense, line width=0.8pt] ([yshift=-2.2pt]axis cs:\xd,\ys) -- ([yshift=2.2pt]axis cs:\xd,\ys)
        node[\dplace, exps annot, text=expDense, inner sep=0.5pt] {\ld};}
    \fi
  }
  \pgfplotsinvokeforeach{0,1,2,3}{
    \pgfplotstablegetelem{#1}{x}\of\scalingrungs\let\xr\pgfplotsretval
    \pgfplotstablegetelem{#1}{label}\of\scalingrungs\let\lr\pgfplotsretval
    \ifnum\scalingwide=1
      \expsnow{\noexpand\draw[expMuted, line width=0.6pt] (axis cs:\xr,\yhi) -- ([yshift=-3pt]axis cs:\xr,\yhi)
        (axis cs:\xr,\yhi) node[above, exps annot, text=expMuted, align=center, inner sep=1.5pt] {\lr};}
    \else
      \expsnow{\noexpand\draw[expGrid, dashed, line width=0.6pt] ({axis cs:\xr,0}|-{rel axis cs:0,0}) -- ({axis cs:\xr,0}|-{rel axis cs:0,1})
        node[above, exps annot, text=expMuted, align=center, inner sep=1.5pt] {\lr};}
    \fi
  }
\end{axis}
\ifnum\scalingresid=1
  \def\scalingpanel{1}
  \pgfplotstablegetelem{1}{A}\of\scalingfits\let\Am\pgfplotsretval
  \pgfplotstablegetelem{1}{alpha}\of\scalingfits\let\alm\pgfplotsretval
  \pgfplotstablegetelem{1}{B}\of\scalingfits\let\Bm\pgfplotsretval
  \pgfplotstablegetelem{0}{rello}\of\scalingaxis\let\rlo\pgfplotsretval
  \pgfplotstablegetelem{0}{relhi}\of\scalingaxis\let\rhi\pgfplotsretval
  \begin{axis}[exps,
    at={(scalingmain.south west)}, anchor=north west, yshift=-10pt,
    width=\scalingwidth, height=\scalingresidheight,
    xmode=log, xmin=\xlo, xmax=\xhi, ymin=\rlo, ymax=\rhi, ytick distance=0.01,
    yticklabel={\pgfmathparse{abs(\tick)}\ifdim\tick pt<0pt \textminus\fi\pgfmathprintnumber{\pgfmathresult}},
    xtick={0.01,0.1,1,10}, xticklabels={0.01,0.1,1,10}, log ticks with fixed point,
    xlabel={Training compute (PFLOP/s-days)}, ylabel={Loss relative to\\the MoE law (nats)}, ylabel style={align=center},
    clip=false,
  ]
    \expsnow{\noexpand\addplot[draw=none, fill=expGrid, fill opacity=0.45, forget plot]
      coordinates {(\xdatamax,\rlo) (\xhi,\rlo) (\xhi,\rhi) (\xdatamax,\rhi)} \noexpand\closedcycle;}
    \pgfplotsinvokeforeach{0,1,2,3}{
      \pgfplotstablegetelem{#1}{x}\of\scalingrungs\let\xr\pgfplotsretval
      \expsnow{\noexpand\draw[expGrid, dashed, line width=0.6pt] ({axis cs:\xr,0}|-{rel axis cs:0,0}) -- ({axis cs:\xr,0}|-{rel axis cs:0,1});}
    }
    \scalingallseries
  \end{axis}
\fi
\end{tikzpicture}
\endgroup
    \caption{\textbf{Pretraining with sliding window baselines.}
        \Cref{fig:pretraining} without the dense transformer and with three
        sliding window baselines, described in
        \Cref{app:pretraining-results}: the MoE transformer with sliding
        window attention in place of the column attention of \ours{}: at
        the depth of \ours{}, with windows that give the same number of
        attention scores as \ours{} in each layer (triangles); at the depth
        of the MoE, with the same windows (inverted triangles); and at the
        depth of the MoE, with one uniform window and periodic full attention
        layers (pentagons). Top: held-out loss against training compute,
        with the fitted laws and the compute multiples of
        \Cref{fig:pretraining}. Bottom: the same losses minus the fitted law
        of the MoE at the same training compute (the zero line), on a scale
        where differences of a few hundredths of a nat are visible; at
        200M, a second seed is drawn as a whisker. The two windowed
        models at the
        depth of the MoE use fewer FLOPs per token than the MoE, so the MoE
        law is evaluated at their smaller compute, and their values in the
        bottom panel are lower than the differences between their losses and
    the losses of the MoE.}
    \label{fig:pretraining-swa}
\end{figure}

In \Cref{fig:pretraining-swa}, we compare these runs with three sliding
window baselines. Each baseline is the MoE transformer with sliding window
attention in place of the column attention of \ours{}, pretrained from
scratch at every scale with the same recipe, token budget, and attention
residuals as the other models, with a single seed. The three baselines
differ in how the attention FLOPs are distributed over depth. The first has
the depth and the blocks of \ours{}. In each block, we choose the window so
that a layer has as many attention scores as the corresponding layer
of \ours{}, which gives windows of \(798\), \(1{,}200\), \(1{,}460\),
\(2{,}477\), \(2{,}477\), and \(2{,}113\) tokens in the six windowed blocks
and full attention in the last block. We choose its number of experts so
that its stored parameters are within \(1.6\%\) of those of \ours{}, and
its FLOPs per token are then within \(0.5\%\) of those of \ours{}. The
second has the depth, the blocks, and the experts of the MoE transformer.
Its layers have the same windows at the same relative depth, and its top two
to four layers have full attention. The attention FLOPs saved by the
windows are not used elsewhere, so this baseline has the stored parameters of
the MoE transformer and \(5\)--\(9\%\) fewer FLOPs per token. The third has
the shape and the FLOPs per token of the second, but the same attention
FLOPs are distributed uniformly over depth: every windowed layer has a window of
about \(1{,}800\) tokens, and a full attention layer follows every five
windowed layers at 100M and every seven at the other scales. We choose the
window so that the model has as many attention scores as the second
baseline.

\begin{table}[ht]
    \centering
    \small
    \setlength{\tabcolsep}{5pt}
    \begin{tabular}{lcc@{\hspace{1.6em}}lrrr}
        \toprule
        Size & Tokens &
        \begin{tabular}[c]{@{}c@{}}
            Compute\\(PFLOP/s-days)
        \end{tabular} &
        Model & Total params & Loss &
        \begin{tabular}[c]{@{}c@{}}
            Estimated compute\\at equal loss
        \end{tabular} \\
        \midrule
        \multirow{3}{*}{100M}
        & \multirow{3}{*}{5B}
        & \multirow{3}{*}{0.02}
        & Dense transformer & 0.14B & 3.578 & $3.1\times$ \\
        & & & MoE           & 0.37B & 3.386 & $1.17\times$ \\
        & & & \ours{}       & 0.38B & \textbf{3.356} & $1.00\times$ \\
        \midrule
        \multirow{3}{*}{200M}
        & \multirow{3}{*}{10B}
        & \multirow{3}{*}{0.10}
        & Dense transformer & 0.23B & 3.293 & $3.2\times$ \\
        & & & MoE           & 0.79B & 3.108 & $1.13\times$ \\
        & & & \ours{}       & 0.76B & \textbf{3.092} & $1.00\times$ \\
        \midrule
        \multirow{3}{*}{500M}
        & \multirow{3}{*}{25B}
        & \multirow{3}{*}{0.78}
        & Dense transformer & 0.55B & 2.953 & $4.1\times$ \\
        & & & MoE           & 2.41B & 2.793 & $1.29\times$ \\
        & & & \ours{}       & 2.44B & \textbf{2.758} & $1.00\times$ \\
        \midrule
        \multirow{3}{*}{1B}
        & \multirow{3}{*}{50B}
        & \multirow{3}{*}{3.49}
        & Dense transformer & 1.10B & 2.743 & $4.1\times$ \\
        & & & MoE           & 5.49B & 2.592 & $1.22\times$ \\
        & & & \ours{}       & 5.53B & \textbf{2.571} & $1.00\times$ \\
        \bottomrule
    \end{tabular}
    \caption{\textbf{Pretraining results at four model scales.} These are
        the numbers plotted in \Cref{fig:pretraining} of
        \Cref{sec:pretraining}. The last column is the estimated compute at
        which each baseline reaches the held-out loss of \ours{}
        at the same scale, from the fitted law $L(F) = A\,F^{-\alpha}
        + L_{\infty}$,
    normalized by the training compute of \ours{}.}
    \label{tab:exp-rungs}
\end{table}

\paragraph{Sliding window attention and \ours{}.} The sliding window
baselines are more compute-efficient than the MoE, and at some
scales they are comparable to \ours{} at similar compute. However, the
compute efficiency of each baseline relative to \ours{} decreases
with scale. The fitted scaling laws agree with this: the sliding window
baselines have a higher irreducible loss (the loss at infinite compute) than
\ours{}. One possible reason is that a sliding window restricts the context
of a token to the most recent tokens, whereas in \ours{} the context of a
token in a column depends on the input and can contain tokens from anywhere
in the sequence. This class of attention patterns is strictly more
expressive, and could scale better.

\subsection{Architecture Ablations}
\label{app:ablations}

\paragraph{Attribution of the improvement over the MoE transformer.} The MLP
of \ours{} differs from that of the MoE transformer baseline in three
ways. First, it has two groups of experts with separate routers: the
\(E_{\mathrm{s}}\) experts whose weights are tied across columns, and the
\(E\) experts of each column. Second, the two groups are weighted by the gate
\eqref{eq:app-path-gate}. Third, each token uses more experts of smaller
width: at the 200M active parameter scale, a token uses \(7\) to
\(10\) experts per
layer in \ours{}, and \(3\) in the baseline. We test whether these
differences, or column-routed attention alone, explain the improvement in
\Cref{fig:pretraining}. The held-out losses of eight models at the
200M active parameter scale and of three models at the 500M scale are in
\Cref{tab:attribution}. All models have the setup of \Cref{sec:pretraining}, the
attention residual, and an active MLP width of \(4D\).

\begin{table}[ht]
    \centering
    \small
    \setlength{\tabcolsep}{4pt}
    \begin{tabular}{cl>{\raggedright\arraybackslash}p{0.46\linewidth}lcc}
        \toprule
        & Attention & Experts & Stored & FLOPs & Held-out loss \\
        \midrule
        \multicolumn{6}{l}{\emph{200M active parameters, 10B tokens}} \\
        1 & full & \(E = 24\), \(K_{\mathrm{e}} = 3\) (MoE baseline) & 787M &
        \(1.00\times\) & 3.108 / 3.113 \\
        2 & full & \(E = 48\), \(K_{\mathrm{e}} = 6\), half width & 781M &
        \(1.00\times\) & 3.107 \\
        3 & full & \(E = 12\), \(K_{\mathrm{e}} = 3\) & 471M & \(1.00\times\) &
        3.170 \\
        4 & full (\(C = 1\)) & experts of \ours{}: \(E_{\mathrm{s}} = 8\),
        \(K_{\mathrm{s}} = 2\), \(E = 8\), \(K_{\mathrm{e}} = 2\), width
        \(D\) & 470M & \(1.00\times\) & 3.170 \\
        5 & full (\(C = 1\)) & experts of \ours{} at the width of the
        baseline: \(E_{\mathrm{s}} = 16\), \(K_{\mathrm{s}} = 2\), \(E = 8\),
        \(K_{\mathrm{e}} = 1\) & 787M & \(1.00\times\) & 3.116 / 3.113 \\
        6 & column-routed & \(E = 24\), \(K_{\mathrm{e}} = 3\), tied across
        columns, with an earlier schedule of active columns & 935M &
        \(1.06\times\) & 3.107 / 3.124 \\
        7 & column-routed & \(E = 24\), \(K_{\mathrm{e}} = 3\), tied across
        columns, with the schedule of \ours{} & 843M & \(0.98\times\) &
        3.097 \\
        8 & column-routed & experts of \ours{} (full model) & 762M &
        \(0.98\times\) & 3.092 / 3.088 \\
        \midrule
        \multicolumn{6}{l}{\emph{500M active parameters, 25B tokens}} \\
        9 & full & \(E = 24\), \(K_{\mathrm{e}} = 3\) (MoE baseline) & 2.41B &
        \(1.00\times\) & 2.793 \\
        10 & full (\(C = 1\)) & experts of \ours{} at the width of the
        baseline, as in row 5 & 2.41B & \(1.00\times\) & 2.794 \\
        11 & column-routed & experts of \ours{} (full model) & 2.44B &
        \(0.98\times\) & 2.758 \\
        \bottomrule
    \end{tabular}
    \caption{Held-out loss of eight models at the 200M scale and three
        models at the 500M scale. FLOPs are forward FLOPs per token relative to
        the dense transformer at \(S = 4096\). Where a second seed was run, both
    losses are given (seed 42 / seed 43).}
    \label{tab:attribution}
\end{table}

An MoE transformer with twice as many experts of
half the width (row 2) has the same loss as the baseline (row 1), so the
smaller experts of \ours{} do not explain the improvement. \ours{} with \(C =
1\) has the same loss as an MoE transformer with the same number of stored
parameters, both at 470M stored parameters (rows 3 and 4) and at the storage
of the baseline (rows 5 and 1; at the 500M scale, rows 10 and 9). With one
column, every token can use the experts of the column, so the models in rows
4, 5, and 10 have all three differences listed above and no column routing.
Column-routed attention with the MLP of the baseline (rows 6 and 7) gives
losses between 3.097 and 3.124 over three runs, which is the loss of the
baseline within the variation between seeds. The full model (row 8), in
which the experts are nested within the columns, has a loss lower than the
baseline by \(0.016\) and \(0.025\) nats on the two seeds, and by
\(0.035\) nats at the
500M scale (row 11).

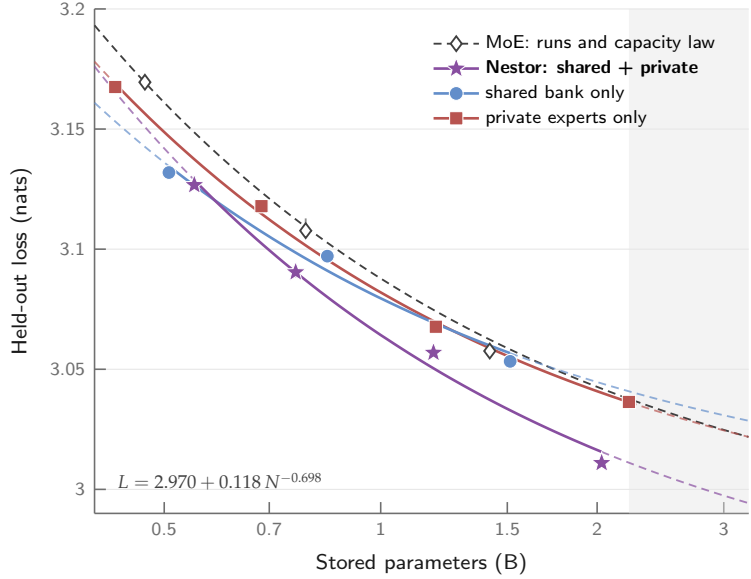
\begin{figure}[t]
    \centering
    \pgfplotstableread[col sep=comma]{figures/exps/raw_data/derived/banks_fits.csv}\banksfits
\pgfplotstableread[col sep=comma]{figures/exps/raw_data/derived/banks_axis.csv}\banksaxis
\def\bankspoints{figures/exps/raw_data/derived/banks_points.csv}
\pgfplotstablegetelem{0}{xlo}\of\banksaxis\let\xlo\pgfplotsretval
\pgfplotstablegetelem{0}{xhi}\of\banksaxis\let\xhi\pgfplotsretval
\pgfplotstablegetelem{0}{ylo}\of\banksaxis\let\ylo\pgfplotsretval
\pgfplotstablegetelem{0}{yhi}\of\banksaxis\let\yhi\pgfplotsretval
\pgfplotstablegetelem{0}{xmax}\of\banksaxis\let\xdatamax\pgfplotsretval
\pgfplotstablegetelem{0}{c}\of\banksfits\let\lawc\pgfplotsretval
\pgfplotstablegetelem{0}{a}\of\banksfits\let\lawa\pgfplotsretval
\pgfplotstablegetelem{0}{alpha}\of\banksfits\let\lawal\pgfplotsretval
\pgfplotstablegetelem{0}{eq}\of\banksfits\let\laweq\pgfplotsretval
\pgfplotsset{
  banks hybrid line/.style={expCortex, line width=1pt},
  banks hybrid mark/.style={mark=fstar, mark size=4pt, mark options={fill=expCortex, draw=white, line width=0.4pt}},
  banks shared line/.style={expBankShared, line width=1pt},
  banks shared mark/.style={mark=*, mark size=2.6pt, mark options={fill=expBankShared, draw=white, line width=0.4pt}},
  banks private line/.style={expBankPrivate, line width=1pt},
  banks private mark/.style={mark=square*, mark size=2.5pt, mark options={fill=expBankPrivate, draw=white, line width=0.4pt}},
  banks moe line/.style={expMoe, densely dashed, line width=0.7pt},
  banks moe mark/.style={mark=diamond*, mark size=2.8pt, mark options={fill=white, draw=expMoe, line width=0.7pt, solid}},
}
\newcommand{\banksseries}[4]{%
  \pgfplotstablegetelem{#2}{c}\of\banksfits\let\C\pgfplotsretval
  \pgfplotstablegetelem{#2}{a}\of\banksfits\let\A\pgfplotsretval
  \pgfplotstablegetelem{#2}{xmin}\of\banksfits\let\xminv\pgfplotsretval
  \pgfplotstablegetelem{#2}{xmax}\of\banksfits\let\xmaxv\pgfplotsretval
  \expsnow{\noexpand\addplot[#3, domain=\xminv:\xmaxv, samples=60, forget plot] {\C+\A*x^(-\lawal)};}
  \expsnow{\noexpand\addplot[#3, line width=0.7pt, densely dashed, opacity=0.7, domain=\xlo:\xminv, samples=30, forget plot] {\C+\A*x^(-\lawal)};}
  \expsnow{\noexpand\addplot[#3, line width=0.7pt, densely dashed, opacity=0.7, domain=\xmaxv:\xhi, samples=30, forget plot] {\C+\A*x^(-\lawal)};}
  \addplot[#3, forget plot, mark=none, draw=none,
           error bars/.cd, y dir=both, y explicit, error mark=none, error bar style={#3, solid, opacity=0.6, line width=0.7pt}]
    table[col sep=comma, x=x, y=ce, y error plus=eplus, y error minus=eminus, discard if not={series}{#1}] {\bankspoints};
  \addplot[only marks, #4, forget plot] table[col sep=comma, x=x, y=ce, discard if not={series}{#1}] {\bankspoints};
}
\begin{tikzpicture}
\begin{axis}[exps,
  width=0.62\textwidth, height=0.5\textwidth,
  xmode=log, xmin=\xlo, xmax=\xhi, ymin=\ylo, ymax=\yhi, ytick distance=0.05,
  xtick={0.5,0.7,1,1.5,2,3}, xticklabels={0.5,0.7,1,1.5,2,3}, minor xtick={}, log ticks with fixed point,
  xlabel={Stored parameters (B)}, ylabel={Held-out loss (nats)},
  legend pos=north east, legend style={row sep=-1pt},
  clip=false,
]
  \expsnow{\noexpand\addplot[draw=none, fill=expGrid, fill opacity=0.45, forget plot]
    coordinates {(\xdatamax,\ylo) (\xhi,\ylo) (\xhi,\yhi) (\xdatamax,\yhi)} \noexpand\closedcycle;}
  \expsnow{\noexpand\addplot[banks moe line, forget plot, domain=\xlo:\xhi, samples=80] {\lawc+\lawa*x^(-\lawal)};}
  \addplot[expMoe, forget plot, mark=none, draw=none,
           error bars/.cd, y dir=both, y explicit, error mark=none, error bar style={expMoe, solid, opacity=0.6, line width=0.7pt}]
    table[col sep=comma, x=x, y=ce, y error plus=eplus, y error minus=eminus, discard if not={series}{moe}] {\bankspoints};
  \addplot[only marks, banks moe mark, forget plot] table[col sep=comma, x=x, y=ce, discard if not={series}{moe}] {\bankspoints};
  \banksseries{private}{3}{banks private line}{banks private mark}
  \banksseries{shared}{2}{banks shared line}{banks shared mark}
  \banksseries{hybrid}{1}{banks hybrid line}{banks hybrid mark}
  \addlegendimage{banks moe line, banks moe mark}\addlegendentry{MoE: runs and capacity law}
  \addlegendimage{banks hybrid line, banks hybrid mark}\addlegendentry{\bfseries\ours{}: shared + private}
  \addlegendimage{banks shared line, banks shared mark}\addlegendentry{shared bank only}
  \addlegendimage{banks private line, banks private mark}\addlegendentry{private experts only}
  \expsnow{\noexpand\node[anchor=south west, font=\noexpand\sffamily\noexpand\scriptsize, text=expMoe, inner sep=1pt]
    at (rel axis cs:0.03, 0.04) {\laweq};}
\end{axis}
\end{tikzpicture}
    \caption{Ablation of the placement of the experts of \ours{}, at
        the 200M scale. We vary the stored parameters of three designs, with
        the columns, the schedule of active columns, the attention residual,
        the training recipe and the \(10\)B-token budget of
        \Cref{sec:pretraining} held fixed, so that every model has the same
        active FLOPs: the design of \ours{}, \(E_{\mathrm{s}} = 8\) shared
        experts with \(K_{\mathrm{s}} = 2\) and \(E \in \{4, 8, 16, 32\}\)
        experts per column with \(K_{\mathrm{e}} = 2\); a shared bank only,
        \(E_{\mathrm{s}} \in \{12, 24, 48\}\) with \(K_{\mathrm{s}} = 3\) and
        no experts in the columns, which is the MLP of the MoE baseline
        with column-routed attention; and experts in the columns only,
        \(E \in \{2, 4, 8, 16\}\) with \(K_{\mathrm{e}} = 1\) and no shared
        experts. The dashed
        line is the capacity law \(L(N) = c + a N^{-0.698}\) of the MoE
        baseline, fitted to its runs with \(16\), \(24\) and \(48\) experts
        (diamonds), and each design is fitted with the same exponent, solid
        over its measured span and dashed beyond it. Where two seeds were
        run, they are joined by a whisker. \ours{} is
        \(0.018\)--\(0.031\) nats below
        the law at every size. With experts in the columns only, the loss
        is below the law at small sizes and closer to the law at larger
        sizes. With a shared bank only, the loss is on the law from \(24\)
        to \(48\) experts. The shared bank with \(96\)
        experts is omitted: its routing collapsed, and \(54\) of its \(96\)
    experts per layer were selected by under \(1\%\) of the tokens.}
    \label{fig:bank-designs}
\end{figure}

\begin{figure}[t]
    \centering
    \pgfplotstableread[col sep=comma]{figures/exps/raw_data/derived/attnres_fits.csv}\attnresfits
\pgfplotstableread[col sep=comma]{figures/exps/raw_data/derived/attnres_multiples.csv}\attnresmult
\pgfplotstableread[col sep=comma]{figures/exps/raw_data/derived/attnres_rungs.csv}\attnresrungs
\pgfplotstableread[col sep=comma]{figures/exps/raw_data/derived/attnres_axis.csv}\attnresaxis
\pgfplotstablegetelem{0}{xlo}\of\attnresaxis\let\xlo\pgfplotsretval
\pgfplotstablegetelem{0}{xhi}\of\attnresaxis\let\xhi\pgfplotsretval
\pgfplotstablegetelem{0}{ylo}\of\attnresaxis\let\ylo\pgfplotsretval
\pgfplotstablegetelem{0}{yhi}\of\attnresaxis\let\yhi\pgfplotsretval
\pgfplotstablegetelem{0}{xmin}\of\attnresfits\let\xdatamin\pgfplotsretval
\pgfplotstablegetelem{0}{xmax}\of\attnresfits\let\xdatamax\pgfplotsretval
\def\attnrespoints{figures/exps/raw_data/derived/attnres_points.csv}
\newcommand{\attnresseries}[8]{%
  \pgfplotstablegetelem{#2}{A}\of\attnresfits\let\A\pgfplotsretval
  \pgfplotstablegetelem{#2}{alpha}\of\attnresfits\let\al\pgfplotsretval
  \pgfplotstablegetelem{#2}{B}\of\attnresfits\let\B\pgfplotsretval
  \pgfplotstablegetelem{#2}{xmin}\of\attnresfits\let\xminv\pgfplotsretval
  \pgfplotstablegetelem{#2}{xmax}\of\attnresfits\let\xmaxv\pgfplotsretval
  \pgfplotstablegetelem{#2}{eq}\of\attnresfits\let\eq\pgfplotsretval
  \expsnow{\noexpand\addplot[#3, line width=1pt, domain=\xminv:\xmaxv, samples=60, forget plot] {\A*x^(-\al)+\B};}
  \expsnow{\noexpand\addplot[#3, line width=0.7pt, densely dashed, opacity=0.7, domain=\xlo:\xminv, samples=30, forget plot] {\A*x^(-\al)+\B};}
  \expsnow{\noexpand\addplot[#3, line width=0.7pt, densely dashed, opacity=0.7, domain=\xmaxv:\xhi, samples=30, forget plot] {\A*x^(-\al)+\B};}
  \addplot[#3, forget plot, mark=none, draw=none,
           error bars/.cd, y dir=plus, y explicit, error mark=none, error bar style={#3, opacity=0.6, line width=0.7pt}]
    table[col sep=comma, x=x, y=ce, y error=twin, discard if not={series}{#1}] {\attnrespoints};
  \addplot[exps mark, mark=#4, mark size=#5, mark options={#7}]
    table[col sep=comma, x=x, y=ce, discard if not={series}{#1}] {\attnrespoints};
  \addlegendentry{#6}
  \expsnow{\noexpand\node[anchor=south west, font=\noexpand\sffamily\noexpand\scriptsize, text=#8, inner sep=1pt]
    at (rel axis cs:0.015, {0.01 + 0.075*(2-#2)}) {\eq};}
}
\colorlet{readOursB}{expCortex!62!white}
\colorlet{readOursC}{expCortex!44!white}
\colorlet{readOursBtext}{expCortex!85!black}
\colorlet{readOursCtext}{expCortex!55!black}
\begin{tikzpicture}
\begin{axis}[exps,
  width=0.62\textwidth, height=0.5\textwidth,
  xmode=log, xmin=\xlo, xmax=\xhi, ymin=\ylo, ymax=\yhi, ytick distance=0.2,
  xtick={0.01,0.1,1,10}, xticklabels={0.01,0.1,1,10}, log ticks with fixed point,
  xlabel={Training compute (PFLOP/s-days)}, ylabel={Held-out loss (nats)},
  legend pos=north east, legend style={row sep=-1pt},
  clip=false,
]
  \expsnow{\noexpand\addplot[draw=none, fill=expGrid, fill opacity=0.45, forget plot]
    coordinates {(\xdatamax,\ylo) (\xhi,\ylo) (\xhi,\yhi) (\xdatamax,\yhi)} \noexpand\closedcycle;}
  \attnresseries{nores}{0}{readOursC}{triangle*}{2.8pt}{None}{fill=readOursC, draw=white, line width=0.4pt}{readOursCtext}
  \attnresseries{kimi}{1}{readOursB}{*}{2.3pt}{AttnRes}{fill=readOursB, draw=white, line width=0.4pt}{readOursBtext}
  \attnresseries{d0h}{2}{expCortex}{fstar}{4pt}{\bfseries Col.\ AttnRes (ours)}{fill=expCortex, draw=white, line width=0.4pt}{expCortex}
  \pgfplotsinvokeforeach{0,1,2,3}{
    \pgfplotstablegetelem{#1}{xstar}\of\attnresmult\let\xs\pgfplotsretval
    \pgfplotstablegetelem{#1}{y}\of\attnresmult\let\ys\pgfplotsretval
    \pgfplotstablegetelem{#1}{xkimi}\of\attnresmult\let\xk\pgfplotsretval
    \pgfplotstablegetelem{#1}{xnores}\of\attnresmult\let\xn\pgfplotsretval
    \pgfplotstablegetelem{#1}{xfar}\of\attnresmult\let\xf\pgfplotsretval
    \pgfplotstablegetelem{#1}{kimi}\of\attnresmult\let\lk\pgfplotsretval
    \pgfplotstablegetelem{#1}{nores}\of\attnresmult\let\ln\pgfplotsretval
    \expsnow{\noexpand\draw[expMuted, dotted, line width=0.6pt] (axis cs:\xs,\ys) -- (axis cs:\xf,\ys);}
    \expsnow{\noexpand\draw[readOursCtext, line width=0.8pt] ([yshift=-2.2pt]axis cs:\xn,\ys) -- ([yshift=2.2pt]axis cs:\xn,\ys)
      node[above right, exps annot, text=readOursCtext, inner sep=0.5pt, yshift=-1pt] {\ln};}
    \expsnow{\noexpand\draw[readOursBtext, line width=0.8pt] ([yshift=-2.2pt]axis cs:\xk,\ys) -- ([yshift=2.2pt]axis cs:\xk,\ys)
      node[below right, exps annot, text=readOursBtext, inner sep=0.5pt, xshift=3pt, yshift=-1pt] {\lk};}
  }
  \pgfplotsinvokeforeach{0,1,2,3}{
    \pgfplotstablegetelem{#1}{x}\of\attnresrungs\let\xr\pgfplotsretval
    \pgfplotstablegetelem{#1}{label}\of\attnresrungs\let\lr\pgfplotsretval
    \expsnow{\noexpand\draw[expGrid, dashed, line width=0.6pt] ({axis cs:\xr,0}|-{rel axis cs:0,0}) -- ({axis cs:\xr,0}|-{rel axis cs:0,1})
      node[above, exps annot, text=expMuted, align=center, inner sep=1.5pt] {\lr};}
  }
\end{axis}
\end{tikzpicture}
    \caption{Ablation of the attention residual of \ours{} at 100M, 200M,
        500M, and 1B, on the axes of \Cref{fig:pretraining}. We compare the
        column attention residual (Col.\ AttnRes), the merged attention
        residual (AttnRes), which is that of \citet{kimi2026attnres} with the
        changes of \Cref{app:architecture}, and no attention
        residual. The rest of the model is unchanged. Each series is fitted by
        $L(F) = A\,F^{-\alpha} + L_{\infty}$, and a horizontal tick
        marks the projected
        compute at which each variant reaches the loss of
        Col.\ AttnRes. Removing the attention
        residual increases the loss by $0.073$--$0.086$ nats at every scale,
        which corresponds to $1.5$--$2\times$ the compute. The two attention
    residuals have the same loss up to 500M and differ at 1B.}
    \label{fig:read-ablation}
\end{figure}

\subsection{Domain Adaptation Settings}
\label{app:domain-adaptation-settings}

\paragraph{Starting checkpoints.}
We use the 500M \ours{}, 500M MoE transformer, and 1B dense transformer
from \Cref{sec:pretraining}. Before domain adaptation, the context length of
each model is extended from \(4\)k to \(32\)k on \(0.5\)B DCLM-baseline tokens,
using the equal-token protocol of \Cref{sec:long-context-extension}.
The extension documents have at least \(32{,}768\) tokens, and
the rotary embedding base is \(2\times 10^{6}\).
The held-out losses of the resulting checkpoints are in
\Cref{tab:domain-adaptation-initial}. Each model has one starting
checkpoint for every target domain and both replay settings. The
adaptation runs are independent and do not start from a model adapted
to another domain.

\begin{table}[ht]
    \centering
    \caption{Held-out loss before domain adaptation, after context extension
        to \(32\)k. All losses are from the \(32\)k evaluation described below.
        These are the starting losses for adaptation and forgetting
    in \Cref{fig:domain-adaptation}.}
    \label{tab:domain-adaptation-initial}
    \begin{tabular}{lrrrrr}
        \toprule
        Model & DCLM & PG-19 & OpenWebMath & AlgebraicStack & German \\
        \midrule
        500M \ours{} & 2.779 & 2.732 & 2.417 & 1.789 & 4.171 \\
        500M MoE & 2.810 & 2.765 & 2.449 & 1.813 & 4.224 \\
        1B dense & 2.779 & 2.719 & 2.412 & 1.720 & 4.164 \\
        \bottomrule
    \end{tabular}
\end{table}

\paragraph{Data.}
We use the published training and test splits of PG-19, OpenWebMath,
AlgebraicStack, and the German subset of FineWeb-2, as cited in
\Cref{sec:domain-adaptation}. OpenWebMath and AlgebraicStack are taken
from Proof-Pile-2. The available training data consist of the PG-19 training
split, the first 12 OpenWebMath training files, all 79 AlgebraicStack
training files, and the first two German training files of FineWeb-2.
All corpora are tokenized with the \texttt{cl100k\_base} tokenizer.
We separate documents
with an end-of-text token, concatenate them, and train on shuffled
\(32{,}768\)-token sequences. Sequences may cross document boundaries;
causal attention is not reset at those boundaries. The same data and
sampling order are used for all three models within each setting.

\paragraph{Adaptation protocol.}
Each run has 477 optimizer steps with 32 sequences per step,
or \(1{,}048{,}576\) tokens per step. This gives \(500{,}170{,}752\)
tokens, rounded to \(0.5\)B in the text.
For both sparse models, we freeze the router parameters and
load-balancing biases and disable routing noise. All other trainable
parameters, including attention projections, experts, normalization
parameters, embeddings, the output head, and attention residuals, are
updated. Routing decisions are recomputed from the current hidden states,
so freezing the routers does not fix the token assignments.
The dense model is fully fine-tuned. Each model, domain, and replay
setting has one run, with seed 42. The data sampler uses seed 0 to
shuffle the training sequences.

\paragraph{Optimization.}
The optimizer settings are the same for all three models in both
replay conditions.
We initialize fresh optimizer states and restart the learning-rate schedule
for each run. Attention and expert matrices are trained by Muon with base
learning rate \(0.0025\) and the matrix-shape scaling of
\Cref{app:experimental-parameterization}. The remaining trainable
parameters are trained by AdamW with base learning rate \(7.5\times10^{-5}\),
including token embeddings, the output head, normalization parameters,
attention residual queries and gains, and the path gates of \ours{}.
Muon uses momentum \(0.95\), Nesterov updates, and five Newton--Schulz
iterations; AdamW uses \(\beta=(0.9,0.95)\) and \(\epsilon=10^{-8}\).
The decoupled weight decay is \(0.1\) for attention and expert
matrices, token embeddings, the output head, and path gates. Normalization
parameters, biases, and attention residual queries and gains have no
weight decay. We clip the global gradient norm to \(1.0\).
The schedule starts with 47 steps of linear warmup, followed by cosine
annealing for the remaining 430 steps. The cosine scheduler uses
\(T_{\max}=477\) and \(\eta_{\min}=10^{-5}\), so the run ends before
the scheduled minimum.

\paragraph{Compute reference.}
At the \(32\)k context length used for adaptation, the estimated forward
FLOPs per token for \ours{} are approximately \(0.84\times\) those of
the 500M MoE transformer and \(0.43\times\) those of the 1B dense
transformer. In these estimates we count attention projections,
causal attention,
and MLPs, assume balanced token assignments across columns, and exclude
routers, attention residuals, normalization, and the output head.
Since all models are trained on the same number of tokens, the same ratios apply
to these forward FLOP totals over each adaptation run. We report them
as a reference for computational cost; the adaptation and forgetting
results compare models trained on equal token budgets.

\paragraph{Evaluation.}
We evaluate the starting checkpoint and the final checkpoint at step 477
on identical held-out sequences. For PG-19, we use the first
\(32{,}768\) tokens of each of the 77 test books with at least this length.
For OpenWebMath, AlgebraicStack, and German, we use the first 200
non-overlapping \(32{,}768\)-token sequences of each test stream,
allowing sequences to contain multiple documents. To measure forgetting,
we use the first \(32{,}768\) tokens of each of 200 DCLM-baseline
validation documents with at least this length. This DCLM evaluation set is
separate from the extension and replay data. We average next-token
cross-entropy over all prediction targets, excluding the first token of
each sequence. The evaluation sequences are the same for all models,
and there are no
parameter updates during evaluation. Adaptation is the initial loss minus
the final loss on the target corpus; forgetting is the final loss minus
the initial loss on DCLM. In the averages in \Cref{fig:domain-adaptation}
and \Cref{tab:domain-adaptation-results}, each of the four target domains
has equal weight.

\paragraph{Adaptation with replay of the pretraining corpus.}
In the runs without replay, all training tokens come from the target
corpus. We repeat these runs with a mixture containing approximately
\(75\%\) target-domain tokens and \(25\%\) DCLM-baseline tokens, keeping
the total budget and all other training settings unchanged. The DCLM
sequences are taken from those already used during context extension.
The target sequences are a subset of those used in the corresponding
run without replay. We shuffle the combined sequences before training;
the \(25\%\) share is approximate because the data are sampled in whole
sequences.
The results are in \Cref{fig:domain-adaptation} (faded marks) and
\Cref{tab:domain-adaptation-results}. Averaged over the
four domains, replay reduces forgetting by about \(70\%\) for each model,
while reducing the adaptation gain by about \(6\%\).
With replay, \ours{} and the MoE transformer have mean gains of
\(0.687\) and \(0.687\), respectively, and mean forgetting of
\(0.073\) and \(0.076\). \ours{} has less forgetting on each of the four
corpora. The 1B dense transformer has a mean gain of \(0.696\) and
mean forgetting of \(0.087\).
The per-domain results and averages for both replay settings, including
the values in \Cref{sec:domain-adaptation}, are listed in
\Cref{tab:domain-adaptation-results}.

\begin{table}[htbp]
    \centering
    \caption{\textbf{Loss before and after domain adaptation.}
        Target-corpus and DCLM held-out losses at \(32\)k, before adaptation
        and after \(0.5\)B training tokens. Gain is target loss before minus
        after; forgetting is DCLM loss after minus before, corresponding to
        \Cref{fig:domain-adaptation}.
        In the average, the four domains have equal weight.
        Differences and averages are computed before rounding to three
    decimal places.}
    \label{tab:domain-adaptation-results}
    \begin{tabular}{llrrrrrr}
        \toprule
        & & \multicolumn{3}{c}{Target-corpus loss}
        & \multicolumn{3}{c}{DCLM loss} \\
        \cmidrule(lr){3-5}\cmidrule(lr){6-8}
        Domain & Model & Before & After & Gain $\uparrow$
        & Before & After & Forgetting $\downarrow$ \\
        \midrule
        \multicolumn{8}{l}{\textit{No replay}} \\
        Books & 500M \ours{} & 2.732 & 2.548 & 0.184 & 2.779 & 2.996 & 0.217 \\
        & 500M MoE & 2.765 & 2.582 & 0.182 & 2.810 & 3.039 & 0.229 \\
        & 1B dense & 2.719 & 2.538 & 0.181 & 2.779 & 3.032 & 0.252 \\
        \addlinespace
        Math & 500M \ours{} & 2.417 & 2.084 & 0.332 & 2.779 & 2.885 & 0.106 \\
        & 500M MoE & 2.449 & 2.111 & 0.338 & 2.810 & 2.920 & 0.110 \\
        & 1B dense & 2.412 & 2.075 & 0.338 & 2.779 & 2.914 & 0.134 \\
        \addlinespace
        Code & 500M \ours{} & 1.789 & 1.084 & 0.705 & 2.779 & 3.073 & 0.294 \\
        & 500M MoE & 1.813 & 1.102 & 0.712 & 2.810 & 3.106 & 0.296 \\
        & 1B dense & 1.720 & 1.031 & 0.688 & 2.779 & 3.164 & 0.385 \\
        \addlinespace
        German & 500M \ours{} & 4.171 & 2.465 & 1.706 & 2.779 & 3.176 & 0.398 \\
        & 500M MoE & 4.224 & 2.521 & 1.704 & 2.810 & 3.221 & 0.410 \\
        & 1B dense & 4.164 & 2.431 & 1.733 & 2.779 & 3.273 & 0.494 \\
        \addlinespace
        \midrule
        Average & 500M \ours{} & 2.777 & 2.045 & 0.732 & 2.779 &
        3.033 & 0.254 \\
        & 500M MoE & 2.813 & 2.079 & 0.734 & 2.810 & 3.072 & 0.261 \\
        & 1B dense & 2.754 & 2.019 & 0.735 & 2.779 & 3.096 & 0.316 \\
        \midrule
        \multicolumn{8}{l}{\textit{\(25\%\) replay}} \\
        Books & 500M \ours{} & 2.732 & 2.573 & 0.158 & 2.779 & 2.828 & 0.049 \\
        & 500M MoE & 2.765 & 2.607 & 0.158 & 2.810 & 2.863 & 0.053 \\
        & 1B dense & 2.719 & 2.560 & 0.159 & 2.779 & 2.837 & 0.058 \\
        \addlinespace
        Math & 500M \ours{} & 2.417 & 2.113 & 0.304 & 2.779 & 2.832 & 0.053 \\
        & 500M MoE & 2.449 & 2.140 & 0.309 & 2.810 & 2.867 & 0.057 \\
        & 1B dense & 2.412 & 2.101 & 0.311 & 2.779 & 2.842 & 0.063 \\
        \addlinespace
        Code & 500M \ours{} & 1.789 & 1.131 & 0.658 & 2.779 & 2.867 & 0.088 \\
        & 500M MoE & 1.813 & 1.146 & 0.668 & 2.810 & 2.901 & 0.091 \\
        & 1B dense & 1.720 & 1.074 & 0.646 & 2.779 & 2.885 & 0.106 \\
        \addlinespace
        German & 500M \ours{} & 4.171 & 2.546 & 1.626 & 2.779 & 2.879 & 0.101 \\
        & 500M MoE & 4.224 & 2.611 & 1.614 & 2.810 & 2.913 & 0.102 \\
        & 1B dense & 4.164 & 2.497 & 1.667 & 2.779 & 2.899 & 0.120 \\
        \addlinespace
        \midrule
        Average & 500M \ours{} & 2.777 & 2.091 & 0.687 & 2.779 &
        2.851 & 0.073 \\
        & 500M MoE & 2.813 & 2.126 & 0.687 & 2.810 & 2.886 & 0.076 \\
        & 1B dense & 2.754 & 2.058 & 0.696 & 2.779 & 2.866 & 0.087 \\
        \bottomrule
    \end{tabular}
\end{table}
\subsection{Long Context Extension Settings}
\label{app:long-context-extension-settings}

This subsection gives the details of the long context extension experiments in
\Cref{sec:long-context-extension}. We report results through \(64\)k at
the 100M, 200M, 500M, and 1B scales.

\begin{figure}[t]
    \centering
    \pgfplotstableread[col sep=comma]{figures/exps/raw_data/derived/flops_length_notes.csv}\flnotes
\def\flcurves{figures/exps/raw_data/derived/flops_length_curves.csv}
\colorlet{flChrome}{black!35}
\colorlet{flBand}{expCortex!16!white}
\pgfplotsset{
  fl ours line/.style={expCortex, line width=0.9pt},
  fl ours mark/.style={mark=fstar, mark size=2.8pt, mark options={fill=expCortex, draw=white, line width=0.3pt, solid}},
  fl base line/.style={expMoe, line width=0.6pt},
  fl base mark/.style={mark=diamond*, mark size=1.9pt, mark options={fill=white, draw=expMoe, line width=0.5pt, solid}},
  fl ours legend/.style={fl ours line, fl ours mark},
  fl base legend/.style={fl base line, fl base mark},
}
\newcommand{\flseries}[5]{%
  \addplot[#4, mark=none, forget plot]
    table[col sep=comma, x=x, y=#3, discard if not={scale}{#1}, discard if not={model}{#2}, discard if not={span}{run}] {\flcurves};
  \addplot[#4, densely dashed, mark=none, forget plot]
    table[col sep=comma, x=x, y=#3, discard if not={scale}{#1}, discard if not={model}{#2}, discard if not={span}{ext}] {\flcurves};
  \addplot[only marks, #5, forget plot]
    table[col sep=comma, x=x, y=#3, discard if not={scale}{#1}, discard if not={model}{#2}, discard if not={span}{mark}] {\flcurves};
}
\newcommand{\fltop}[1]{%
  \addplot[draw=none, fill=flBand, forget plot]
    table[col sep=comma, x=x, y=gflops, discard if not={scale}{#1}, discard if not={span}{band}] {\flcurves} \closedcycle;
  \flseries{#1}{base}{gflops}{fl base line}{fl base mark}
  \flseries{#1}{cortex}{gflops}{fl ours line}{fl ours mark}
}
\newcommand{\flbottom}[2]{%
  \pgfplotstablegetelem{#2}{limit}\of\flnotes\let\fllimit\pgfplotsretval
  \pgfplotstablegetelem{#2}{label_limit}\of\flnotes\let\fllimitlabel\pgfplotsretval
  \pgfplotstablegetelem{#2}{x_last}\of\flnotes\let\flxlast\pgfplotsretval
  \pgfplotstablegetelem{#2}{ratio_last}\of\flnotes\let\flrlast\pgfplotsretval
  \pgfplotstablegetelem{#2}{label_last}\of\flnotes\let\fllastlabel\pgfplotsretval
  \expsnow{\noexpand\draw[expCortex!55!white, dotted, line width=0.6pt] (axis cs:4,\fllimit) -- (axis cs:1024,\fllimit);}
  \expsnow{\noexpand\node[exps annot, text=expCortex!85!black, anchor=north west, inner xsep=1.2pt, inner ysep=2.4pt] at (axis cs:4,\fllimit) {\fllimitlabel};}
  \flseries{#1}{base}{ratio}{fl base line}{fl base mark}
  \flseries{#1}{cortex}{ratio}{fl ours line}{fl ours mark}
  \expsnow{\noexpand\node[exps annot, text=expCortex!85!black, anchor=south west, inner sep=1.5pt] at (axis cs:\flxlast,\flrlast) {\fllastlabel};}
}
\begin{tikzpicture}
\begin{groupplot}[exps,
  group style={group size=4 by 2, horizontal sep=0.34cm, vertical sep=0.28cm,
               yticklabels at=edge left, xticklabels at=edge bottom},
  scale only axis, width=0.178\textwidth,
  xmode=log, log basis x=2, xmin=3.3, xmax=1300,
  xtick={4,16,64,256,1024}, xticklabels={4k,16k,64k,256k,1M}, minor xtick={},
  tick label style={font=\sffamily\fontsize{6}{7}\selectfont, black!55},
  axis line style={flChrome, line width=0.4pt}, tick style={flChrome, line width=0.4pt},
  grid style={black!7, line width=0.3pt},
  every axis title/.append style={font=\sffamily\scriptsize, yshift=-3pt},
]
\nextgroupplot[title={100M}, height=0.23\textwidth, ymode=log, ymin=0.1, ymax=110,
               ytick={0.1,1,10,100}, yticklabels={0.1,1,10,100}]  \fltop{100m}
\nextgroupplot[title={200M}, height=0.23\textwidth, ymode=log, ymin=0.1, ymax=110, ytick={0.1,1,10,100}]  \fltop{200m}
\nextgroupplot[title={500M}, height=0.23\textwidth, ymode=log, ymin=0.1, ymax=110, ytick={0.1,1,10,100}]  \fltop{500m}
\nextgroupplot[title={1B},   height=0.23\textwidth, ymode=log, ymin=0.1, ymax=110, ytick={0.1,1,10,100}]  \fltop{1b}
\nextgroupplot[height=0.15\textwidth, ymin=0.69, ymax=1.04, ytick={0.7,0.8,0.9,1.0},
               yticklabels={0.7$\times$,0.8$\times$,0.9$\times$,1.0$\times$}]  \flbottom{100m}{0}
\nextgroupplot[height=0.15\textwidth, ymin=0.69, ymax=1.04, ytick={0.7,0.8,0.9,1.0}]  \flbottom{200m}{1}
\nextgroupplot[height=0.15\textwidth, ymin=0.69, ymax=1.04, ytick={0.7,0.8,0.9,1.0}]  \flbottom{500m}{2}
\nextgroupplot[height=0.15\textwidth, ymin=0.69, ymax=1.04, ytick={0.7,0.8,0.9,1.0},
               legend to name=exps:flops-length, legend columns=3,
               legend style={font=\sffamily\fontsize{6.5}{8}\selectfont, /tikz/every even column/.append style={column sep=0.9em}}]
  \flbottom{1b}{3}
  \addlegendimage{fl ours legend}\addlegendentry{\bfseries\ours{}}
  \addlegendimage{fl base legend}\addlegendentry{MoE transformer (dense within $0.2\%$)}
  \addlegendimage{black!55, densely dashed, line width=0.6pt}\addlegendentry{dashed: beyond the trained lengths}
\end{groupplot}
\node[rotate=90, anchor=south, font=\sffamily\scriptsize, text=expInk, align=center] at ([xshift=-1pt]group c1r1.outer west) {Forward FLOPs\\per token (GFLOPs)};
\node[rotate=90, anchor=south, font=\sffamily\scriptsize, text=expInk, align=center] at ([xshift=-1pt]group c1r2.outer west) {Relative to\\the MoE};
\node[below=0.26cm, font=\sffamily\scriptsize, text=expInk] at ($(group c1r2.south)!0.5!(group c4r2.south)$) {Context length $S$};
\node[below=0.56cm] at ($(group c1r2.south)!0.5!(group c4r2.south)$) {\pgfplotslegendfromname{exps:flops-length}};
\end{tikzpicture}
    \caption{\textbf{Forward FLOPs per token against the context length.}
        Top: the forward FLOPs per token of \ours{} and of the baselines at
        all scales, with the accounting of \Cref{app:configurations}. The
        baseline curve is that of the MoE transformer. The dense transformer
        has the same FLOPs at 100M and 500M; at 200M and 1B, because of the
        rounding of the MoE expert widths, the two differ by less than
        $0.2\%$ over the plotted lengths. Bottom: the FLOPs per token of
        \ours{} relative to the MoE transformer. The markers are the context
        lengths of \Cref{fig:long-context-extension}, and the same formula is
        plotted up to $1$M tokens as dashed curves. The projections and MLPs
        of \ours{} cost $4$--$8\%$ more than those of the baselines, and its
        attention costs $0.75$--$0.81\times$ as much. The share of attention
        in the total grows with $S$, so the relative cost of \ours{} decreases
        from $0.98$--$0.99\times$ at $4$k to $0.80$--$0.84\times$ at $64$k,
    and tends to the attention ratio as $S \to \infty$ (dotted).}
    \label{fig:flops-vs-length}
\end{figure}
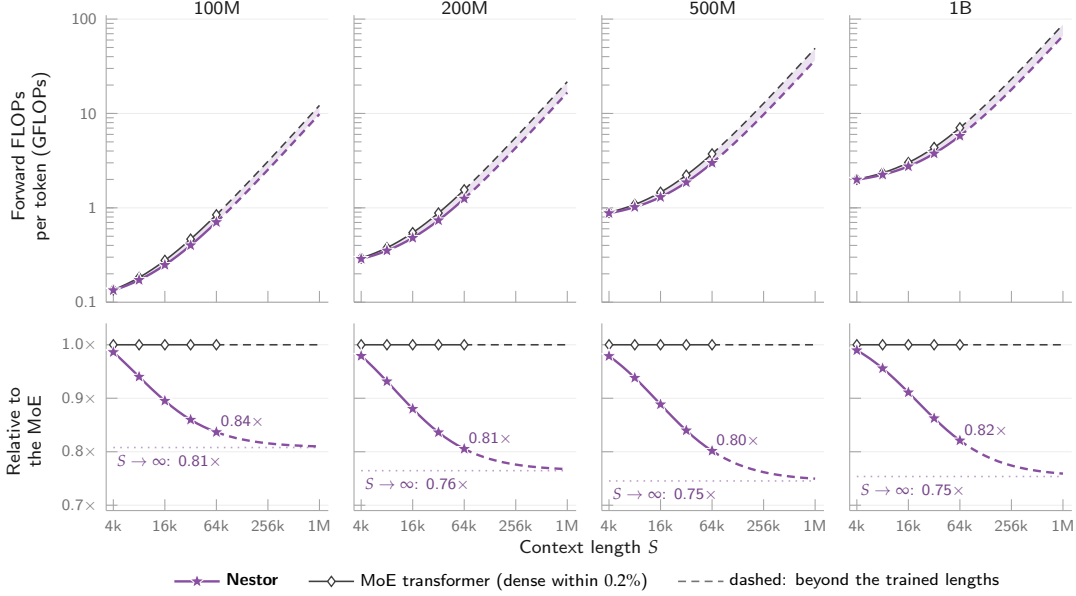

\paragraph{Starting checkpoints.}
We extend the pretrained dense transformer, MoE transformer, and \ours{}
of \Cref{sec:pretraining} at all four scales. These models were pretrained
at context length \(4\)k with rotary base \(10^{4}\); their \(4\)k held-out
losses are the first column of \Cref{tab:long-context-losses}. Each target
length is a separate run from the \(4\)k checkpoint, and no run continues
from a model extended to a shorter length. At 200M, \ours{} was pretrained
with two seeds, and every setting is run from both checkpoints, with a
data order for the extension run that depends on the seed; the
baselines, and all
models at the other scales, have one run per length.

\paragraph{Data.}
For a target length \(S\), the extension corpus consists of DCLM-baseline
documents of at least \(S\) tokens, taken from a part of the corpus that
was not used in pretraining. Documents are separated by an end-of-text
token and concatenated, and the training sequences are consecutive
\(S\)-token windows of this stream, so a sequence may cross a document
boundary; causal attention is not reset at those boundaries. At a given
length, all models are trained on the same sequences in the same order (data
sampler seed \(0\)). The held-out set at length \(S\) consists of the first
\(S\) tokens of each of \(200\) documents of at least \(S\) tokens
(\(100\) at \(64\)k), taken from a third part of the corpus that is
disjoint from both the pretraining and the extension data. The losses at
\(4\)k are those of \Cref{sec:pretraining}, on its held-out set.

\paragraph{Extension at equal tokens.}
Each run has \(477\) optimizer steps with \(1{,}048{,}576\) tokens
per step (\(128\), \(64\), \(32\), and \(16\) sequences per step at \(8\)k,
\(16\)k, \(32\)k, and \(64\)k), which gives \(500{,}170{,}752\) tokens,
rounded to \(0.5\)B in the text. The rotary base is raised from \(10^{4}\)
to \(5 \times 10^{5}\), \(10^{6}\), \(2 \times 10^{6}\), and \(5 \times
10^{6}\) at the four lengths, in training and in evaluation. All parameters
are trained, including the routers. The optimizer, the weight decay, and
the gradient clipping are those of \Cref{app:domain-adaptation-settings}:
Muon with peak learning rate \(0.0025\) and AdamW with peak learning rate
\(7.5 \times 10^{-5}\), a quarter of the pretraining peaks, with fresh
optimizer states, \(47\) steps of linear warmup (\(10\%\)), and cosine
annealing over the remaining steps.

\paragraph{Extension at equal FLOPs.}
Let \(r(S)\) be the forward FLOPs per token of \ours{} relative to the MoE
transformer at length \(S\), with the accounting of
\Cref{app:configurations}; its values are given under the panels of
\Cref{fig:long-context-extension} and listed in
\Cref{tab:long-context-budgets}, and \(1 / r(S)\) is the measured
counterpart of the multiplier \(R(S)\) of \Cref{sec:long-context-extension}.
The equal-FLOP runs of \ours{} are trained on \(0.5\mathrm{B} / r(S)\) tokens at
the same tokens per step, that is for \(T = \lceil 477 / r(S) \rceil\)
steps, with a warmup of \(10\%\) of \(T\) and the cosine schedule spread
over the \(T\) steps, and with both peak learning rates multiplied by
\(r(S)\), so that the product of the number of steps and the peak learning
rate is that of the equal-token run. At the unscaled peak learning rate,
the losses of the same runs were higher by \(0.002\) to \(0.014\) nats at
500M and 1B, with a difference that increases with \(S\). In the
setting with new documents,
the sequences are drawn from the full extension corpus, so the additional
\(T - 477\) steps use documents that are not in the equal-token run. In
the setting with repeated documents, the corpus is restricted to the
sequences used in the equal-token run, extended by \((T - 477)\) steps'
worth of sequences drawn from that set without replacement, so that no
sequence appears more than twice, and the run is one shuffled pass over
this multiset; the two copies of a repeated sequence are therefore at
random positions of the run, at a mean distance of \(T / 3\) steps. The two
settings have the same number of steps, tokens, and FLOPs, and differ only
in the data.

\begin{table}[ht]
    \centering
    \caption{FLOP ratio \(r(S)\) of \ours{} to the MoE transformer at each
        context length, and, in parentheses, the number of optimizer steps
        of the equal-FLOP runs, \(\lceil 477 / r(S) \rceil\). The
    equal-token runs have \(477\) steps.}
    \label{tab:long-context-budgets}
    \begin{tabular}{lcccc}
        \toprule
        Scale & \(8\)k & \(16\)k & \(32\)k & \(64\)k \\
        \midrule
        100M & 0.940 (508) & 0.895 (533) & 0.860 (555) & 0.837 (570) \\
        200M & 0.933 (512) & 0.881 (542) & 0.837 (570) & 0.806 (592) \\
        500M & 0.938 (509) & 0.888 (537) & 0.840 (568) & 0.802 (595) \\
        1B & 0.956 (499) & 0.911 (524) & 0.863 (553) & 0.821 (581) \\
        \bottomrule
    \end{tabular}
\end{table}

\paragraph{Evaluation.}
Each run is evaluated at its final step on the held-out sequences of its
length, with the rotary base of the run and in full precision. We average
the next-token cross-entropy over all prediction targets of all sequences,
excluding the first token of each sequence. The same sequences are used for
every model and setting at a length. In \Cref{fig:long-context-extension},
we plot these losses minus the loss of the MoE transformer of the same
scale at the same length; at 200M, the two seeds are joined by a whisker.
The losses are listed in \Cref{tab:long-context-losses}.

\paragraph{Results.}
The loss of the dense transformer is higher than that of the MoE
transformer by \(0.13\) to \(0.19\)
nats at every length and scale. At equal tokens, the margin of \ours{} over
the MoE transformer is about the same as in pretraining at every
length: \(-0.028\) to
\(-0.030\) nats at 100M, \(-0.012\) to \(-0.015\) at 200M, \(-0.031\) to
\(-0.035\) at 500M, and \(-0.017\) to \(-0.025\) at 1B. At equal FLOPs the
margin increases with the context length. With new documents it is
\(-0.047\), \(-0.037\), \(-0.061\), and \(-0.058\) nats at \(64\)k at the
four scales; with repeated documents, \(-0.037\), \(-0.023\), \(-0.047\),
and \(-0.033\). With repeated documents, the losses are within \(0.006\)
nats of those with new documents at \(8\)k and \(16\)k at every
scale; at \(32\)k and
\(64\)k, the losses with new documents are lower by \(0.004\) to \(0.025\) nats.

\begin{table}[ht]
    \centering
    \footnotesize
    \setlength{\tabcolsep}{3.5pt}
    \caption{Held-out loss at each context length, before extension
        (\(4\)k) and after extension to \(8\)k, \(16\)k, \(32\)k, and
        \(64\)k, for the runs of \Cref{fig:long-context-extension}. The
        three settings of \ours{} start from the same \(4\)k checkpoint;
        repeated and new refer to the documents of the equal-FLOP runs. At
        200M, \ours{} is given for both seeds (seed 42 / seed 43); the
    baselines and the other scales have one seed.}
    \label{tab:long-context-losses}
    \begin{tabular}{lccccc}
        \toprule
        Model & \(4\)k & \(8\)k & \(16\)k & \(32\)k & \(64\)k \\
        \midrule
        \multicolumn{6}{l}{\textit{100M}} \\
        Dense transformer & 3.578 & 3.511 & 3.495 & 3.501 & 3.490 \\
        MoE transformer & 3.386 & 3.338 & 3.330 & 3.330 & 3.325 \\
        \ours{}, equal tokens & 3.356 & 3.311 & 3.302 & 3.302 & 3.296 \\
        \ours{}, equal FLOPs, repeated & -- & 3.309 & 3.298 & 3.297 & 3.287 \\
        \ours{}, equal FLOPs, new & -- & 3.307 & 3.295 & 3.293 & 3.277 \\
        \addlinespace
        \multicolumn{6}{l}{\textit{200M}} \\
        Dense transformer & 3.293 & 3.256 & 3.248 & 3.244 & 3.237 \\
        MoE transformer & 3.108 & 3.096 & 3.095 & 3.084 & 3.080 \\
        \ours{}, equal tokens & 3.092/3.088 & 3.081/3.078 &
        3.084/3.083 & 3.070/3.068 & 3.066/3.063 \\
        \ours{}, equal FLOPs, repeated & -- & 3.079/3.076 &
        3.075/3.073 & 3.061/3.058 & 3.056/3.051 \\
        \ours{}, equal FLOPs, new & -- & 3.077/3.073 & 3.074/3.071 &
        3.056/3.054 & 3.043/3.042 \\
        \addlinespace
        \multicolumn{6}{l}{\textit{500M}} \\
        Dense transformer & 2.953 & 2.971 & 2.970 & 2.952 & 2.948 \\
        MoE transformer & 2.793 & 2.827 & 2.842 & 2.810 & 2.805 \\
        \ours{}, equal tokens & 2.758 & 2.792 & 2.811 & 2.779 & 2.773 \\
        \ours{}, equal FLOPs, repeated & -- & 2.789 & 2.802 & 2.766 & 2.757 \\
        \ours{}, equal FLOPs, new & -- & 2.787 & 2.800 & 2.758 & 2.744 \\
        \addlinespace
        \multicolumn{6}{l}{\textit{1B}} \\
        Dense transformer & 2.743 & 2.797 & 2.810 & 2.779 & 2.780 \\
        MoE transformer & 2.592 & 2.633 & 2.683 & 2.634 & 2.632 \\
        \ours{}, equal tokens & 2.571 & 2.616 & 2.661 & 2.612 & 2.607 \\
        \ours{}, equal FLOPs, repeated & -- & 2.616 & 2.655 & 2.603 & 2.599 \\
        \ours{}, equal FLOPs, new & -- & 2.610 & 2.650 & 2.591 & 2.574 \\
        \bottomrule
    \end{tabular}
\end{table}

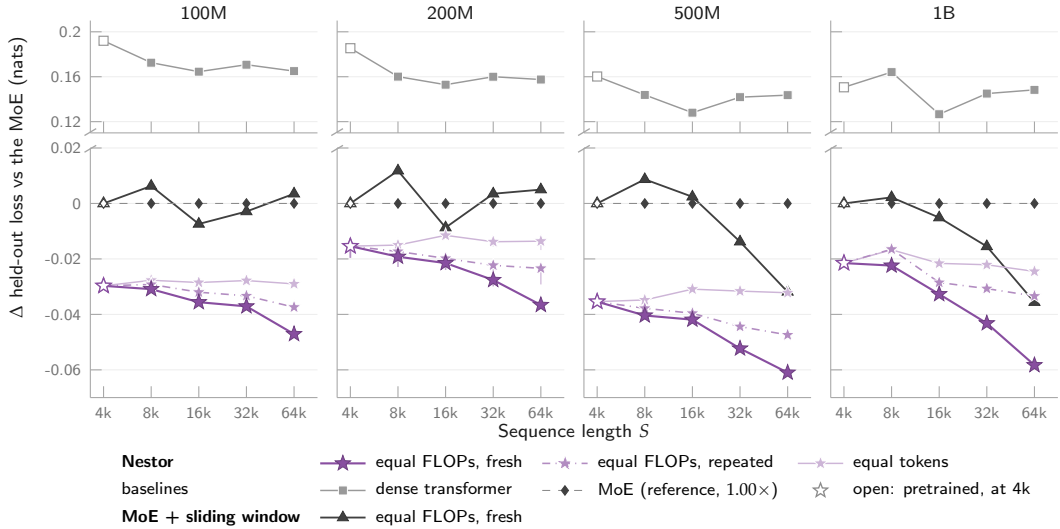
\begin{figure}[ht]
    \centering
    \def\cptprefix{cptswa}
\def\cptyticks{0.02,0,-0.02,-0.04,-0.06}
\def\cptextra#1{\cptseries{#1}{swafm_isoflop_lr}{cpt swa fresh line}{cpt swa fresh mark}{cpt swa fresh open}}
\def\cptlegend{%
  \cptlegenddefault
  \addlegendimage{empty legend}\addlegendentry{\bfseries MoE + sliding window}
  \addlegendimage{cpt swa fresh legend}\addlegendentry{equal FLOPs, fresh}}
\providecommand{\cptprefix}{cpt}
\providecommand{\cptdense}{dense}
\providecommand{\cptdenselabel}{dense transformer}
\providecommand{\cptextra}[1]{}
\providecommand{\cptyticks}{0,-0.02,-0.04,-0.06}
\providecommand{\cptlegend}{\cptlegenddefault}
\providecommand{\cptlegendcols}{4}
\providecommand{\cptbottomheight}{0.2\textwidth}  %
\expsnow{\noexpand\pgfplotstableread[col sep=comma]{figures/exps/raw_data/derived/\cptprefix_axis.csv}\noexpand\cptaxis}
\edef\cptdeltafile{figures/exps/raw_data/derived/\cptprefix_delta.csv}
\edef\cpttwinsfile{figures/exps/raw_data/derived/\cptprefix_twins.csv}
\newcommand{\cptlim}[3]{\pgfplotstablegetelem{#1}{#2}\of\cptaxis\let#3\pgfplotsretval}
\cptlim{0}{ylo}{\cptyloA}\cptlim{0}{yhi}{\cptyhiA}\cptlim{0}{tlo}{\cpttloA}\cptlim{0}{thi}{\cptthiA}\cptlim{0}{xmax}{\cptxmaxA}
\cptlim{1}{ylo}{\cptyloB}\cptlim{1}{yhi}{\cptyhiB}\cptlim{1}{tlo}{\cpttloB}\cptlim{1}{thi}{\cptthiB}\cptlim{1}{xmax}{\cptxmaxB}
\cptlim{2}{ylo}{\cptyloC}\cptlim{2}{yhi}{\cptyhiC}\cptlim{2}{tlo}{\cpttloC}\cptlim{2}{thi}{\cptthiC}\cptlim{2}{xmax}{\cptxmaxC}
\cptlim{3}{ylo}{\cptyloD}\cptlim{3}{yhi}{\cptyhiD}\cptlim{3}{tlo}{\cpttloD}\cptlim{3}{thi}{\cptthiD}\cptlim{3}{xmax}{\cptxmaxD}
\newcommand{\cptseries}[5]{%
  \addplot[#3, mark=none, forget plot]
    table[col sep=comma, x=x, y=delta, discard if not={scale}{#1}, discard if not={series}{#2}]
    {\cptdeltafile};
  \addplot[only marks, #4, forget plot]
    table[col sep=comma, x=x, y=delta, discard if not={scale}{#1}, discard if not={series}{#2}, discard if={x}{4}]
    {\cptdeltafile};
  \addplot[only marks, #5, forget plot]
    table[col sep=comma, x=x, y=delta, discard if not={scale}{#1}, discard if not={series}{#2}, discard if not={x}{4}]
    {\cptdeltafile};
}
\newcommand{\cpttwin}[3]{%
  \addplot[#3, opacity=0.5, line width=0.6pt, mark=none, forget plot]
    table[col sep=comma, x=x, y=y, discard if not={scale}{#1}, discard if not={series}{#2}]
    {\cpttwinsfile};
}
\colorlet{cptChrome}{black!35}
\colorlet{cptOursA}{expCortex}
\colorlet{cptOursB}{expCortexMid}    %
\colorlet{cptOursC}{expCortexLight}
\providecolor{cptSwaA}{named}{expMoe}
\colorlet{cptSwaB}{cptSwaA!65!white}
\colorlet{cptSwaC}{cptSwaA!42!white}
\pgfplotsset{
  cpt fresh line/.style={cptOursA, line width=0.8pt},
  cpt fresh mark/.style={mark=fstar, mark size=3pt, mark options={fill=cptOursA, draw=cptOursA!80!black, line width=0.3pt}},
  cpt fresh open/.style={mark=fstar, mark size=3pt, mark options={fill=white, draw=cptOursA, line width=0.5pt}},
  cpt repeat line/.style={cptOursB, dash dot, line width=0.6pt},
  cpt repeat mark/.style={mark=fstar, mark size=2.8pt, mark options={fill=cptOursB, draw=white, line width=0.3pt}},
  cpt repeat open/.style={mark=fstar, mark size=2.8pt, mark options={fill=white, draw=cptOursB, line width=0.5pt}},
  cpt isotok line/.style={cptOursC, line width=0.55pt},
  cpt isotok mark/.style={mark=fstar, mark size=2.8pt, mark options={fill=cptOursC, draw=white, line width=0.3pt}},
  cpt isotok open/.style={mark=fstar, mark size=2.8pt, mark options={fill=white, draw=cptOursC, line width=0.5pt}},
  cpt moe line/.style={black!45, dashed, line width=0.4pt},
  cpt moe mark/.style={mark=diamond*, mark size=1.8pt, mark options={fill=expMoe, draw=expMoe, line width=0.3pt, solid}},
  cpt moe open/.style={mark=diamond*, mark size=1.8pt, mark options={fill=white, draw=expMoe, line width=0.45pt, solid}},
  cpt dense line/.style={black!40, line width=0.55pt},
  cpt dense mark/.style={mark=square*, mark size=1.7pt, mark options={fill=black!40, draw=white, line width=0.3pt}},
  cpt dense open/.style={mark=square*, mark size=1.7pt, mark options={fill=white, draw=black!40, line width=0.45pt}},
  cpt swa fresh line/.style={cptSwaA, line width=0.7pt},
  cpt swa fresh mark/.style={mark=triangle*, mark size=2.4pt, mark options={fill=cptSwaA, draw=cptSwaA!80!black, line width=0.3pt, solid}},
  cpt swa fresh open/.style={mark=triangle*, mark size=2.4pt, mark options={fill=white, draw=cptSwaA, line width=0.5pt, solid}},
  cpt swa repeat line/.style={cptSwaB, dash dot, line width=0.55pt},
  cpt swa repeat mark/.style={mark=triangle*, mark size=2.2pt, mark options={fill=cptSwaB, draw=white, line width=0.3pt, solid}},
  cpt swa repeat open/.style={mark=triangle*, mark size=2.2pt, mark options={fill=white, draw=cptSwaB, line width=0.5pt, solid}},
  cpt swa isotok line/.style={cptSwaC, line width=0.5pt},
  cpt swa isotok mark/.style={mark=triangle*, mark size=2.2pt, mark options={fill=cptSwaC, draw=white, line width=0.3pt, solid}},
  cpt swa isotok open/.style={mark=triangle*, mark size=2.2pt, mark options={fill=white, draw=cptSwaC, line width=0.5pt, solid}},
  cpt fresh legend/.style={cpt fresh line, cpt fresh mark},
  cpt repeat legend/.style={cpt repeat line, cpt repeat mark},
  cpt isotok legend/.style={cpt isotok line, cpt isotok mark},
  cpt moe legend/.style={cpt moe line, cpt moe mark},
  cpt dense legend/.style={cpt dense line, cpt dense mark},
  cpt swa fresh legend/.style={cpt swa fresh line, cpt swa fresh mark},
  cpt swa repeat legend/.style={cpt swa repeat line, cpt swa repeat mark},
  cpt swa isotok legend/.style={cpt swa isotok line, cpt swa isotok mark},
  cpt open legend/.style={only marks, mark=fstar, mark size=2.8pt, mark options={fill=white, draw=black!45, line width=0.5pt}},
}
\newcommand{\cpttop}[1]{\cptseries{#1}{\cptdense}{cpt dense line}{cpt dense mark}{cpt dense open}}
\newcommand{\cpttwins}[1]{\cpttwin{#1}{isotok}{cptOursC}\cpttwin{#1}{isoflop_lr}{cptOursA}\cpttwin{#1}{isoflop_ep_lr2}{cptOursB}}
\newcommand{\cptbottom}[1]{%
  \cptextra{#1}%
  \cptseries{#1}{moe}{cpt moe line}{cpt moe mark}{cpt moe open}
  \cptseries{#1}{isotok}{cpt isotok line}{cpt isotok mark}{cpt isotok open}
  \cptseries{#1}{isoflop_ep_lr2}{cpt repeat line}{cpt repeat mark}{cpt repeat open}
  \cptseries{#1}{isoflop_lr}{cpt fresh line}{cpt fresh mark}{cpt fresh open}
}
\def\cptlegenddefault{%
  \addlegendimage{empty legend}\addlegendentry{\bfseries\ours{}}
  \addlegendimage{cpt fresh legend}\addlegendentry{equal FLOPs, fresh}
  \addlegendimage{cpt repeat legend}\addlegendentry{equal FLOPs, repeated}
  \addlegendimage{cpt isotok legend}\addlegendentry{equal tokens}
  \addlegendimage{empty legend}\addlegendentry{baselines}
  \addlegendimage{cpt dense legend}\addlegendentry{\cptdenselabel}
  \addlegendimage{cpt moe legend}\addlegendentry{MoE (reference, $1.00\times$)}
  \addlegendimage{cpt open legend}\addlegendentry{open: pretrained, at 4k}}
\newcommand{\cptbreak}[1]{%
  \draw[cptChrome, line width=0.5pt] ([xshift=-2.2pt, yshift=-1pt]group c#1r1.south west) -- ([xshift=2.2pt, yshift=1.6pt]group c#1r1.south west);
  \draw[cptChrome, line width=0.5pt] ([xshift=-2.2pt, yshift=-1.6pt]group c#1r2.north west) -- ([xshift=2.2pt, yshift=1pt]group c#1r2.north west);
}
\begin{tikzpicture}
\begin{groupplot}[exps,
  group style={group size=4 by 2, horizontal sep=0.26cm, vertical sep=0.2cm,
               yticklabels at=edge left, xticklabels at=edge bottom},
  scale only axis, width=0.182\textwidth,
  xmode=log, log basis x=2, xmin=3.3,
  xtick={4,8,16,32,64}, xticklabels={4k,8k,16k,32k,64k}, minor xtick={},
  tick label style={font=\sffamily\fontsize{6}{7}\selectfont, black!55},
  axis line style={cptChrome, line width=0.4pt}, tick style={cptChrome, line width=0.4pt},
  grid style={black!7, line width=0.3pt},
  every axis title/.append style={font=\sffamily\scriptsize, yshift=-3pt},
]
\nextgroupplot[title={100M}, height=0.09\textwidth, xmax=\cptxmaxA, ymin=\cpttloA, ymax=\cptthiA, ytick={0.12,0.16,0.20}]  \cpttop{100m}
\nextgroupplot[title={200M}, height=0.09\textwidth, xmax=\cptxmaxB, ymin=\cpttloB, ymax=\cptthiB, ytick={0.12,0.16,0.20}]  \cpttop{200m}
\nextgroupplot[title={500M}, height=0.09\textwidth, xmax=\cptxmaxC, ymin=\cpttloC, ymax=\cptthiC, ytick={0.12,0.16,0.20}]  \cpttop{500m}
\nextgroupplot[title={1B},   height=0.09\textwidth, xmax=\cptxmaxD, ymin=\cpttloD, ymax=\cptthiD, ytick={0.12,0.16,0.20}]  \cpttop{1b}
\nextgroupplot[height=\cptbottomheight, xmax=\cptxmaxA, ymin=\cptyloA, ymax=\cptyhiA, ytick={\cptyticks}]   \cptbottom{100m}
\nextgroupplot[height=\cptbottomheight, xmax=\cptxmaxB, ymin=\cptyloB, ymax=\cptyhiB, ytick={\cptyticks}]   \cpttwins{200m}\cptbottom{200m}
\nextgroupplot[height=\cptbottomheight, xmax=\cptxmaxC, ymin=\cptyloC, ymax=\cptyhiC, ytick={\cptyticks}]   \cptbottom{500m}
\nextgroupplot[height=\cptbottomheight, xmax=\cptxmaxD, ymin=\cptyloD, ymax=\cptyhiD, ytick={\cptyticks},
               legend to name=exps:cpt\cptprefix, legend columns=\cptlegendcols,
               legend style={font=\sffamily\fontsize{6.5}{8}\selectfont, /tikz/every even column/.append style={column sep=0.5em},
                             /tikz/every odd column/.append style={column sep=0.15em}}]
  \cptbottom{1b}
  \cptlegend
\end{groupplot}
\cptbreak{1}\cptbreak{2}\cptbreak{3}\cptbreak{4}
\node[rotate=90, anchor=south, font=\sffamily\scriptsize, text=expInk] at ([xshift=-1pt]$(group c1r1.outer west)!0.5!(group c1r2.outer west)$) {$\Delta$ held-out loss vs the MoE (nats)};
\node[below=0.22cm, font=\sffamily\scriptsize, text=expInk] at ($(group c1r2.south)!0.5!(group c4r2.south)$) {Sequence length $S$};
\node[below=0.5cm] at ($(group c1r2.south)!0.5!(group c4r2.south)$) {\pgfplotslegendfromname{exps:cpt\cptprefix}};
\end{tikzpicture}
\def\cptprefix{cpt}\def\cptyticks{0,-0.02,-0.04,-0.06}\def\cptextra#1{}\def\cptlegend{\cptlegenddefault}\def\cptlegendcols{4}
    \caption{\textbf{Long context extension with a sliding window baseline.}
        The panels of \Cref{fig:long-context-extension} with one more
        baseline in the bottom row (triangles): the MoE transformer with
        sliding window attention in as many layers as needed to match
        the FLOPs per token of \ours{}, extended at equal FLOPs with new
        documents. It starts from the \(4\)k checkpoint of the MoE
    transformer, so both have the same open marker at \(4\)k.}
    \label{fig:long-context-extension-swa}
\end{figure}

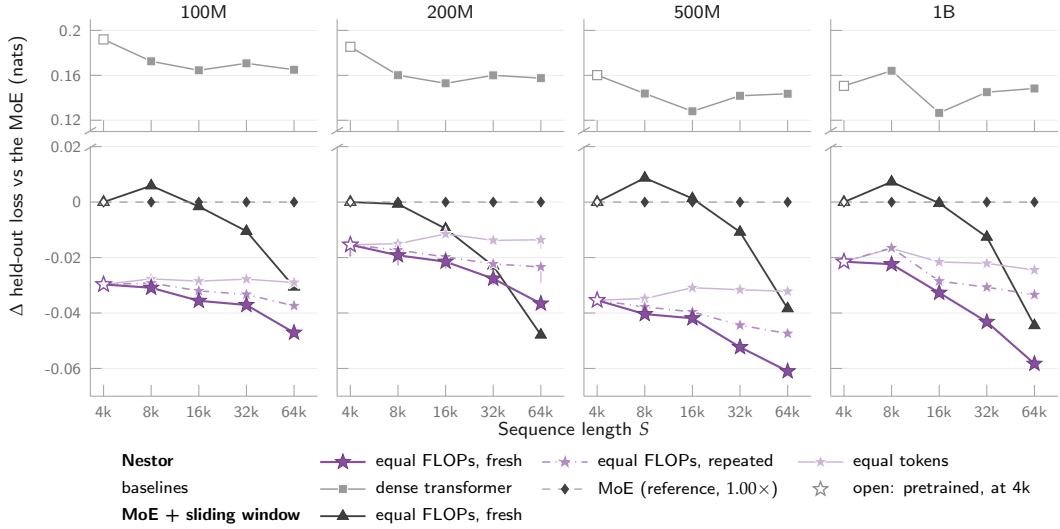
\begin{figure}[ht]
    \centering
    \def\cptprefix{cptswap2}
\def\cptyticks{0.02,0,-0.02,-0.04,-0.06}
\def\cptextra#1{\cptseries{#1}{swaperiod2_isoflop_lr}{cpt swa fresh line}{cpt swa fresh mark}{cpt swa fresh open}}
\def\cptlegend{%
  \cptlegenddefault
  \addlegendimage{empty legend}\addlegendentry{\bfseries MoE + sliding window}
  \addlegendimage{cpt swa fresh legend}\addlegendentry{equal FLOPs, fresh}}
\providecommand{\cptprefix}{cpt}
\providecommand{\cptdense}{dense}
\providecommand{\cptdenselabel}{dense transformer}
\providecommand{\cptextra}[1]{}
\providecommand{\cptyticks}{0,-0.02,-0.04,-0.06}
\providecommand{\cptlegend}{\cptlegenddefault}
\providecommand{\cptlegendcols}{4}
\providecommand{\cptbottomheight}{0.2\textwidth}  %
\expsnow{\noexpand\pgfplotstableread[col sep=comma]{figures/exps/raw_data/derived/\cptprefix_axis.csv}\noexpand\cptaxis}
\edef\cptdeltafile{figures/exps/raw_data/derived/\cptprefix_delta.csv}
\edef\cpttwinsfile{figures/exps/raw_data/derived/\cptprefix_twins.csv}
\newcommand{\cptlim}[3]{\pgfplotstablegetelem{#1}{#2}\of\cptaxis\let#3\pgfplotsretval}
\cptlim{0}{ylo}{\cptyloA}\cptlim{0}{yhi}{\cptyhiA}\cptlim{0}{tlo}{\cpttloA}\cptlim{0}{thi}{\cptthiA}\cptlim{0}{xmax}{\cptxmaxA}
\cptlim{1}{ylo}{\cptyloB}\cptlim{1}{yhi}{\cptyhiB}\cptlim{1}{tlo}{\cpttloB}\cptlim{1}{thi}{\cptthiB}\cptlim{1}{xmax}{\cptxmaxB}
\cptlim{2}{ylo}{\cptyloC}\cptlim{2}{yhi}{\cptyhiC}\cptlim{2}{tlo}{\cpttloC}\cptlim{2}{thi}{\cptthiC}\cptlim{2}{xmax}{\cptxmaxC}
\cptlim{3}{ylo}{\cptyloD}\cptlim{3}{yhi}{\cptyhiD}\cptlim{3}{tlo}{\cpttloD}\cptlim{3}{thi}{\cptthiD}\cptlim{3}{xmax}{\cptxmaxD}
\newcommand{\cptseries}[5]{%
  \addplot[#3, mark=none, forget plot]
    table[col sep=comma, x=x, y=delta, discard if not={scale}{#1}, discard if not={series}{#2}]
    {\cptdeltafile};
  \addplot[only marks, #4, forget plot]
    table[col sep=comma, x=x, y=delta, discard if not={scale}{#1}, discard if not={series}{#2}, discard if={x}{4}]
    {\cptdeltafile};
  \addplot[only marks, #5, forget plot]
    table[col sep=comma, x=x, y=delta, discard if not={scale}{#1}, discard if not={series}{#2}, discard if not={x}{4}]
    {\cptdeltafile};
}
\newcommand{\cpttwin}[3]{%
  \addplot[#3, opacity=0.5, line width=0.6pt, mark=none, forget plot]
    table[col sep=comma, x=x, y=y, discard if not={scale}{#1}, discard if not={series}{#2}]
    {\cpttwinsfile};
}
\colorlet{cptChrome}{black!35}
\colorlet{cptOursA}{expCortex}
\colorlet{cptOursB}{expCortexMid}    %
\colorlet{cptOursC}{expCortexLight}
\providecolor{cptSwaA}{named}{expMoe}
\colorlet{cptSwaB}{cptSwaA!65!white}
\colorlet{cptSwaC}{cptSwaA!42!white}
\pgfplotsset{
  cpt fresh line/.style={cptOursA, line width=0.8pt},
  cpt fresh mark/.style={mark=fstar, mark size=3pt, mark options={fill=cptOursA, draw=cptOursA!80!black, line width=0.3pt}},
  cpt fresh open/.style={mark=fstar, mark size=3pt, mark options={fill=white, draw=cptOursA, line width=0.5pt}},
  cpt repeat line/.style={cptOursB, dash dot, line width=0.6pt},
  cpt repeat mark/.style={mark=fstar, mark size=2.8pt, mark options={fill=cptOursB, draw=white, line width=0.3pt}},
  cpt repeat open/.style={mark=fstar, mark size=2.8pt, mark options={fill=white, draw=cptOursB, line width=0.5pt}},
  cpt isotok line/.style={cptOursC, line width=0.55pt},
  cpt isotok mark/.style={mark=fstar, mark size=2.8pt, mark options={fill=cptOursC, draw=white, line width=0.3pt}},
  cpt isotok open/.style={mark=fstar, mark size=2.8pt, mark options={fill=white, draw=cptOursC, line width=0.5pt}},
  cpt moe line/.style={black!45, dashed, line width=0.4pt},
  cpt moe mark/.style={mark=diamond*, mark size=1.8pt, mark options={fill=expMoe, draw=expMoe, line width=0.3pt, solid}},
  cpt moe open/.style={mark=diamond*, mark size=1.8pt, mark options={fill=white, draw=expMoe, line width=0.45pt, solid}},
  cpt dense line/.style={black!40, line width=0.55pt},
  cpt dense mark/.style={mark=square*, mark size=1.7pt, mark options={fill=black!40, draw=white, line width=0.3pt}},
  cpt dense open/.style={mark=square*, mark size=1.7pt, mark options={fill=white, draw=black!40, line width=0.45pt}},
  cpt swa fresh line/.style={cptSwaA, line width=0.7pt},
  cpt swa fresh mark/.style={mark=triangle*, mark size=2.4pt, mark options={fill=cptSwaA, draw=cptSwaA!80!black, line width=0.3pt, solid}},
  cpt swa fresh open/.style={mark=triangle*, mark size=2.4pt, mark options={fill=white, draw=cptSwaA, line width=0.5pt, solid}},
  cpt swa repeat line/.style={cptSwaB, dash dot, line width=0.55pt},
  cpt swa repeat mark/.style={mark=triangle*, mark size=2.2pt, mark options={fill=cptSwaB, draw=white, line width=0.3pt, solid}},
  cpt swa repeat open/.style={mark=triangle*, mark size=2.2pt, mark options={fill=white, draw=cptSwaB, line width=0.5pt, solid}},
  cpt swa isotok line/.style={cptSwaC, line width=0.5pt},
  cpt swa isotok mark/.style={mark=triangle*, mark size=2.2pt, mark options={fill=cptSwaC, draw=white, line width=0.3pt, solid}},
  cpt swa isotok open/.style={mark=triangle*, mark size=2.2pt, mark options={fill=white, draw=cptSwaC, line width=0.5pt, solid}},
  cpt fresh legend/.style={cpt fresh line, cpt fresh mark},
  cpt repeat legend/.style={cpt repeat line, cpt repeat mark},
  cpt isotok legend/.style={cpt isotok line, cpt isotok mark},
  cpt moe legend/.style={cpt moe line, cpt moe mark},
  cpt dense legend/.style={cpt dense line, cpt dense mark},
  cpt swa fresh legend/.style={cpt swa fresh line, cpt swa fresh mark},
  cpt swa repeat legend/.style={cpt swa repeat line, cpt swa repeat mark},
  cpt swa isotok legend/.style={cpt swa isotok line, cpt swa isotok mark},
  cpt open legend/.style={only marks, mark=fstar, mark size=2.8pt, mark options={fill=white, draw=black!45, line width=0.5pt}},
}
\newcommand{\cpttop}[1]{\cptseries{#1}{\cptdense}{cpt dense line}{cpt dense mark}{cpt dense open}}
\newcommand{\cpttwins}[1]{\cpttwin{#1}{isotok}{cptOursC}\cpttwin{#1}{isoflop_lr}{cptOursA}\cpttwin{#1}{isoflop_ep_lr2}{cptOursB}}
\newcommand{\cptbottom}[1]{%
  \cptextra{#1}%
  \cptseries{#1}{moe}{cpt moe line}{cpt moe mark}{cpt moe open}
  \cptseries{#1}{isotok}{cpt isotok line}{cpt isotok mark}{cpt isotok open}
  \cptseries{#1}{isoflop_ep_lr2}{cpt repeat line}{cpt repeat mark}{cpt repeat open}
  \cptseries{#1}{isoflop_lr}{cpt fresh line}{cpt fresh mark}{cpt fresh open}
}
\def\cptlegenddefault{%
  \addlegendimage{empty legend}\addlegendentry{\bfseries\ours{}}
  \addlegendimage{cpt fresh legend}\addlegendentry{equal FLOPs, fresh}
  \addlegendimage{cpt repeat legend}\addlegendentry{equal FLOPs, repeated}
  \addlegendimage{cpt isotok legend}\addlegendentry{equal tokens}
  \addlegendimage{empty legend}\addlegendentry{baselines}
  \addlegendimage{cpt dense legend}\addlegendentry{\cptdenselabel}
  \addlegendimage{cpt moe legend}\addlegendentry{MoE (reference, $1.00\times$)}
  \addlegendimage{cpt open legend}\addlegendentry{open: pretrained, at 4k}}
\newcommand{\cptbreak}[1]{%
  \draw[cptChrome, line width=0.5pt] ([xshift=-2.2pt, yshift=-1pt]group c#1r1.south west) -- ([xshift=2.2pt, yshift=1.6pt]group c#1r1.south west);
  \draw[cptChrome, line width=0.5pt] ([xshift=-2.2pt, yshift=-1.6pt]group c#1r2.north west) -- ([xshift=2.2pt, yshift=1pt]group c#1r2.north west);
}
\begin{tikzpicture}
\begin{groupplot}[exps,
  group style={group size=4 by 2, horizontal sep=0.26cm, vertical sep=0.2cm,
               yticklabels at=edge left, xticklabels at=edge bottom},
  scale only axis, width=0.182\textwidth,
  xmode=log, log basis x=2, xmin=3.3,
  xtick={4,8,16,32,64}, xticklabels={4k,8k,16k,32k,64k}, minor xtick={},
  tick label style={font=\sffamily\fontsize{6}{7}\selectfont, black!55},
  axis line style={cptChrome, line width=0.4pt}, tick style={cptChrome, line width=0.4pt},
  grid style={black!7, line width=0.3pt},
  every axis title/.append style={font=\sffamily\scriptsize, yshift=-3pt},
]
\nextgroupplot[title={100M}, height=0.09\textwidth, xmax=\cptxmaxA, ymin=\cpttloA, ymax=\cptthiA, ytick={0.12,0.16,0.20}]  \cpttop{100m}
\nextgroupplot[title={200M}, height=0.09\textwidth, xmax=\cptxmaxB, ymin=\cpttloB, ymax=\cptthiB, ytick={0.12,0.16,0.20}]  \cpttop{200m}
\nextgroupplot[title={500M}, height=0.09\textwidth, xmax=\cptxmaxC, ymin=\cpttloC, ymax=\cptthiC, ytick={0.12,0.16,0.20}]  \cpttop{500m}
\nextgroupplot[title={1B},   height=0.09\textwidth, xmax=\cptxmaxD, ymin=\cpttloD, ymax=\cptthiD, ytick={0.12,0.16,0.20}]  \cpttop{1b}
\nextgroupplot[height=\cptbottomheight, xmax=\cptxmaxA, ymin=\cptyloA, ymax=\cptyhiA, ytick={\cptyticks}]   \cptbottom{100m}
\nextgroupplot[height=\cptbottomheight, xmax=\cptxmaxB, ymin=\cptyloB, ymax=\cptyhiB, ytick={\cptyticks}]   \cpttwins{200m}\cptbottom{200m}
\nextgroupplot[height=\cptbottomheight, xmax=\cptxmaxC, ymin=\cptyloC, ymax=\cptyhiC, ytick={\cptyticks}]   \cptbottom{500m}
\nextgroupplot[height=\cptbottomheight, xmax=\cptxmaxD, ymin=\cptyloD, ymax=\cptyhiD, ytick={\cptyticks},
               legend to name=exps:cpt\cptprefix, legend columns=\cptlegendcols,
               legend style={font=\sffamily\fontsize{6.5}{8}\selectfont, /tikz/every even column/.append style={column sep=0.5em},
                             /tikz/every odd column/.append style={column sep=0.15em}}]
  \cptbottom{1b}
  \cptlegend
\end{groupplot}
\cptbreak{1}\cptbreak{2}\cptbreak{3}\cptbreak{4}
\node[rotate=90, anchor=south, font=\sffamily\scriptsize, text=expInk] at ([xshift=-1pt]$(group c1r1.outer west)!0.5!(group c1r2.outer west)$) {$\Delta$ held-out loss vs the MoE (nats)};
\node[below=0.22cm, font=\sffamily\scriptsize, text=expInk] at ($(group c1r2.south)!0.5!(group c4r2.south)$) {Sequence length $S$};
\node[below=0.5cm] at ($(group c1r2.south)!0.5!(group c4r2.south)$) {\pgfplotslegendfromname{exps:cpt\cptprefix}};
\end{tikzpicture}
\def\cptprefix{cpt}\def\cptyticks{0,-0.02,-0.04,-0.06}\def\cptextra#1{}\def\cptlegend{\cptlegenddefault}\def\cptlegendcols{4}
    \caption{\textbf{Long context extension with a sliding window baseline,
        one global layer in two.} As
        \Cref{fig:long-context-extension-swa}, with a different baseline
        (triangles): the MoE transformer with sliding window attention of
        window \(4096\) in every other layer, starting from the first, and
        global attention elsewhere. Its FLOPs per token are
        \(0.94\)--\(0.96\times\) those of the MoE transformer at \(8\)k and
        \(0.61\)--\(0.67\times\) at \(64\)k, lower than those of \ours{}
        from \(16\)k on. At equal FLOPs it is therefore trained on more tokens
        than \ours{}: \(0.75\)--\(0.82\)B against \(0.60\)--\(0.62\)B at
    \(64\)k.}
    \label{fig:long-context-extension-swa-period2}
\end{figure}

\begin{figure}[ht]
    \centering
    \def\cptprefix{cptswap6}
\def\cptyticks{0.02,0,-0.02,-0.04,-0.06,-0.08}
\def\cptbottomheight{0.289\textwidth}
\def\cptextra#1{\cptseries{#1}{swaperiod6_isoflop_lr}{cpt swa fresh line}{cpt swa fresh mark}{cpt swa fresh open}}
\def\cptlegend{%
  \cptlegenddefault
  \addlegendimage{empty legend}\addlegendentry{\bfseries MoE + sliding window}
  \addlegendimage{cpt swa fresh legend}\addlegendentry{equal FLOPs, fresh}}
\providecommand{\cptprefix}{cpt}
\providecommand{\cptdense}{dense}
\providecommand{\cptdenselabel}{dense transformer}
\providecommand{\cptextra}[1]{}
\providecommand{\cptyticks}{0,-0.02,-0.04,-0.06}
\providecommand{\cptlegend}{\cptlegenddefault}
\providecommand{\cptlegendcols}{4}
\providecommand{\cptbottomheight}{0.2\textwidth}  %
\expsnow{\noexpand\pgfplotstableread[col sep=comma]{figures/exps/raw_data/derived/\cptprefix_axis.csv}\noexpand\cptaxis}
\edef\cptdeltafile{figures/exps/raw_data/derived/\cptprefix_delta.csv}
\edef\cpttwinsfile{figures/exps/raw_data/derived/\cptprefix_twins.csv}
\newcommand{\cptlim}[3]{\pgfplotstablegetelem{#1}{#2}\of\cptaxis\let#3\pgfplotsretval}
\cptlim{0}{ylo}{\cptyloA}\cptlim{0}{yhi}{\cptyhiA}\cptlim{0}{tlo}{\cpttloA}\cptlim{0}{thi}{\cptthiA}\cptlim{0}{xmax}{\cptxmaxA}
\cptlim{1}{ylo}{\cptyloB}\cptlim{1}{yhi}{\cptyhiB}\cptlim{1}{tlo}{\cpttloB}\cptlim{1}{thi}{\cptthiB}\cptlim{1}{xmax}{\cptxmaxB}
\cptlim{2}{ylo}{\cptyloC}\cptlim{2}{yhi}{\cptyhiC}\cptlim{2}{tlo}{\cpttloC}\cptlim{2}{thi}{\cptthiC}\cptlim{2}{xmax}{\cptxmaxC}
\cptlim{3}{ylo}{\cptyloD}\cptlim{3}{yhi}{\cptyhiD}\cptlim{3}{tlo}{\cpttloD}\cptlim{3}{thi}{\cptthiD}\cptlim{3}{xmax}{\cptxmaxD}
\newcommand{\cptseries}[5]{%
  \addplot[#3, mark=none, forget plot]
    table[col sep=comma, x=x, y=delta, discard if not={scale}{#1}, discard if not={series}{#2}]
    {\cptdeltafile};
  \addplot[only marks, #4, forget plot]
    table[col sep=comma, x=x, y=delta, discard if not={scale}{#1}, discard if not={series}{#2}, discard if={x}{4}]
    {\cptdeltafile};
  \addplot[only marks, #5, forget plot]
    table[col sep=comma, x=x, y=delta, discard if not={scale}{#1}, discard if not={series}{#2}, discard if not={x}{4}]
    {\cptdeltafile};
}
\newcommand{\cpttwin}[3]{%
  \addplot[#3, opacity=0.5, line width=0.6pt, mark=none, forget plot]
    table[col sep=comma, x=x, y=y, discard if not={scale}{#1}, discard if not={series}{#2}]
    {\cpttwinsfile};
}
\colorlet{cptChrome}{black!35}
\colorlet{cptOursA}{expCortex}
\colorlet{cptOursB}{expCortexMid}    %
\colorlet{cptOursC}{expCortexLight}
\providecolor{cptSwaA}{named}{expMoe}
\colorlet{cptSwaB}{cptSwaA!65!white}
\colorlet{cptSwaC}{cptSwaA!42!white}
\pgfplotsset{
  cpt fresh line/.style={cptOursA, line width=0.8pt},
  cpt fresh mark/.style={mark=fstar, mark size=3pt, mark options={fill=cptOursA, draw=cptOursA!80!black, line width=0.3pt}},
  cpt fresh open/.style={mark=fstar, mark size=3pt, mark options={fill=white, draw=cptOursA, line width=0.5pt}},
  cpt repeat line/.style={cptOursB, dash dot, line width=0.6pt},
  cpt repeat mark/.style={mark=fstar, mark size=2.8pt, mark options={fill=cptOursB, draw=white, line width=0.3pt}},
  cpt repeat open/.style={mark=fstar, mark size=2.8pt, mark options={fill=white, draw=cptOursB, line width=0.5pt}},
  cpt isotok line/.style={cptOursC, line width=0.55pt},
  cpt isotok mark/.style={mark=fstar, mark size=2.8pt, mark options={fill=cptOursC, draw=white, line width=0.3pt}},
  cpt isotok open/.style={mark=fstar, mark size=2.8pt, mark options={fill=white, draw=cptOursC, line width=0.5pt}},
  cpt moe line/.style={black!45, dashed, line width=0.4pt},
  cpt moe mark/.style={mark=diamond*, mark size=1.8pt, mark options={fill=expMoe, draw=expMoe, line width=0.3pt, solid}},
  cpt moe open/.style={mark=diamond*, mark size=1.8pt, mark options={fill=white, draw=expMoe, line width=0.45pt, solid}},
  cpt dense line/.style={black!40, line width=0.55pt},
  cpt dense mark/.style={mark=square*, mark size=1.7pt, mark options={fill=black!40, draw=white, line width=0.3pt}},
  cpt dense open/.style={mark=square*, mark size=1.7pt, mark options={fill=white, draw=black!40, line width=0.45pt}},
  cpt swa fresh line/.style={cptSwaA, line width=0.7pt},
  cpt swa fresh mark/.style={mark=triangle*, mark size=2.4pt, mark options={fill=cptSwaA, draw=cptSwaA!80!black, line width=0.3pt, solid}},
  cpt swa fresh open/.style={mark=triangle*, mark size=2.4pt, mark options={fill=white, draw=cptSwaA, line width=0.5pt, solid}},
  cpt swa repeat line/.style={cptSwaB, dash dot, line width=0.55pt},
  cpt swa repeat mark/.style={mark=triangle*, mark size=2.2pt, mark options={fill=cptSwaB, draw=white, line width=0.3pt, solid}},
  cpt swa repeat open/.style={mark=triangle*, mark size=2.2pt, mark options={fill=white, draw=cptSwaB, line width=0.5pt, solid}},
  cpt swa isotok line/.style={cptSwaC, line width=0.5pt},
  cpt swa isotok mark/.style={mark=triangle*, mark size=2.2pt, mark options={fill=cptSwaC, draw=white, line width=0.3pt, solid}},
  cpt swa isotok open/.style={mark=triangle*, mark size=2.2pt, mark options={fill=white, draw=cptSwaC, line width=0.5pt, solid}},
  cpt fresh legend/.style={cpt fresh line, cpt fresh mark},
  cpt repeat legend/.style={cpt repeat line, cpt repeat mark},
  cpt isotok legend/.style={cpt isotok line, cpt isotok mark},
  cpt moe legend/.style={cpt moe line, cpt moe mark},
  cpt dense legend/.style={cpt dense line, cpt dense mark},
  cpt swa fresh legend/.style={cpt swa fresh line, cpt swa fresh mark},
  cpt swa repeat legend/.style={cpt swa repeat line, cpt swa repeat mark},
  cpt swa isotok legend/.style={cpt swa isotok line, cpt swa isotok mark},
  cpt open legend/.style={only marks, mark=fstar, mark size=2.8pt, mark options={fill=white, draw=black!45, line width=0.5pt}},
}
\newcommand{\cpttop}[1]{\cptseries{#1}{\cptdense}{cpt dense line}{cpt dense mark}{cpt dense open}}
\newcommand{\cpttwins}[1]{\cpttwin{#1}{isotok}{cptOursC}\cpttwin{#1}{isoflop_lr}{cptOursA}\cpttwin{#1}{isoflop_ep_lr2}{cptOursB}}
\newcommand{\cptbottom}[1]{%
  \cptextra{#1}%
  \cptseries{#1}{moe}{cpt moe line}{cpt moe mark}{cpt moe open}
  \cptseries{#1}{isotok}{cpt isotok line}{cpt isotok mark}{cpt isotok open}
  \cptseries{#1}{isoflop_ep_lr2}{cpt repeat line}{cpt repeat mark}{cpt repeat open}
  \cptseries{#1}{isoflop_lr}{cpt fresh line}{cpt fresh mark}{cpt fresh open}
}
\def\cptlegenddefault{%
  \addlegendimage{empty legend}\addlegendentry{\bfseries\ours{}}
  \addlegendimage{cpt fresh legend}\addlegendentry{equal FLOPs, fresh}
  \addlegendimage{cpt repeat legend}\addlegendentry{equal FLOPs, repeated}
  \addlegendimage{cpt isotok legend}\addlegendentry{equal tokens}
  \addlegendimage{empty legend}\addlegendentry{baselines}
  \addlegendimage{cpt dense legend}\addlegendentry{\cptdenselabel}
  \addlegendimage{cpt moe legend}\addlegendentry{MoE (reference, $1.00\times$)}
  \addlegendimage{cpt open legend}\addlegendentry{open: pretrained, at 4k}}
\newcommand{\cptbreak}[1]{%
  \draw[cptChrome, line width=0.5pt] ([xshift=-2.2pt, yshift=-1pt]group c#1r1.south west) -- ([xshift=2.2pt, yshift=1.6pt]group c#1r1.south west);
  \draw[cptChrome, line width=0.5pt] ([xshift=-2.2pt, yshift=-1.6pt]group c#1r2.north west) -- ([xshift=2.2pt, yshift=1pt]group c#1r2.north west);
}
\begin{tikzpicture}
\begin{groupplot}[exps,
  group style={group size=4 by 2, horizontal sep=0.26cm, vertical sep=0.2cm,
               yticklabels at=edge left, xticklabels at=edge bottom},
  scale only axis, width=0.182\textwidth,
  xmode=log, log basis x=2, xmin=3.3,
  xtick={4,8,16,32,64}, xticklabels={4k,8k,16k,32k,64k}, minor xtick={},
  tick label style={font=\sffamily\fontsize{6}{7}\selectfont, black!55},
  axis line style={cptChrome, line width=0.4pt}, tick style={cptChrome, line width=0.4pt},
  grid style={black!7, line width=0.3pt},
  every axis title/.append style={font=\sffamily\scriptsize, yshift=-3pt},
]
\nextgroupplot[title={100M}, height=0.09\textwidth, xmax=\cptxmaxA, ymin=\cpttloA, ymax=\cptthiA, ytick={0.12,0.16,0.20}]  \cpttop{100m}
\nextgroupplot[title={200M}, height=0.09\textwidth, xmax=\cptxmaxB, ymin=\cpttloB, ymax=\cptthiB, ytick={0.12,0.16,0.20}]  \cpttop{200m}
\nextgroupplot[title={500M}, height=0.09\textwidth, xmax=\cptxmaxC, ymin=\cpttloC, ymax=\cptthiC, ytick={0.12,0.16,0.20}]  \cpttop{500m}
\nextgroupplot[title={1B},   height=0.09\textwidth, xmax=\cptxmaxD, ymin=\cpttloD, ymax=\cptthiD, ytick={0.12,0.16,0.20}]  \cpttop{1b}
\nextgroupplot[height=\cptbottomheight, xmax=\cptxmaxA, ymin=\cptyloA, ymax=\cptyhiA, ytick={\cptyticks}]   \cptbottom{100m}
\nextgroupplot[height=\cptbottomheight, xmax=\cptxmaxB, ymin=\cptyloB, ymax=\cptyhiB, ytick={\cptyticks}]   \cpttwins{200m}\cptbottom{200m}
\nextgroupplot[height=\cptbottomheight, xmax=\cptxmaxC, ymin=\cptyloC, ymax=\cptyhiC, ytick={\cptyticks}]   \cptbottom{500m}
\nextgroupplot[height=\cptbottomheight, xmax=\cptxmaxD, ymin=\cptyloD, ymax=\cptyhiD, ytick={\cptyticks},
               legend to name=exps:cpt\cptprefix, legend columns=\cptlegendcols,
               legend style={font=\sffamily\fontsize{6.5}{8}\selectfont, /tikz/every even column/.append style={column sep=0.5em},
                             /tikz/every odd column/.append style={column sep=0.15em}}]
  \cptbottom{1b}
  \cptlegend
\end{groupplot}
\cptbreak{1}\cptbreak{2}\cptbreak{3}\cptbreak{4}
\node[rotate=90, anchor=south, font=\sffamily\scriptsize, text=expInk] at ([xshift=-1pt]$(group c1r1.outer west)!0.5!(group c1r2.outer west)$) {$\Delta$ held-out loss vs the MoE (nats)};
\node[below=0.22cm, font=\sffamily\scriptsize, text=expInk] at ($(group c1r2.south)!0.5!(group c4r2.south)$) {Sequence length $S$};
\node[below=0.5cm] at ($(group c1r2.south)!0.5!(group c4r2.south)$) {\pgfplotslegendfromname{exps:cpt\cptprefix}};
\end{tikzpicture}
\def\cptprefix{cpt}\def\cptyticks{0,-0.02,-0.04,-0.06}\def\cptextra#1{}\def\cptlegend{\cptlegenddefault}\def\cptlegendcols{4}
\def\cptbottomheight{0.2\textwidth}
    \caption{\textbf{Long context extension with a sliding window baseline,
        one global layer in six.} As
        \Cref{fig:long-context-extension-swa}, with a different baseline
        (triangles): the MoE transformer with sliding window attention of
        window \(4096\) in five of every six layers and global attention in
        the sixth. Its FLOPs per token are \(0.89\)--\(0.94\times\) those of
        the MoE transformer at \(8\)k and \(0.34\)--\(0.43\times\) at
        \(64\)k, so at equal FLOPs it is trained on \(1.15\)--\(1.49\)B tokens
        at \(64\)k, against \(0.60\)--\(0.62\)B for \ours{}. The \(64\)k
        extension corpus has \(1.2\)B tokens; at 100M, 200M, and 500M,
        \(17\%\), \(20\%\), and \(2\%\) of the tokens of the \(64\)k run
        are therefore seen a second time. The vertical range of the bottom
    row is larger than in \Cref{fig:long-context-extension-swa}.}
    \label{fig:long-context-extension-swa-period6}
\end{figure}
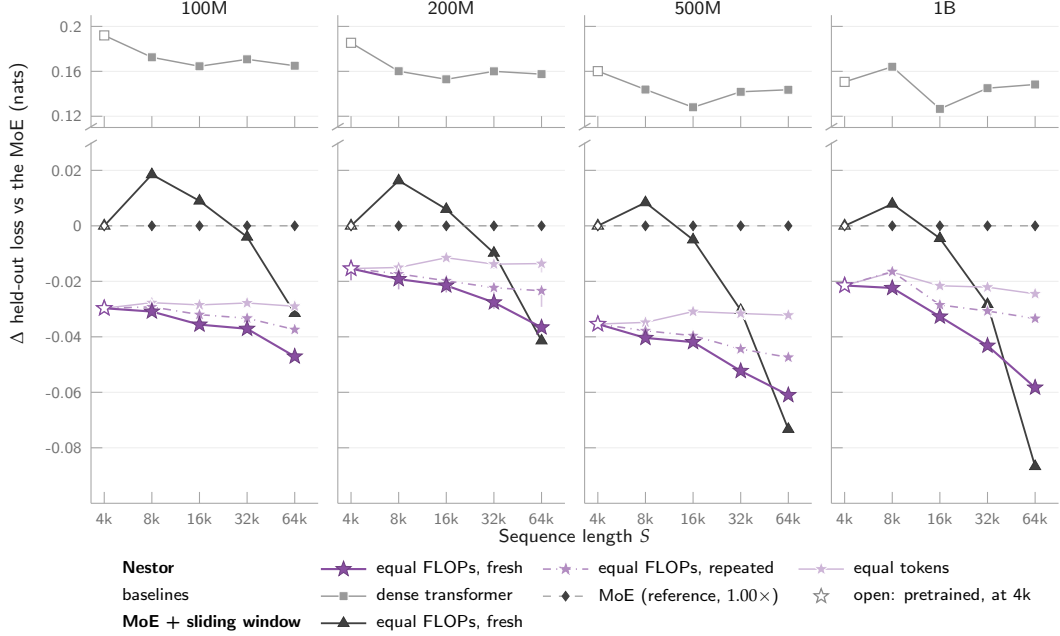
\subsection{Routing Statistics and Routing Counterfactuals}
\label{app:routing-analysis}

This section describes how balanced the trained routers of \ours{} and of
the MoE transformer are, what information each level of the routing
contains, and how much the held-out loss depends on each level. For the
first two questions we record the routing decisions of the \(500\)M models
of \Cref{sec:domain-adaptation}; for the third we evaluate the pretrained
models of \Cref{sec:pretraining} with modified routing.

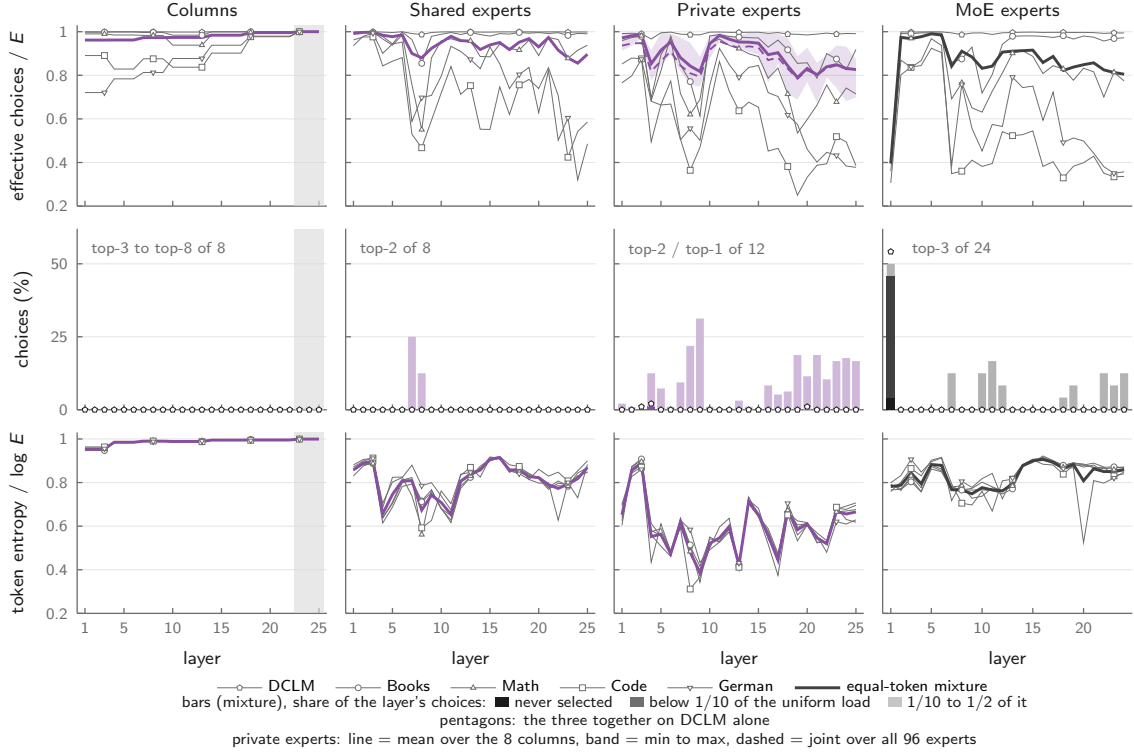
\begin{figure}[t]
    \centering
    \newcommand{\rtdatadir}{figures/exps/raw_data/routing}%
    \providecommand{\rtdatadir}{data}
\def\rtdepth{\rtdatadir/balance_depth.csv}
\newcommand{\rtdepthline}[5]{%
  \addplot[#5] table[col sep=comma, x expr=\thisrow{layer}+1, y=#4, discard if not={arm}{#1},
    discard if not={family}{#2}, discard if not={view}{#3}] {\rtdepth};}
\newcommand{\rtdepthviews}[4]{%
  \rtdepthline{#1}{#2}{dclm}{#3}{rt thin, rt dclm}
  \rtdepthline{#1}{#2}{pg19}{#3}{rt thin, rt pg19}
  \rtdepthline{#1}{#2}{owm}{#3}{rt thin, rt owm}
  \rtdepthline{#1}{#2}{algstack}{#3}{rt thin, rt algstack}
  \rtdepthline{#1}{#2}{fw2de}{#3}{rt thin, rt fw2de}
  \rtdepthline{#1}{#2}{mix}{#3}{#4, line width=1.1pt, mark=none}
}
\newcommand{\rtdepthbars}[3]{%
  \addplot[ybar stacked, bar width=0.8, fill=expInk, draw=none] table[col sep=comma, x expr=\thisrow{layer}+1, y expr=100*\thisrow{share_zero},
    discard if not={arm}{#1}, discard if not={family}{#2}, discard if not={view}{mix}] {\rtdepth};
  \addplot[ybar stacked, bar width=0.8, fill=#3, draw=none] table[col sep=comma, x expr=\thisrow{layer}+1, y expr=100*\thisrow{share_tenth},
    discard if not={arm}{#1}, discard if not={family}{#2}, discard if not={view}{mix}] {\rtdepth};
  \addplot[ybar stacked, bar width=0.8, fill=#3!40!white, draw=none] table[col sep=comma, x expr=\thisrow{layer}+1, y expr=100*\thisrow{share_half},
    discard if not={arm}{#1}, discard if not={family}{#2}, discard if not={view}{mix}] {\rtdepth};
  \addplot[only marks, rt dclm, mark size=1.1pt, mark options={fill=white, draw=expInk, line width=0.35pt}]
    table[col sep=comma, x expr=\thisrow{layer}+1, y expr=100*\thisrow{share_under},
    discard if not={arm}{#1}, discard if not={family}{#2}, discard if not={view}{dclm}] {\rtdepth};
}
\newcommand{\rtstructural}{%
  \fill[rtStruct, opacity=0.6] (axis cs:22.5,\pgfkeysvalueof{/pgfplots/ymin}) rectangle (axis cs:25.5,\pgfkeysvalueof{/pgfplots/ymax});}
\pgfplotsset{rt ya/.style={ymin=0.2, ymax=1.03, ytick={0.2,0.4,0.6,0.8,1}},
  rt yb/.style={ymin=0, ymax=62, ytick={0,25,50}}}
\begin{tikzpicture}
\begin{groupplot}[exps, group style={group size=4 by 3, horizontal sep=0.25cm, vertical sep=0.3cm,
    xticklabels at=edge bottom, yticklabels at=edge left},
  scale only axis, width=0.2\textwidth, height=0.145\textwidth, xmin=0.2, xmax=25.8, xtick={1,5,10,15,20,25},
  tick label style={font=\sffamily\fontsize{6}{7}\selectfont, expMuted}, label style={font=\sffamily\scriptsize},
  every axis title/.append style={font=\sffamily\scriptsize}, major tick length=2pt,
  axis line style={expMuted, line width=0.5pt}, tick style={expMuted, line width=0.5pt}]
\nextgroupplot[title={Columns}, rt ya, ylabel={effective choices / \textit{E}}]
  \rtstructural \rtdepthviews{flatq12}{group}{eff_norm}{expCortex}
\nextgroupplot[title={Shared experts}, rt ya]
  \rtdepthviews{flatq12}{public}{eff_norm}{expCortex}
\nextgroupplot[title={Private experts}, rt ya]
  \addplot[name path=lo, draw=none, forget plot] table[col sep=comma, x expr=\thisrow{layer}+1, y=eff_norm_min, discard if not={arm}{flatq12}, discard if not={family}{private}, discard if not={view}{mix}] {\rtdepth};
  \addplot[name path=hi, draw=none, forget plot] table[col sep=comma, x expr=\thisrow{layer}+1, y=eff_norm_max, discard if not={arm}{flatq12}, discard if not={family}{private}, discard if not={view}{mix}] {\rtdepth};
  \addplot[expCortex, opacity=0.18, forget plot] fill between[of=lo and hi];
  \rtdepthviews{flatq12}{private}{eff_norm}{expCortex}
  \rtdepthline{flatq12}{private}{mix}{eff_joint_norm}{expCortex, densely dashed, line width=0.7pt, mark=none}
\nextgroupplot[title={MoE experts}, rt ya, xmax=24.8]
  \rtdepthviews{moe500}{expert}{eff_norm}{expMoe}
\nextgroupplot[rt yb, ylabel={choices (\%)}]
  \rtstructural \rtdepthbars{flatq12}{group}{expCortex}
  \node[exps annot, text=expMuted, anchor=north west] at (rel axis cs:0.02,0.98) {top-3 to top-8 of 8};
\nextgroupplot[rt yb] \rtdepthbars{flatq12}{public}{expCortex}
  \node[exps annot, text=expMuted, anchor=north west] at (rel axis cs:0.02,0.98) {top-2 of 8};
\nextgroupplot[rt yb] \rtdepthbars{flatq12}{private}{expCortex}
  \node[exps annot, text=expMuted, anchor=north west] at (rel axis cs:0.02,0.98) {top-2 / top-1 of 12};
\nextgroupplot[rt yb, xmax=24.8] \rtdepthbars{moe500}{expert}{expMoe}
  \node[exps annot, text=expMuted, anchor=north west] at (rel axis cs:0.08,0.98) {top-3 of 24};
\nextgroupplot[rt ya, xlabel={layer}, ylabel={token entropy / log \textit{E}}]
  \rtstructural \rtdepthviews{flatq12}{group}{token_entropy_norm}{expCortex}
\nextgroupplot[rt ya, xlabel={layer}] \rtdepthviews{flatq12}{public}{token_entropy_norm}{expCortex}
\nextgroupplot[rt ya, xlabel={layer}] \rtdepthviews{flatq12}{private}{token_entropy_norm}{expCortex}
\nextgroupplot[rt ya, xlabel={layer}, xmax=24.8, legend to name=rt:depth, legend columns=6,
    legend style={font=\sffamily\fontsize{6}{7}\selectfont, /tikz/every even column/.append style={column sep=0.55em}}]
  \rtdepthviews{moe500}{expert}{token_entropy_norm}{expMoe}
  \legend{DCLM, Books, Math, Code, German, equal-token mixture}
\end{groupplot}
\node[below=0.62cm] at ($(group c1r3.south)!0.5!(group c4r3.south)$) {\pgfplotslegendfromname{rt:depth}};
\node[below=0.95cm, font=\sffamily\fontsize{6}{7}\selectfont, text=expInk, align=center] at ($(group c1r3.south)!0.5!(group c4r3.south)$) {%
  bars (mixture), share of the layer's choices:\enspace \tikz\fill[expInk] (0,0) rectangle (0.18,0.12); never selected\quad
  \tikz\fill[expMuted] (0,0) rectangle (0.18,0.12); below 1/10 of the uniform load\quad
  \tikz\fill[expMuted!40!white] (0,0) rectangle (0.18,0.12); 1/10 to 1/2 of it\\
  pentagons: the three together on DCLM alone\\
  private experts: line = mean over the 8 columns, band = min to max, dashed = joint over all 96 experts};
\end{tikzpicture}
    \caption{\small \textbf{Routing statistics of the trained \(500\)M
        models} across depth, one column of panels per router family.
        \textbf{Top:} effective number of choices of the aggregate
        assignment distribution on each held-out corpus (thin lines) and on
        their equal-token mixture (heavy line). \textbf{Middle:} share of
        the layer's choices that are never selected or selected well below
        the uniform load. \textbf{Bottom:} mean per-token entropy of the
        router's softmax as a fraction of \(\log E\), which measures how
        concentrated the router's distribution is for each token, as opposed
        to how balanced the load is. In the last block of \ours{}, every
    column is selected (shaded).}
    \label{fig:routing-balance}
\end{figure}

\paragraph{Routing statistics.} For the \(500\)M \ours{} and the \(500\)M MoE
transformer after context extension to \(32\)k (the starting checkpoints of
\Cref{app:domain-adaptation-settings}), we record the selections of every
router on the first \(64\) held-out sequences of \(32{,}768\) tokens of each
of the five corpora of \Cref{sec:domain-adaptation} (DCLM-baseline, books,
mathematics, code, and German; \(2.1\)M tokens per corpus), in evaluation
mode. We repeat this for the models adapted to each corpus, with and
without frozen routers, on the same sequences, to count the assignments
that change during adaptation. In \Cref{fig:routing-balance} we plot, per
router family and layer, the effective number of choices \(\exp H(p)
/ E\) of the
distribution \(p\) of the assignments over the \(E\) choices, the share of
choices that are rarely or never selected, and the mean per-token entropy
of the router's softmax as a fraction of \(\log E\).

\paragraph{Balance and entropy.} On the pretraining corpus the load of
both models is almost uniform over the choices (effective choices of at
least \(0.97 E\) in every router family and layer), with one exception:
the routing in the first layer of the MoE transformer has collapsed: two
experts are each selected by about \(70\%\) of the tokens and one expert is
never selected (\(0.41 E\)). On the other corpora the load is more concentrated
in both models, most of all on code and German, where a third to a half of
the experts of some layers have less than half of the uniform load; for
the columns this share is smaller (\(15\)--\(21\%\)). For each token, the
softmax of the column router is close to uniform: in the six routed blocks
its entropy is \(0.95\)--\(0.997\) of \(\log C\), so the gating weights of
the selected columns are within a few percent of uniform, and its scores
have an RMS of \(0.15\)--\(0.41\), below the mean magnitude
\(0.44\) of the balancing bias. However, the selection depends on the
input and is stable under noise: routing with the training-time noise
agrees with noise-free routing on \(85\)--\(98\%\) of tokens per
block, and for only \(1\)--\(16\%\) of tokens is the selection the
same as that by the bias alone. The softmax of the expert routers is
more concentrated, with per-token entropies of \(0.4\)--\(0.9\) of
\(\log E\).

\begin{table}[ht]
    \centering
    \small
    \setlength{\tabcolsep}{5pt}
    \begin{tabular}{lccc}
        \toprule
        & Corpus information & \multicolumn{2}{c}{Assignments changed
        by adaptation} \\
        Routing decision & (bits; range over layers) & routers frozen
        & all parameters trained \\
        \midrule
        \ours{}: which column & \(0.05\) (\(0.01\)--\(0.08\)) &
        \(4.7\%\) & \(13.7\%\) \\
        \ours{}: which shared expert & \(0.13\) (\(0.01\)--\(0.36\))
        & \(15.6\%\) & \(24.4\%\) \\
        \ours{}: which expert of the column & \(0.28\)
        (\(0.04\)--\(0.54\)) & \(25.8\%\) & \(32.6\%\) \\
        MoE transformer: which expert & \(0.25\) (\(0.02\)--\(0.48\))
        & \(21.7\%\) & \(30.7\%\) \\
        \bottomrule
    \end{tabular}
    \caption{Information in each routing level, for the \(500\)M models of
        \Cref{fig:routing-balance}. \emph{Corpus information:} mutual
        information, in bits, between the corpus (five corpora with equal
        token counts, entropy \(2.32\) bits) and the selection,
        counting each selection as one draw; mean over the routed layers,
        range in parentheses. For the experts of a column, the information
        is conditional on the column, so the column and expert rows
        add up to the information of the pair. \emph{Assignments
        changed:} share of tokens of the adaptation corpus whose
        selected set differs from that before adaptation, after
        \(0.5\)B tokens of adaptation without replay, averaged over the
        four corpora of \Cref{sec:domain-adaptation}; with frozen routers,
        every change is due to a change of the routers' inputs. Training on
        the pretraining corpus alone changes \(2\) / \(8\) /
    \(15\) / \(11\%\) of the assignments in the four rows.}
    \label{tab:routing-information}
\end{table}

\paragraph{Information in each level.} The column assignment contains almost
no information about the corpus (\Cref{tab:routing-information}; \(0.05\)
bits of a possible \(2.32\)), while the choice of expert within a column
contains as much as the choice of expert in the MoE transformer (\(0.28\)
and \(0.25\) bits). The \(\log_{2}\)
enrichment of a corpus at an expert of a column is the sum of its
enrichment at the column and its enrichment within the column, and the mean
absolute value of the second term is three times that of the first
(\(0.84\) against \(0.29\)). The column assignment is also the most
stable decision
under domain adaptation: with frozen routers, \(4.7\%\) of the column
assignments on the target corpus change, against \(26\%\) of the
assignments to the experts of the columns and \(22\%\) of the assignments
in the MoE transformer. Thus the column routing is largely the same across
corpora, and the selections that depend on the corpus are those of the
expert routers within each column.

\paragraph{Routing counterfactuals.} To measure how much the held-out
loss depends on each routing level, we re-evaluate the pretrained models of
\Cref{sec:pretraining} on the held-out set of \Cref{fig:pretraining}, at
context length \(4\)k, with the decisions of one router family replaced at
evaluation time and without any training. In the \emph{shuffled}
condition, each token is given the routing output (selected set and gating
weights) of another token of the same sequence, chosen uniformly at
random. The number of tokens in every column and expert, and hence the
attention FLOPs, are then unchanged, and only the assignment of decisions
to tokens changes. In the \emph{random} condition, each token is assigned
a uniformly random subset of \(K\) choices, weighted by the router's
scores at those choices. When the column routing is replaced, different tokens
are routed to each column, and the expert routers of the column route
these tokens with their learned parameters, so this condition changes only
the column decision. With the learned routing, we obtain the losses of
\Cref{tab:exp-rungs} to four decimals. The increase in held-out loss for
\ours{} and the MoE transformer at \(200\)M, \(500\)M, and \(1\)B, and for
the column-routed model with the MoE's shared expert bank (row 7 of
\Cref{tab:attribution}), is in
\Cref{tab:routing-counterfactual}.

\begin{table}[ht]
    \centering
    \small
    \setlength{\tabcolsep}{3pt}
    \begin{tabular}{lcccccc}
        \toprule
        & \multicolumn{3}{c}{\ours{}} & \multicolumn{2}{c}{MoE
        transformer} & Column-routed, \\
        \cmidrule(lr){2-4} \cmidrule(lr){5-6}
        Replaced routing & 200M & 500M & 1B & 200M & 500M / 1B & shared bank \\
        \midrule
        columns, shuffled & \(0.40\) / \(0.48\) & \(0.74\) & \(0.62\)
        & -- & -- & \(0.50\) \\
        columns, random & \(0.39\) / \(0.46\) & \(0.69\) & \(0.79\) &
        -- & -- & \(0.53\) \\
        shared experts, shuffled & \(2.59\) / \(2.53\) & \(2.11\) &
        \(2.13\) & -- & -- & -- \\
        experts of the columns, shuffled & \(0.99\) / \(1.08\) &
        \(1.36\) & \(1.31\) & -- & -- & -- \\
        both expert levels, shuffled & \(3.96\) / \(3.77\) & \(3.82\)
        & \(3.61\) & -- & -- & -- \\
        experts, shuffled & -- & -- & -- & \(4.01\) / \(4.23\) &
        \(4.47\) / \(4.51\) & \(4.08\) \\
        experts, random & -- & -- & -- & \(3.88\) / \(3.99\) &
        \(4.16\) / \(4.57\) & \(4.16\) \\
        all routing, shuffled & \(4.14\) / \(3.94\) & \(4.09\) &
        \(4.23\) & -- & -- & \(4.39\) \\
        \bottomrule
    \end{tabular}
    \caption{Increase in held-out loss (nats) of the pretrained models when
        one routing level is replaced at evaluation time. The columns of
        the table are the models of \Cref{tab:exp-rungs} and the
        column-routed model with the shared bank of the MoE transformer
        (row 7 of \Cref{tab:attribution}, at \(200\)M); the \(200\)M models
        are given for both seeds, as seed 42 / seed 43. For
        reference, the loss of
        the MoE transformer exceeds the loss of \ours{} by \(0.016\) /
        \(0.025\) at \(200\)M, \(0.035\) at \(500\)M, and \(0.022\) at
    \(1\)B.}
    \label{tab:routing-counterfactual}
\end{table}

\begin{table}[ht]
    \centering
    \small
    \setlength{\tabcolsep}{4pt}
    \begin{tabular}{lccccccccc}
        \toprule
        & \multicolumn{7}{c}{Block (active columns of \(8\))} & & \\
        \cmidrule(lr){2-8}
        Model & 1 (3) & 2 (4) & 3 (5) & 4 (6) & 5 (6) & 6 (7) & 7 (8)
        & Sum & All blocks \\
        \midrule
        \ours{}, seed 42 & \(0.003\) & \(0.124\) & \(0.049\) &
        \(0.045\) & \(0.064\) & \(0.027\) & \(0.001\) & \(0.312\) & \(0.398\) \\
        \ours{}, seed 43 & \(0.002\) & \(0.124\) & \(0.057\) &
        \(0.060\) & \(0.070\) & \(0.027\) & \(0.002\) & \(0.341\) & \(0.479\) \\
        Column-routed, shared bank & \(0.054\) & \(0.046\) &
        \(0.058\) & \(0.078\) & \(0.039\) & \(0.030\) & \(0.009\) &
        \(0.314\) & \(0.500\) \\
        \bottomrule
    \end{tabular}
    \caption{Increase in held-out loss (nats) of the \(200\)M column-routed
        models when the column routing of one block is shuffled and the
        other blocks have their learned routing, next to the sum over
        blocks and the cost of shuffling all blocks at once (the first row
        of \Cref{tab:routing-counterfactual}). In the last block every
    column is selected, so shuffling changes only its gating weights.}
    \label{tab:routing-counterfactual-blocks}
\end{table}

\paragraph{Results.} First, the loss depends
strongly on the column assignment at every scale: replacing it increases
the loss by \(0.40\)--\(0.48\) nats at \(200\)M, \(0.74\) at \(500\)M, and
\(0.62\) at \(1\)B, which is \(18\)--\(28\) times the difference between
the losses of the MoE transformer and of \ours{} at the same scale, and
the shuffled and random conditions agree. The loss depends on which
columns a token is routed to, although the gating weights of the columns
are nearly uniform, so the learned columns are not equivalent to a
random partition of the tokens. Second, replacing the column assignment
increases the loss by a similar amount (\(0.50\) nats) in the
column-routed model whose experts are shared by all columns. The
importance of the column decision is therefore due to column-routed
attention and not to the nesting, and the evidence that the improvement of
\ours{} is due to the nesting comes from the trained ablations of
\Cref{app:ablations}, not from these counterfactuals. Third, at \(200\)M,
replacing both expert levels of \ours{} increases the loss about as much
as replacing the single expert level of the MoE transformer (\(3.8\)--\(4.0\)
against \(4.0\)--\(4.2\) nats). At larger scales the two increases differ
more (\(3.6\) against \(4.5\) at \(1\)B), while the increase for the column
level is larger than at \(200\)M, so at larger scales a larger share of
the effect of the routing is due to the columns. Within the expert levels
of \ours{}, the loss depends most on the shared experts.
The effect of the column assignment of each block is given in
\Cref{tab:routing-counterfactual-blocks}. In \ours{}, shuffling the
assignment of the
first block, which is computed before any attention layer, has almost no
effect, shuffling the second block has the largest effect, and the effect
decreases towards the last block. The sum over blocks is below the
increase from replacing all blocks at once, so the routers of later blocks
depend on features computed under the learned routing of the earlier
blocks. In the column-routed model with the shared
bank, shuffling the assignment of the first block has an effect
(\(0.054\) nats).

\section{Optimization and Initialization}
\label{app:experimental-parameterization}

\paragraph{Optimization.} The matrices of the attention operators and of the
experts are trained by Muon \citep{jordan2024muon} with the base learning rate
\(\eta_{\mathrm{M}} = 0.01\). The update of a matrix in
\(\R^{D_{\mathrm{out}} \times D_{\mathrm{in}}}\) is multiplied by
\(\sqrt{D_{\mathrm{out}} / D_{\mathrm{in}}}\), so that its spectral norm
scales as in the spectral condition of \citet{yang2023spectral}. All other
parameters are trained by AdamW \citep{loshchilov2019decoupled} with the base
learning rate \(\eta_{\mathrm{A}} = 3 \times 10^{-4}\). These are the token
embeddings, the output head, the expert routers of the MoE baseline and of
\ours{} (including the routers of the shared experts and of the experts of
each column, and the gate \eqref{eq:app-path-gate}), the queries and gains of
the attention residual, and the normalization gains. The only exception is
the last layer of the column router, which has the learning rate
\(\eta_{\mathrm{A}} / (D_{\mathrm{in}} \sqrt{M})\), where \(D_{\mathrm{in}} =
128\) is the hidden width of the router and \(M\) is the number of layers in
its block: this router is evaluated once and used by all \(M\) layers, so its
gradient is a sum of \(M\) terms. The Muon base learning rate was selected at
200M and reused at the other scales. The AdamW base learning rate was fixed,
without a separate sweep.

\begin{table}[ht]
    \centering
    \small
    \begin{tabular}{ll}
        \toprule
        Setting & Value \\
        \midrule
        Learning rate, Muon & \(0.01\) \\
        Learning rate, AdamW & \(3 \times 10^{-4}\) \\
        Muon momentum & \(0.95\), Nesterov \\
        Newton--Schulz iterations & \(5\) \\
        AdamW \((\beta_1, \beta_2)\), \(\epsilon\) & \((0.9, 0.95)\),
        \(10^{-8}\) \\
        Warmup & linear, \(8\%\) of the steps \\
        Decay & cosine, minimum \(10^{-5}\) \\
        Weight decay & \(0.1\) on matrices, \(0\) on vectors \\
        Gradient clipping & \(1.0\) \\
        Dropout & \(0\) \\
        Sequence length & \(4{,}096\) \\
        Precision & bfloat16 autocast \\
        Seed & \(42\) (\(43\) for the second seed) \\
        \bottomrule
    \end{tabular}
    \caption{Pretraining hyperparameters, shared by all architectures and
    scales.}
    \label{tab:pretraining-hparams}
\end{table}

\paragraph{Initialization.} The three architectures have the same
initialization. The query, key, and value projections, the input and gate
matrices of the experts, the expert routers, and the gate
\eqref{eq:app-path-gate} are initialized with standard deviation \(0.1 /
\sqrt{8}\). The output projections of the attention operators and the output
matrices of the experts are initialized with this standard deviation divided by
\(\sqrt{2L}\), where \(L\) is the number of layers of the model. The token
embeddings and the output head have standard deviation \(0.1\); the output of
the embedding is multiplied by \(12\), and the output logits are divided by
\(8\). The last layer of the column router is initialized with standard
deviation \(1 / D_{\mathrm{in}}\), so that the column routing is close to
uniform at initialization, and the queries of the attention residual
are initialized at zero.

\section{Compute Multipliers under Scaling Laws}\label{app:scaling-laws}

In addition to \Cref{sec:pretraining,sec:long-context-extension}, this
section gives a compute accounting for the full architectures, analogous to
the one in \Cref{thm:nested}(c) for \Cref{model:linear} at a fixed model size.
We take as a hypothesis that the losses of the full architectures follow a
scaling law with common exponents and coefficients that depend on the
architecture. We show that the compute multiplier between the two
architectures is then the same at every loss, and that it grows with the
context length because \ours{} has fewer attention FLOPs.

\subsection{The Compute Multiplier}\label{app:multiplier}

\paragraph{The multiplier at every scale.} \Cref{thm:nested} holds for fixed
\(C\), \(E\), \(r\), \(D\), and \(S\). There, the
excess risks of the two architectures have the same exponent in \(T\), the
different coefficients \(\sigma^{2} d_{s}\), and the different costs
\(F_{s}\) per sample, and \eqref{eq:compute-multiplier} is the compute
multiplier. We take as a hypothesis that the
losses of the full architectures have the same form when the model size also
grows, which \Cref{model:linear} does not capture: a scaling law of
the form of \citet{hoffmann2022training}, with
common exponents and coefficients that depend on
the architecture. Under this hypothesis, the compute multiplier is the same
at every scale. Let \(N\) be the
number of active parameters, \(T\) the number of training tokens, and \(F\)
the training FLOPs. Our compute accounting includes the attention FLOPs
(\Cref{app:configurations}). At a fixed context length,
\begin{equation}\label{eq:compute}
    F = k_{0} (1 + \nu_{s}) N T,
\end{equation}
where \(k_{0} N T\) is the FLOPs of the projections and the experts, and
\(\nu_{s}\) is the ratio of the attention FLOPs of architecture \(s\) to
these FLOPs. For equal shapes and balanced routing, \eqref{eq:attn-cost} gives
\(\nu_{\mathrm{N}} = \rho\, \nu_{\mathrm{F}}\) for \ours{} (N) and the MoE
transformer (F), where \(\rho \in (0, 1)\) is the average of \(K_{\mathrm{c}}
/ C\) over the layers of \ours{}.

\begin{proposition}{}{scaling}
    For each architecture \(s\), let \(L_{s}(N, T) = L_{\infty} + a_{s}
    N^{-\alpha} + b_{s} T^{-\beta}\) and \(F = k_{0} (1 + \nu_{s}) N T\),
    where \(\alpha, \beta, a_{s}, b_{s}, k_{0} > 0\) and \(\nu_{s} \geq 0\) are
    constants. Let \(p \doteq \alpha \beta / (\alpha + \beta)\) and \(q_{s}
    \doteq (1 + \nu_{s})\, a_{s}^{1 / \alpha} b_{s}^{1 / \beta}\). Then
    \begin{equation}\label{eq:frontier}
        \min_{N, T \colon k_{0} (1 + \nu_{s}) N T = F} L_{s}(N, T) =
        L_{\infty} + H\, (k_{0} q_{s} / F)^{p},
        \qquad
        H \doteq (\beta / \alpha)^{\alpha / (\alpha + \beta)} + (\alpha /
        \beta)^{\beta / (\alpha + \beta)}.
    \end{equation}
    For every loss \(L > L_{\infty}\), the compute at which the MoE transformer
    reaches \(L\), divided by the compute at which \ours{} reaches \(L\), is
    \begin{equation}\label{eq:multiplier}
        \frac{q_{\mathrm{F}}}{q_{\mathrm{N}}}
        = \frac{1 + \nu_{\mathrm{F}}}{1 + \nu_{\mathrm{N}}} \cdot
        \Big(\frac{a_{\mathrm{F}}}{a_{\mathrm{N}}}\Big)^{1 / \alpha}
        \Big(\frac{b_{\mathrm{F}}}{b_{\mathrm{N}}}\Big)^{1 / \beta}.
    \end{equation}
\end{proposition}

The multiplier \eqref{eq:multiplier} has two factors. The first is the saving
in attention FLOPs, and the second is the ratio of the coefficients of the
law. The ratio \(d_{\mathrm{F}} / d_{\mathrm{N}}\) of \Cref{thm:nested} is an
example of the latter at a fixed size, and the second factor of
\eqref{eq:compute-multiplier} is the former. Neither factor depends on \(L\), so
under the hypotheses of the proposition the multiplier stays constant while
the difference of the losses goes to zero. In \Cref{fig:pretraining},
the fitted exponents of \ours{} and of the MoE are equal to three
digits, and the measured multiplier \(1.13\)--\(1.29\times\) has no trend
over our four scales. The fits have separate offsets, which the proposition
does not allow; we discuss their effect at the end of
\Cref{app:theory-scaling}. \Cref{app:theory-scaling} also extends the
proposition to a ratio \(\nu_{s}\) that changes with the model size, and to
the training path of our experiments, which has a fixed ratio \(T / N\).

\subsection{Long Contexts}\label{app:theory-long}

The attention FLOPs of one token are proportional
to the context length \(S\), and the other FLOPs do not depend on \(S\). Let
\(S_{0}\) be the context length at which the two are equal for the MoE
transformer, so that \(\nu_{\mathrm{F}} = S / S_{0}\) and \(\nu_{\mathrm{N}}
= \rho S / S_{0}\). The first factor of \eqref{eq:multiplier} is then
\begin{equation}\label{eq:token-multiplier}
    R(S) \doteq \frac{1 + \nu_{\mathrm{F}}}{1 + \nu_{\mathrm{N}}} =
    \frac{S_{0} + S}{S_{0} + \rho S}.
\end{equation}
In the context extension experiments the models are fixed and \(S\) grows. For
equal shapes and balanced routing, at equal FLOPs \ours{} is trained on \(R(S)\)
times as many tokens as the MoE transformer. For our models the measured
counterpart is \(1 / r(S)\) (\Cref{tab:long-context-budgets}), which also
accounts for the slightly different shapes of the two architectures
(\Cref{fig:flops-vs-length}). The multiplier \(R(S)\) increases with
\(S\) and converges to \(1
/ \rho\), as we show below. This explains why the margin at equal
FLOPs grows with the context length in \Cref{fig:long-context-extension}. At
equal tokens the margin is unchanged, which agrees with the assumption of
\Cref{sec:theory} that the restricted context is sufficient for the
prediction. In
pretraining, \(S\) is fixed and the width \(D\) grows. The length \(S_{0}\) is
proportional to \(D\), because the attention FLOPs grow as \(S D\) and the
other FLOPs as \(D^{2}\). Then \(R(S) \to 1\), and the pretraining advantage
at large width must come from the second factor of \eqref{eq:multiplier}.

We now derive \(S_{0}\) and the properties of \(R(S)\). In one layer, let
a token cost \(\phi\) FLOPs in the projections and the
experts of either architecture and \(\omega S\) FLOPs in the attention of the
MoE transformer. By \eqref{eq:attn-cost}, it costs \(\omega (K_{\mathrm{c}} /
C) S\) FLOPs in the attention of \ours{}. We sum over the layers and obtain
\(\nu_{\mathrm{F}} = \omega S / \phi = S / S_{0}\) and \(\nu_{\mathrm{N}} =
\rho S / S_{0}\) with \(S_{0} \doteq \phi / \omega\). In the backward pass,
all costs are multiplied by the same factor. For a layer of width
\(D\), \(\phi\)
is proportional to \(D^{2}\) and \(\omega\) to \(D\), so \(S_{0}\) is
proportional to \(D\). The accounting of \Cref{app:configurations} gives
\(S_{0} = 16 D\) for our models. The ratio \(R(S)\) of
\eqref{eq:token-multiplier} has the derivative \(R'(S) = S_{0} (1 - \rho) /
(S_{0} + \rho S)^{2} > 0\) and the limit \(1 / \rho\) as \(S \to \infty\).

To express the saving in terms of the loss, we assume a learning
curve for the continued
training of \ours{} at the context length \(S\),
\begin{equation}\label{eq:continuation}
    L_{\mathrm{N}}(T; S) = \ell(S) + b(S) (T + \tau_{S})^{-\beta},
    \qquad \tau_{S} \geq 0,
\end{equation}
where \(\tau_{S}\) accounts for the pretrained initialization. Let
\(\Delta_{T}(S)\) be the loss of the MoE transformer minus the loss of
\ours{} when both are trained on \(T\) tokens. When \ours{} is trained on
\(R(S) T\) tokens at equal FLOPs, the difference is
\begin{equation}\label{eq:equal-flops-gap}
    \Delta_{F}(S) = \Delta_{T}(S) + b(S) \big[(T + \tau_{S})^{-\beta} - (R(S)
    T + \tau_{S})^{-\beta}\big].
\end{equation}
The second term is positive and increasing in \(R(S)\). Therefore
\(\Delta_{F}(S) - \Delta_{T}(S)\) grows with \(S\) when \(b(S)\) and
\(\tau_{S}\) change little with \(S\), which agrees with
\Cref{fig:long-context-extension}.

\subsection{Proof of \Cref{prop:scaling} and
Extensions}\label{app:theory-scaling}

The law has the form of \citet{hoffmann2022training}.
\citet{clark2022unified} study routed models, where the exponents can depend
on the number of experts. At the fixed sparsity of our experiments, the
relation between the stored and the active parameters is absorbed in
\(a_{s}\).

\begin{proof}[Proof of \Cref{prop:scaling}]
    Let \(\kappa_{s} \doteq k_{0} (1 + \nu_{s})\), and substitute \(T = F /
    (\kappa_{s} N)\). As a function of \(\log N\), the excess loss \(a_{s}
    N^{-\alpha} + b_{s} (\kappa_{s} N / F)^{\beta}\) is the sum of
    decreasing and increasing exponentials. It is strictly convex and
    diverges at both
    ends, so it has a unique minimizer, where \(\alpha a_{s} N^{-\alpha} =
    \beta b_{s} T^{-\beta}\). With \(u \doteq a_{s} N^{-\alpha}\),
    \(b_{s} T^{-\beta} = \alpha u / \beta\), and
    \begin{equation}
        \frac{F}{\kappa_{s}} = N T
        = \Big(\frac{a_{s}}{u}\Big)^{1 / \alpha} \Big(\frac{\beta
        b_{s}}{\alpha u}\Big)^{1 / \beta}
        = a_{s}^{1 / \alpha} b_{s}^{1 / \beta} \Big(\frac{\beta}{\alpha}\Big)^{1
        / \beta} u^{-1 / p}.
    \end{equation}
    Since \(\kappa_{s} a_{s}^{1 / \alpha} b_{s}^{1 / \beta} = k_{0} q_{s}\) and
    \(p / \beta = \alpha / (\alpha + \beta)\), we obtain
    \begin{equation}
        u = (\beta / \alpha)^{\alpha / (\alpha + \beta)} (k_{0} q_{s} / F)^{p},
        \qquad
        \alpha u / \beta = (\alpha / \beta)^{\beta / (\alpha + \beta)}
        (k_{0} q_{s} / F)^{p}.
    \end{equation}
    Their sum is \eqref{eq:frontier}, and solving it for \(F\) at a
    loss \(L > L_{\infty}\) gives \(F = k_{0} q_{s} (H / (L -
    L_{\infty}))^{1 / p} \propto q_{s}\), giving \eqref{eq:multiplier}.
\end{proof}

\paragraph{Attention ratio that depends on the model size.} At a fixed
context length, the ratio \(\nu_{s}\) decreases as the width grows
(\Cref{app:theory-long}), so it is a function \(\nu_{s}(N)\). The frontier
then has no closed form, but the comparison of the architectures remains
valid. Suppose that \(a_{\mathrm{N}} \leq a_{\mathrm{F}}\), \(b_{\mathrm{N}}
\leq b_{\mathrm{F}}\), and \(\nu_{\mathrm{N}}(N) \leq \nu_{\mathrm{F}}(N)\)
for all \(N\). Let \((N, T)\) be any allocation of the MoE transformer with
compute \(F\). With the same compute, a model of \ours{} of size \(N\) can be
trained on \(T' = T (1 + \nu_{\mathrm{F}}(N)) / (1 +
\nu_{\mathrm{N}}(N)) \geq T\)
tokens, and \(L_{\mathrm{N}}(N, T') \leq L_{\mathrm{F}}(N, T)\). Therefore
\ours{} has the lower frontier at every \(F\).

\paragraph{Fixed ratio of tokens to parameters.} Our models are trained with
\(T = \varrho N\) for a constant \(\varrho\), which is not the optimal
allocation. On this path, \(L_{s} - L_{\infty} = a_{s} N^{-\alpha} + b_{s}
\varrho^{-\beta} N^{-\beta}\) and \(F = k_{0} (1 + \nu_{s}(N)) \varrho
N^{2}\). Let \(\alpha = \beta\) and \(k_{s} \doteq a_{s} + b_{s}
\varrho^{-\alpha}\). Architecture \(s\) reaches the loss \(L\) at the size
\(N_{s} = (k_{s} / (L - L_{\infty}))^{1 / \alpha}\), and the ratio of the
compute of the two architectures at this loss is
\begin{equation}\label{eq:fixed-ratio}
    \frac{F_{\mathrm{F}}}{F_{\mathrm{N}}}
    = \frac{1 + \nu_{\mathrm{F}}(N_{\mathrm{F}})}{1 +
    \nu_{\mathrm{N}}(N_{\mathrm{N}})} \cdot
    \Big(\frac{k_{\mathrm{F}}}{k_{\mathrm{N}}}\Big)^{2 / \alpha}.
\end{equation}
This has the same two factors as \eqref{eq:multiplier}. The second factor is
constant. The first factor is constant when \(\nu_{s}\) is constant, and then
\(L_{s} - L_{\infty} = k_{s} (k_{0} (1 + \nu_{s}) \varrho / F)^{\alpha / 2}\)
is one power law in \(F\), which is the form fitted in
\Cref{fig:pretraining}. When the width grows at a fixed context length, the
first factor converges to \(1\), and the multiplier converges to
\((k_{\mathrm{F}} / k_{\mathrm{N}})^{2 / \alpha}\). If \(\alpha \neq \beta\),
the two terms of \(L_{s} - L_{\infty}\) change their relative size with
\(N\), and the multiplier depends on \(L\). In this case \(a_{\mathrm{N}}
\leq a_{\mathrm{F}}\), \(b_{\mathrm{N}} \leq b_{\mathrm{F}}\), and
\(\nu_{\mathrm{N}} \leq \nu_{\mathrm{F}}\) are still sufficient for \ours{}
to have the lower loss at every \(F\) when the compute \((1 + \nu_{s}(N))
N^{2}\) increases with \(N\): with the same compute, a model of
\ours{} that is at least as
large can then be trained on at least as many tokens. The fits in
\Cref{fig:pretraining} have a separate offset for each architecture, and the
offset of \ours{} is lower. A lower limit \(L_{\infty}\) for \ours{} makes the
multiplier grow with the scale.

\section{Theoretical Analysis}\label{app:theory}

In this section we give the formal versions of the results stated
informally in \Cref{sec:theory}, together with further discussion. The
setup is that of \Cref{sec:theory}: the data are generated by
\Cref{model:linear}, the nested and flat predictors
\eqref{eq:nested-predictor} are fit by least squares, and a fit is
measured by its excess risk, both as in \eqref{eq:least-squares}. We state
the results on sample and compute efficiency in \Cref{app:theory-sample} and
on forgetting in \Cref{app:theory-forget}, prove them in
\Cref{app:theory-nested,app:theory-forgetting}, and compare the nested
predictor with other flat predictors in \Cref{app:theory-comparisons}.

\subsection{Sample and Compute Efficiency}\label{app:theory-sample}

We write \(\ve_{c} \in \R^{C}\) for the \(c\)-th standard basis vector. The
\textit{narrow flat predictor} is the flat predictor in
\eqref{eq:nested-predictor} whose attention score is computed from the raw
tokens \(\vx_{i} = (\ve_{c_{i}}, \vz_{i}) \in \R^{C + D}\) rather than from
the product features, so that its score is \(\vx_{S}^{\top} \vM \vx_{i}\)
for a matrix \(\vM \in \R^{(C + D) \times (C + D)}\) that is shared by all
pairs of labels, and its values are \(\vz_{i}\) as in
\eqref{eq:nested-predictor}.

\begin{theorem}{}{nested}
    In \Cref{model:linear}, let \(C \geq 2\), and let
    \(\hat{\vf}^{\mathrm{N}}\) and \(\hat{\vf}^{\mathrm{F}}\) be least
    squares fits of the nested and flat predictors
    \eqref{eq:nested-predictor} on \(T\) samples.
    \begin{enumerate}[(a), nosep, leftmargin=2em]
        \item Both predictors can represent the regression function
            \(\vf^{\star}\): the nested predictor with \(\vW_{c} =
            \vW_{c}^{\star}\), \(\vA_{c} = \vA_{c}^{\star}\), and
            \(\vB_{ce} = \vB_{ce}^{\star}\), and the flat predictor with
            \(\vW_{cc} = \vW_{c}^{\star}\), \(\vW_{cc'} = \vzero\) for \(c'
            \neq c\), \(\vV = \vI\), \(\vW_{ce}^{\mathrm{in}} =
            \vA_{c}^{\star}\), and \(\vW_{ce}^{\mathrm{out}} =
            \vB_{ce}^{\star}\). The narrow flat predictor cannot represent
            \(\vf^{\star}\) for any value of its weights.
        \item As \(T \to \infty\), we have \(T\, \cE(\hat{\vf}^{\mathrm{N}})
            \rightsquigarrow \sigma^{2} \chi^{2}_{d_{\mathrm{N}}}\) and \(T\,
                \cE(\hat{\vf}^{\mathrm{F}}) \rightsquigarrow \sigma^{2}
            \chi^{2}_{d_{\mathrm{F}}}\) in distribution, where
            \begin{small}
                \begin{equation}\label{eq:dims}
                    d_{\mathrm{N}} = \underbrace{C \cdot
                    D^{2}}_{\text{patterns}} + \underbrace{C \cdot r (D -
                    r)}_{\text{row spaces}} + \underbrace{C E \cdot r
                    D}_{\text{maps}} - C,
                    \quad
                    d_{\mathrm{F}} = \underbrace{C^{2} \cdot
                    D^{2}}_{\text{patterns}} + \underbrace{C E \cdot r (D -
                    r)}_{\text{row spaces}} + \underbrace{C E \cdot r
                    D}_{\text{maps}} - C.
                \end{equation}
            \end{small}
            The limits have the means \(\sigma^{2} d_{\mathrm{N}}\) and
            \(\sigma^{2} d_{\mathrm{F}}\), and \(1 < d_{\mathrm{F}} /
            d_{\mathrm{N}} \leq C\).
        \item Let \(F_{0}, F_{1} > 0\), and let a sample cost \(F_{0} +
            F_{1} k\) FLOPs when scores are computed for \(k\)
            context tokens, so
            that the expected FLOPs per sample are \(F_{\mathrm{N}} \doteq
            F_{0} + F_{1} (S - 1) / C\) and \(F_{\mathrm{F}} \doteq F_{0} +
            F_{1} (S - 1)\). If predictor \(s \in \{\mathrm{N},
            \mathrm{F}\}\) is
            trained on \(T = F / F_{s}\) samples, then \(F\,
                \cE(\hat{\vf}^{s}) \rightsquigarrow \sigma^{2} F_{s}
            \chi^{2}_{d_{s}}\), whose mean is \(\sigma^{2} d_{s} F_{s}\).
            The two means are equal when the flat predictor uses
            \begin{equation}\label{eq:compute-multiplier}
                \kappa(S) \doteq \frac{d_{\mathrm{F}}}{d_{\mathrm{N}}}
                \cdot \frac{S_{0} + S - 1}{S_{0} + (S - 1) / C}, \qquad
                S_{0} \doteq F_{0} / F_{1},
            \end{equation}
            times the FLOPs of the nested predictor, and \(\kappa\)
            increases from \(d_{\mathrm{F}} / d_{\mathrm{N}}\) at \(S = 1\)
            to \(C \cdot d_{\mathrm{F}} / d_{\mathrm{N}}\) as \(S \to \infty\).
    \end{enumerate}
\end{theorem}

\paragraph{Interpretation of part (a).} The two predictors can represent the
same target, because the experts of the flat predictor can include the value
maps of the columns, and the blocks of its pattern can implement the
routing. The second requires that the score of the shared attention depend
on the labels of both tokens, which is impossible for a bilinear score of
the raw tokens: in the narrow flat predictor, the term of the score that is
bilinear in \((\vz_{S}, \vz_{i})\) does not depend on the labels, so it
cannot vanish on the tokens of the other labels and equal
\(\vz_{S}^{\top} \vW_{c}^{\star} \vz_{i}\) on the relevant tokens. With
the product features, the input of the shared attention is \(C\) times
wider, and \Cref{rmk:simulation} gives the same construction for the full
architectures. Part (b) quantifies the statistical cost of this width.

\paragraph{Interpretation of part (b).} The predictors differ in the
quantities estimated from data. The number \(d\) is the dimension of the set
of functions that a predictor can represent near the target, which is the
number of directions in weight space that change the prediction. It is
smaller than the number of stored weights, because \(\vV\) is redundant, an
invertible change of the \(r\) hidden units does not change the
prediction, and multiplying the pattern of a label by a constant and its
maps by the inverse does not change the prediction, which accounts
for the term \(- C\) in both counts. Each dimension adds \(\sigma^{2}\) to
the mean of the limit of \(T\, \cE\), as in the classical asymptotic theory
of least squares over a smooth model \citep{vaart1998asymptotic,
geyer1994asymptotics}.

The three terms of \eqref{eq:dims} are the patterns, the row spaces, and
the maps; the last two are shown in \Cref{fig:theory-model}b,c. In the
nested predictor, one pattern \(\vW_{c} \in \R^{D \times D}\) per column is
estimated from the tokens routed to the column. In the flat predictor, one
block \(\vW_{cc'}\) per pair of labels is estimated, i.e., \(C^{2}\) blocks,
of which the \(C (C - 1)\) blocks with \(c' \neq c\) are zero at the
target: in the flat predictor these zero blocks must be estimated from
data, whereas in \ours{} the tokens of the other labels are removed by the
routing before the pattern is applied. A matrix \(\vG\) of rank \(r\)
is determined
by its row space and by a linear map from the row space to \(\R^{D}\). The
row space is an \(r\)-dimensional subspace of \(\R^{D}\) and has \(r (D -
r)\) free parameters, and the map has \(r D\). In both predictors, a
separate map is estimated for each of the \(C E\) experts. In the
flat predictor each
expert has its own \(\vW_{ce}^{\mathrm{in}}\), so the row space of each
expert is estimated from the samples of that expert only. In the
nested predictor all experts
of a column have the row space of \(\vA_{c}\), which is estimated from the
samples of all \(E\) experts of the column. Altogether,
\begin{equation}\label{eq:dims-gap}
    d_{\mathrm{F}} - d_{\mathrm{N}} = C (C - 1) \cdot D^{2} + C (E - 1)
    \cdot r (D - r),
\end{equation}
where the first term is due to the routed attention and the second to the
nesting of the experts in the columns. If the flat predictor is fitted on
\(k T\) samples, then \(T\, \cE(\hat{\vf}^{\mathrm{F}}) \rightsquigarrow
(\sigma^{2} / k)\, \chi^{2}_{d_{\mathrm{F}}}\). This limit has the same
mean as the limit \(\sigma^{2} \chi^{2}_{d_{\mathrm{N}}}\) of the nested
predictor on \(T\) samples when \(k = d_{\mathrm{F}} / d_{\mathrm{N}}\). We
use this ratio of the limiting means to compare the sample sizes
needed by the two predictors. It equals \((C (C + E) D^{2} - C) / (C (1 +
E) D^{2} - C)\) at \(r = D\), where a row space has no free parameters,
and it tends to \(C\) as \(D \to \infty\) for fixed \(r\) and \(E\). Dropping
the terms of lower order in \(D / r\) and in \(E\) gives the forms
\eqref{eq:dims-informal}: \(d_{\mathrm{N}} \approx C D (D + E r)\),
\(d_{\mathrm{F}} \approx C D (C D + 2 E r)\), and the ratio \((C + 2 x) /
(1 + x)\) with \(x \doteq E r / D\), which decreases in \(x\) from \(C\) to
\(2\) when \(C \geq 2\).

The dimensions in \eqref{eq:dims} do not depend on the label
probabilities or the context length \(S\). An expert with probability
\(\pi_{ce}\) is used on about \(\pi_{ce} T\) samples, so its weight
error is larger by a factor \(1 / \pi_{ce}\). It is also used with
probability \(\pi_{ce}\), and the two factors cancel in
the excess risk \eqref{eq:least-squares}. Every module therefore adds
\(\sigma^{2}\) to the
mean of the limit for each of its dimensions, however few samples it
is used on. The count \(d_{\mathrm{N}}\) includes all \(C\) patterns,
\(C\) value maps, and \(C E\) experts, so it accounts for the cost of
specialization.

Nesting requires both column-routed attention and experts within the
columns, which agrees with \Cref{tab:attribution}, where neither part
alone improves the loss. With \(C = 1\), the nested predictor is a
flat predictor, and the narrow flat predictor represents the target. With
column-routed attention and experts \(\vW_{e}\) shared by all columns, the
predictor is \(\vW_{e} \vA_{c} \sum_{i \in \cI} (\vz_{S}^{\top} \vW_{c}
\vz_{i}) \vz_{i}\). The column space of \(\vW_{e} \vA_{c}\) is contained in
that of \(\vW_{e}\), which has dimension at most \(r\), so this predictor
represents the target only if, for each \(e\), the maps
\(\vB_{ce}^{\star}\) have the same column space for all \(c\).
Nesting also helps only when the coarse variable changes the
computation. If every token were relevant to every query and neither the
patterns nor the row spaces of the \(\vA_{c}^{\star}\) depended on \(c\),
then a shared attention with a label-free pattern and a value map of width
\(r\), followed by the experts, would represent the target with one pattern
and one row space, whereas the nested predictor would estimate a separate
copy of both for each column.

\paragraph{Interpretation of part (c).} The excess risks of the two
predictors decay at the same rate in the number of samples, and they
differ in the coefficient \(\sigma^{2} d_{s}\) and in the cost
\(F_{s}\) of a sample. The multiplier \eqref{eq:compute-multiplier} is the
product of the two ratios. The first factor, \(d_{\mathrm{F}} /
d_{\mathrm{N}}\), is the cost of the estimation of the
\(C^{2}\) pattern blocks and \(C E\) row spaces of the flat
predictor. The second factor,
\((S_{0} + S - 1) / (S_{0} + (S - 1) / C)\), is the cost of scoring all
\(S - 1\) context tokens when only \((S - 1) / C\) of them are relevant; it
is the multiplier \(R(S - 1)\) of \eqref{eq:token-multiplier}
with the fraction \(\rho = 1 / C\) of active columns, and \(S_{0} = F_{0}
/ F_{1}\) is the crossover length of \Cref{sec:long-context-extension}, at
which the cost of scoring the context equals the other cost of a sample.
\Cref{app:scaling-laws} derives the same multiplier for the full
architectures from a scaling law hypothesis. This explains
\Cref{fig:long-context-extension}: at equal tokens the two predictors
differ by the first factor alone, and at equal FLOPs the second factor
grows with the context length. At a fixed context length, \(S_{0}\) grows
with the cost \(F_{0}\) of the maps, so the second factor tends to one as
the width grows, and the advantage at large widths in
\Cref{fig:pretraining} is the first factor.

\subsection{Forgetting}\label{app:theory-forget}

We formalize the continual training of \Cref{sec:theory}. The new domain
\(Q\) is a distribution over \((e, \vX, \vy)\): the fine label \(e \in
[E]\) is drawn with probability \(\pi_{e}^{Q} > 0\), the context \(\vX\) is
drawn as in \Cref{model:linear} with the query label \(c_{Q}\), the summary
\(\vm = \sum_{i \in \cI} (\vz_{S}^{\top} \vW_{c_{Q}}^{\star} \vz_{i})
\vz_{i}\) is computed with the pattern of the label \(c_{Q}\), and \(\vy =
\vG'_{e} \vm + \veps\) for fixed matrices \(\vG'_{e} \in \R^{D \times D}\)
and noise \(\veps\) as in \Cref{model:linear}. Let \(\vSigma_{e}^{Q} \doteq
\Ex_{Q}[\vm \vm^{\top} \mid e]\), and let \(\vSigma_{ce} \doteq \Ex[\vm
\vm^{\top} \mid c, e]\) under \Cref{model:linear}; both are positive
definite. The initial weights of both predictors are those of
\Cref{thm:nested}(a). The weights of each predictor follow the gradient flow
\(\dot{\vtheta} = -\vnabla_{\vtheta} L_{Q}(\vtheta)\) of the population
loss
\begin{equation}\label{eq:new-domain-loss}
    L_{Q}(\vtheta) \doteq \frac{1}{2} \sum_{e \in [E]} \pi_{e}^{Q}\,
    \Ex_{Q}\big[\norm{\vy - \vf(c_{Q}, e, \vX; \vtheta)}_{2}^{2} \,\big|\,
    e\big],
\end{equation}
where \(\vf(\cdot\,; \vtheta)\) is either predictor in
\eqref{eq:nested-predictor}. The forgetting at time \(t\) on the old data
with the labels \((c, e)\) is
\begin{equation}\label{eq:forgetting-measure}
    \Phi_{ce}(t) \doteq \frac{1}{2} \Ex\big[\norm{\vf(c, e, \vX;
    \vtheta(t)) - \vf^{\star}(c, e, \vX)}_{2}^{2} \,\big|\, c, e\big],
\end{equation}
which is the increase of the risk of \Cref{model:linear} on these labels.

\begin{proposition}{}{forgetting}
    In the setting above, let \(c \neq c_{Q}\) and \(e \in [E]\).
    \begin{enumerate}[(a), nosep, leftmargin=2em]
        \item For the nested predictor, \(\vW_{c}\), \(\vA_{c}\), and
            \(\vB_{ce}\) stay at their initial values, so \(\vf(c, e, \vX;
            \vtheta(t)) = \vf^{\star}(c, e, \vX)\) and \(\Phi_{ce}(t) = 0\)
            for all \(t \geq 0\).
        \item For the flat predictor, the blocks \(\vW_{cc'}\) and the
            experts \(\vW_{ce}^{\mathrm{in}}, \vW_{ce}^{\mathrm{out}}\) stay
            at their initial values, so \(\vf(c, e, \vX; \vtheta(t)) =
            \vG_{ce}^{\star} \vV(t) \vm\) for all \(t \geq 0\), and as \(t
            \to 0\),
            \begin{equation}\label{eq:forgetting-rate}
                \Phi_{ce}(t) = \frac{t^{2}}{2} \norm{\vB_{ce}^{\star}
                    \vA_{c}^{\star} \vA_{c_{Q}}^{\star\top} \vXi\,
                \vSigma_{ce}^{1/2}}_{F}^{2} + O(t^{3}),
                \qquad
                \vXi \doteq \sum_{e' \in [E]} \pi_{e'}^{Q}\,
                \vB_{c_{Q} e'}^{\star\top} (\vG_{c_{Q} e'}^{\star} -
                \vG'_{e'}) \vSigma_{e'}^{Q}.
            \end{equation}
            Moreover, \(\dot{\vV}(0) = -\vnabla_{\vV} L_{Q}(\vtheta(0)) =
            -\vA_{c_{Q}}^{\star\top} \vXi\), so \(\vXi \neq \vzero\) if and
            only if \(\dot{\vV}(0) \neq \vzero\). The
            leading coefficient in \eqref{eq:forgetting-rate} is zero if
            \(\vA_{c}^{\star} \vA_{c_{Q}}^{\star\top} = \vzero\), and it is
            positive if \(\vA_{c}^{\star} \vA_{c_{Q}}^{\star\top}\) is
            invertible and \(\vXi \neq \vzero\).
    \end{enumerate}
\end{proposition}

When the rows of \(\vA_{c}^{\star}\) and of \(\vA_{c_{Q}}^{\star}\) are
orthonormal, the singular values of \(\vA_{c}^{\star}
\vA_{c_{Q}}^{\star\top}\) are the cosines of the principal angles between
the two row spaces, so this matrix is the overlap matrix by which
\citet{doan2021theoretical} characterize forgetting in the linearized
regime. In this case, since \(\vB_{ce}^{\star}\) is injective and
\(\vSigma_{ce}^{1/2}\) is invertible, the leading coefficient of
\eqref{eq:forgetting-rate} lies between \(\frac{1}{2} \cos^{2}
    \theta_{r}\, \sigma_{\min}^{2}(\vB_{ce}^{\star})\, \norm{\vXi
\vSigma_{ce}^{1/2}}_{F}^{2}\) and \(\frac{1}{2} \cos^{2} \theta_{1}\,
    \norm{\vB_{ce}^{\star}}_{\mathrm{op}}^{2}\, \norm{\vXi
\vSigma_{ce}^{1/2}}_{F}^{2}\), where \(\theta_{1} \leq \dots \leq
\theta_{r}\) are the principal angles. In the proposition, the two
predictors are compared at equal time under gradient flow. In
training on \(Q\), the pattern blocks and the experts of the label
\(c_{Q}\) of the flat predictor change, as do the weights of the column
\(c_{Q}\) of the nested predictor, and in addition \(\vV\) changes.
The change of \(\vV\) is the only source of forgetting on the
other coarse labels. The nested predictor has no such shared weight,
and only its column \(c_{Q}\) changes. Since \(\vV\) is
redundant, the flat predictor with \(\vV = \vI\) held fixed has the same
function class, and it has no forgetting on the other coarse labels either.
The forgetting in part (b) is therefore due to training a value map
shared by all labels, as in the MoE transformer in
\Cref{sec:domain-adaptation}.

\subsection{Proof of \Cref{thm:nested}}\label{app:theory-nested}

Both predictors are linear in a fixed set of polynomial features of the
context, so least squares over a predictor is least squares over a class
of functions inside one finite-dimensional linear space. Least squares over
a class is a projection of the unconstrained least squares estimator onto
the class. The error of the unconstrained estimator is asymptotically an
isotropic Gaussian vector in the norm that defines the excess risk, and
after the projection only its component in the tangent space of the class
remains.
The limit of \(T\, \cE\) is therefore \(\sigma^{2}\) times a \(\chi^{2}\)
variable whose number of degrees of freedom is the dimension of the class
at the target. The dimension is the number of weights of the predictor
minus the dimension of the set of reparametrizations that leave the
prediction fixed, which we compute with the following lemma.

\begin{lemma}{}{block}
    For \(\vG, \vW \in \R^{D \times D}\), let \(P(\vG, \vW)\) be the map
    \((\vq, \vz) \mapsto \vG (\vq^{\top} \vW \vz)\, \vz\) from
    \(\R^{D} \times \R^{D}\) to \(\R^{D}\), an element of the
    finite-dimensional space \(\cP\) of polynomial maps that are linear in
    \(\vq\) and homogeneous quadratic in \(\vz\), and fix a norm on
    \(\cP\).
    \begin{enumerate}[(i), nosep, leftmargin=2em]
        \item \(P(\vG, \vW) = 0\) if and only if \(\vG = \vzero\) or \(\vW
            = \vzero\), and there is \(\kappa_{0} > 0\) with \(\norm{P(\vG,
            \vW)} \geq \kappa_{0} \norm{\vG}_{F} \norm{\vW}_{F}\) for all
            \(\vG, \vW\).
        \item Let \(\vW\) be invertible and \(\vG \neq \vzero\). Then
            \(P(\vG', \vW') = P(\vG, \vW)\) if and only if \((\vG', \vW')
            = (\mu^{-1} \vG, \mu \vW)\) for some \(\mu \neq 0\), and the
            differential of \(P\) at \((\vG, \vW)\) vanishes on
            \((\dot{\vG}, \dot{\vW})\) if and only if \((\dot{\vG},
            \dot{\vW}) = (\mu \vG, -\mu \vW)\) for some \(\mu \in \R\).
    \end{enumerate}
\end{lemma}

\begin{proof}
    Since \(\vq^{\top} \vW \vz = (\vW \vz)^{\top} \vq\), the map
    \(P(\vG, \vW)\) sends \((\vq, \vz)\) to \((\vG \vz) (\vW
    \vz)^{\top} \vq\), so \(P(\vG, \vW) = P(\vG', \vW')\) if and only
    if \((\vG \vz) (\vW \vz)^{\top} = (\vG' \vz) (\vW' \vz)^{\top}\) for
    all \(\vz \in \R^{D}\).

    \textit{Part (i).} If \(P(\vG, \vW) = 0\), then every \(\vz\) lies in
    \(\ker \vG\) or in \(\ker \vW\), and a vector space is not the union of
    two proper subspaces, so \(\vG = \vzero\) or \(\vW = \vzero\).
    \(P\) is bilinear and continuous and does not vanish on the compact
    set \(\{\norm{\vG}_{F} = \norm{\vW}_{F} = 1\}\), so its minimum
    \(\kappa_{0}\) there is positive, and the bound follows by
    homogeneity.

    \textit{Part (ii).} Let \(U \doteq \R^{D} \setminus \ker \vG\), which
    is open and dense. For \(\vz \in U\), the matrix \((\vG \vz) (\vW
    \vz)^{\top}\) is nonzero of rank one, so \(\vG' \vz = a(\vz) \vG
    \vz\) and \(\vW' \vz = a(\vz)^{-1} \vW \vz\) for a scalar \(a(\vz) \neq
    0\). Hence every \(\vz \in U\) is an eigenvector of \(\vW^{-1} \vW'\).
    A linear map for which every vector of a nonempty open set is an
    eigenvector is a multiple of the identity: for linearly independent
    eigenvectors \(\vz_{1}, \vz_{2} \in U\) and small \(t\), the vector
    \(\vz_{1} + t \vz_{2} \in U\) is an eigenvector only if \(\vz_{1}\) and
    \(\vz_{2}\) have the same eigenvalue, so the eigenvalue is constant on
    \(U\), which spans \(\R^{D}\). Therefore \(\vW' = \mu \vW\) for some
    \(\mu \neq 0\), and then \(\vG' \vz = \mu^{-1} \vG \vz\) on \(U\),
    hence everywhere. The converse is clear. The differential of \(P\) at
    \((\vG, \vW)\) sends \((\dot{\vG}, \dot{\vW})\) to \(\vz \mapsto
    (\dot{\vG} \vz) (\vW \vz)^{\top} + (\vG \vz) (\dot{\vW} \vz)^{\top}\).
    If this vanishes, then for \(\vz \in U\) the two rank-one terms cancel,
    so \(\dot{\vG} \vz = b(\vz) \vG \vz\) and \(\dot{\vW} \vz = -b(\vz) \vW
    \vz\) for a scalar \(b(\vz)\), and the argument above gives \(\dot{\vW}
    = -\mu \vW\) and \(\dot{\vG} = \mu \vG\). The converse is again clear.
\end{proof}

\begin{proof}[Proof of \Cref{thm:nested}]
    \textit{Part (a).} The two constructions represent \(\vf^{\star}\) by
    substitution: in the flat predictor, the terms with \(c_{i} \neq c\)
    have the score zero, and the remaining sum is \(\vm\). For the narrow
    flat predictor, write \(\vG_{ce} \doteq \vW_{ce}^{\mathrm{out}}
    \vW_{ce}^{\mathrm{in}} \vV\) and suppose that it represents
    \(\vf^{\star}\). Then, for every \((c, e)\) and every configuration of
    the labels of the context, the two predictions agree for almost every
    \((\vz_{1}, \dots, \vz_{S})\), hence as polynomials. Setting \(\vz_{i}
    = \vzero\) for \(2 \leq i \leq S - 1\) gives the identity
    \begin{equation}\label{eq:narrow-identity}
        \vG_{ce}\, (\vx_{S}^{\top} \vM \vx_{1})\, \vz_{1} = \indvar\{c_{1}
        = c\}\, \vG_{ce}^{\star} (\vz_{S}^{\top} \vW_{c}^{\star} \vz_{1})\,
        \vz_{1}
        \qquad \text{for all } \vz_{S}, \vz_{1} \in \R^{D}, \ c_{1} \in
        [C],
    \end{equation}
    where \(\vx_{S} = (\ve_{c}, \vz_{S})\) and \(\vx_{1} = (\ve_{c_{1}},
    \vz_{1})\). Write \(\vx_{S}^{\top} \vM \vx_{1} = \alpha + \vbeta^{\top}
    \vz_{1} + \vgamma^{\top} \vz_{S} + \vz_{S}^{\top} \vM_{22} \vz_{1}\),
    where \(\alpha\), \(\vbeta\), and \(\vgamma\) depend on \((c, c_{1})\)
    and \(\vM_{22}\) does not. The right side of
    \eqref{eq:narrow-identity} with \(c_{1} = c\) is nonzero, so
    \(\vG_{ce} \neq \vzero\). Since \(C \geq 2\), there is \(c_{1} \neq
    c\), for which the right side is zero; then the score vanishes for all
    \(\vz_{S}\) and all \(\vz_{1} \notin \ker \vG_{ce}\), hence
    identically, so \(\vM_{22} = \vzero\). For \(c_{1} = c\), the left side
    then has no term of degree one in \(\vz_{S}\) and two in \(\vz_{1}\),
    whereas the right side is \(P(\vG_{ce}^{\star},
    \vW_{c}^{\star})(\vz_{S}, \vz_{1})\) in the notation of
    \Cref{lem:block}, which is of exactly these degrees and nonzero by
    \Cref{lem:block}(i). This contradiction proves the claim.

    \textit{Part (b), the function space.} Let \(\cH\) be the space of
    functions of \((c, e, \vX)\) of the form
    \begin{equation}\label{eq:feature-space}
        \vh(c, e, \vX) = \sum_{i \in [S - 1]} \vp_{c e c_{i}}(\vz_{S},
        \vz_{i}),
        \qquad \vp_{cec'} \in \cP \ \text{ for all } (c, e, c') \in [C]
        \times [E] \times [C],
    \end{equation}
    with \(\cP\) as in \Cref{lem:block}. Both classes lie in \(\cH\): the
    nested predictor has \(\vp_{cec} = P(\vB_{ce} \vA_{c}, \vW_{c})\) and
    \(\vp_{cec'} = 0\) for \(c' \neq c\), and the flat predictor has
    \(\vp_{cec'} = P(\vW_{ce}^{\mathrm{out}} \vW_{ce}^{\mathrm{in}} \vV,
    \vW_{cc'})\). Define \(\norm{\vh}_{\pi}^{2} \doteq \Ex \norm{\vh(c, e,
    \vX)}_{2}^{2}\), so that \(\cE(\vh) = \norm{\vh - \vf^{\star}}_{\pi}^{2}\).
    This is a norm on \(\cH\): if \(\norm{\vh}_{\pi} = 0\), then for every
    \((c, e)\) and every configuration of the labels, which has positive
    probability, the sum in \eqref{eq:feature-space} vanishes for almost
    every \((\vz_{1}, \dots, \vz_{S})\), hence as a polynomial, and
    setting all but one of \(\vz_{1}, \dots, \vz_{S - 1}\) to zero shows
    that every \(\vp_{cec'}
    = 0\). In particular a function in \(\cH\) determines its components
    \(\vp_{cec'}\), and \(\norm{\cdot}_{\pi}\) is finite since the contents
    have finite moments. Let \(\norm{\vh}_{T}^{2} \doteq T^{-1}
    \sum_{i \in [T]} \norm{\vh(c_{i}, e_{i}, \vX_{i})}_{2}^{2}\) be the
    empirical norm. By the law of large numbers applied to a basis of
    \(\cH\), \(\norm{\cdot}_{T} \to \norm{\cdot}_{\pi}\) uniformly on the
    unit sphere of \(\cH\), almost surely, so \(\norm{\cdot}_{T}\) is
    eventually a norm and the unconstrained least squares estimator
    \(\hat{\vh}^{\mathrm{ols}} \in \argmin_{\vh \in \cH} \sum_{i}
    \norm{\vy_{i} - \vh(c_{i}, e_{i}, \vX_{i})}_{2}^{2}\) is eventually
    unique. By the normal equations, for every \(\vh \in \cH\),
    \begin{equation}\label{eq:ls-projection}
        \frac{1}{T} \sum_{i \in [T]} \norm{\vy_{i} - \vh(c_{i}, e_{i},
        \vX_{i})}_{2}^{2}
        = \frac{1}{T} \sum_{i \in [T]} \norm{\vy_{i} -
        \hat{\vh}^{\mathrm{ols}}(c_{i}, e_{i}, \vX_{i})}_{2}^{2}
        + \norm{\vh - \hat{\vh}^{\mathrm{ols}}}_{T}^{2}.
    \end{equation}
    So a least squares estimator over a closed class \(\cM \subseteq
    \cH\) is a projection of \(\hat{\vh}^{\mathrm{ols}}\) onto \(\cM\) in
    the norm \(\norm{\cdot}_{T}\).

    Since \(\vf^{\star} \in \cH\) and \(\veps\) is independent of \((c, e,
    \vX)\) with mean zero and covariance \(\sigma^{2} \vI\), the
    unconstrained problem is a linear regression with the feature map
    given by a basis \(\vh_{1}, \dots, \vh_{p}\) of \(\cH\), and the central
    limit theorem gives \(\sqrt{T} (\hat{\vh}^{\mathrm{ols}} - \vf^{\star})
    \rightsquigarrow \vxi \doteq \sum_{j} \xi_{j} \vh_{j}\), where
    \((\xi_{j})_{j}\) is a
    centered Gaussian vector with covariance \(\sigma^{2} \vGamma^{-1}\) and
    \(\vGamma_{jk} \doteq \Ex[\vh_{j}(c, e, \vX)^{\top} \vh_{k}(c, e,
    \vX)]\) is the Gram matrix of \(\norm{\cdot}_{\pi}\). That is, \(\vxi\)
    is an isotropic Gaussian vector with variance \(\sigma^{2}\) in the
    norm \(\norm{\cdot}_{\pi}\).

    \textit{Part (b), the limit.} Let \(\cM\) be a closed class that
    contains \(\vf^{\star}\) and is, near \(\vf^{\star}\), a smooth
    manifold of dimension \(d\) with tangent space \(\cT\) at
    \(\vf^{\star}\). Its least squares estimator \(\hat{\vf}\) satisfies
    \(\norm{\hat{\vf} - \hat{\vh}^{\mathrm{ols}}}_{T} \leq \norm{\vf^{\star}
    - \hat{\vh}^{\mathrm{ols}}}_{T} = O_{\Pr}(T^{-1/2})\), so \(\hat{\vf} -
    \vf^{\star} = O_{\Pr}(T^{-1/2})\). The objective in
    \eqref{eq:ls-projection} is quadratic, its Hessian converges to that of
    \(\norm{\cdot}_{\pi}^{2}\), and a smooth manifold is Chernoff regular
    with tangent cone \(\cT\). The theorem on constrained estimators of
    \citet[Section 4]{geyer1994asymptotics} then gives \(\sqrt{T}
    (\hat{\vf} - \vf^{\star}) \rightsquigarrow \vP_{\cT} \vxi\), where
    \(\vP_{\cT}\) is the projection onto \(\cT\) that is orthogonal in
    \(\norm{\cdot}_{\pi}\). By the continuous mapping theorem, \(T
        \cE(\hat{\vf}) = \norm{\sqrt{T} (\hat{\vf} - \vf^{\star})}_{\pi}^{2}
    \rightsquigarrow \norm{\vP_{\cT} \vxi}_{\pi}^{2}\), which has the
    distribution \(\sigma^{2} \chi^{2}_{d}\).

    \textit{Part (b), the classes are manifolds.} It remains to show that
    the two classes are smooth manifolds near \(\vf^{\star}\) and to
    compute their dimensions. We parametrize each class by the matrices
    that enter \Cref{lem:block}. The matrices of rank \(r\) in \(\R^{p
    \times q}\) form a smooth manifold of dimension \(r (p + q - r)\), and
    since the rank is lower semicontinuous, the closed set of matrices of
    rank at most \(r\) coincides with this manifold near each of its
    points. For the nested predictor, let \(\bar{\vG}_{c} \in \R^{E D
    \times D}\) stack \(\vG_{c1}, \dots, \vG_{cE}\). A rank factorization
    shows that \((\vG_{ce})_{e} = (\vB_{ce} \vA_{c})_{e}\) for some
    \(\vA_{c}\) and \(\vB_{ce}\) if and only if \(\rank \bar{\vG}_{c} \leq
    r\). The nested class is therefore the image \(\cM_{\mathrm{N}} =
    P_{\mathrm{N}}(\cD_{\mathrm{N}})\) of the polynomial map
    \(P_{\mathrm{N}} \colon (\vG, \vW) \mapsto (P(\vG_{ce}, \vW_{c})
    \indvar\{c' = c\})_{c, e, c'}\) on the domain \(\cD_{\mathrm{N}} \doteq
    \{(\vG, \vW) \colon \rank \bar{\vG}_{c} \leq r \text{ for all } c\}\),
    with \(\vW = (\vW_{c})_{c}\). For the flat predictor,
    \(\vW_{ce}^{\mathrm{out}}
    \vW_{ce}^{\mathrm{in}} \vV\) with \(\vV = \vI\) attains every matrix of
    rank at most \(r\), so the flat class is \(\cM_{\mathrm{F}} =
    P_{\mathrm{F}}(\cD_{\mathrm{F}})\) with \(P_{\mathrm{F}} \colon (\vG,
    \vW) \mapsto (P(\vG_{ce}, \vW_{cc'}))_{c, e, c'}\) on
    \(\cD_{\mathrm{F}} \doteq \{(\vG, \vW) \colon \rank \vG_{ce} \leq r
    \text{ for all } (c, e)\}\), with \(\vW = (\vW_{cc'})_{c, c'}\). The
    targets are \(\vtheta_{\mathrm{N}}^{\star} \doteq (\vG^{\star},
    (\vW_{c}^{\star})_{c})\) and \(\vtheta_{\mathrm{F}}^{\star} \doteq
    (\vG^{\star}, \vW^{\star})\) with \(\vW_{cc}^{\star} = \vW_{c}^{\star}\)
    and \(\vW_{cc'}^{\star} = \vzero\) for \(c' \neq c\). Since
    \(\vB_{ce}^{\star}\) is injective and \(\vA_{c}^{\star}\) is
    surjective, \(\rank \vG_{ce}^{\star} = \rank \bar{\vG}_{c}^{\star} =
    r\), so each domain is a smooth manifold near its target, of dimension
    \begin{equation}\label{eq:domain-dims}
        \dim \cD_{\mathrm{N}} = C r (E D + D - r) + C D^{2},
        \qquad
        \dim \cD_{\mathrm{F}} = C E r (2 D - r) + C^{2} D^{2}.
    \end{equation}
    Both classes are closed, so the least squares estimators exist. Let
    \(P_{s}(\vtheta_{k}) \to \vh\) with \(\vtheta_{k} \in \cD_{s}\), and
    call \(\vW_{c}\), respectively \((\vW_{cc'})_{c'}\), the pattern blocks
    of the label \(c\). If they are all zero, then so are the components of
    \(P_{s}(\vtheta_{k})\) with the query label \(c\), and we may replace
    \(\vG_{ce}\) by \(\vzero\) and the pattern blocks by any blocks of unit
    norm without changing \(P_{s}(\vtheta_{k})\) or leaving \(\cD_{s}\).
    Otherwise, the action of \((\R \setminus \{0\})^{C}\) defined below
    rescales the pattern blocks of the label \(c\) to unit total Frobenius
    norm without changing \(P_{s}(\vtheta_{k})\). After this normalization,
    for each \((c, e)\) some pattern block \(\vW\) of the label \(c\) has
    \(\norm{\vW}_{F} \geq C^{-1/2}\), and \Cref{lem:block}(i) gives
    \(\norm{\vG_{ce}}_{F} \leq C^{1/2} \kappa_{0}^{-1} \norm{P(\vG_{ce},
    \vW)}\), which is bounded because the components of
    \(P_{s}(\vtheta_{k})\) converge. Hence \((\vtheta_{k})_{k}\) is bounded,
    a subsequence converges to some \(\tilde{\vtheta}\) in the closed set
    \(\cD_{s}\), and \(\vh = P_{s}(\tilde{\vtheta}) \in \cM_{s}\) by
    continuity.

    Let \(s \in \{\mathrm{N}, \mathrm{F}\}\). The group \((\R \setminus
    \{0\})^{C}\) acts on \(\cD_{s}\) by \(\vG_{ce} \mapsto \lambda_{c}^{-1}
    \vG_{ce}\) and \(\vW_{c} \mapsto \lambda_{c} \vW_{c}\), respectively
    \(\vW_{cc'} \mapsto \lambda_{c} \vW_{cc'}\) for all \(c'\), and
    \(P_{s}\) is invariant under this action. We claim that near
    \(\vtheta_{s}^{\star}\) the differential of \(P_{s}\) on the tangent
    space of \(\cD_{s}\) has the kernel of dimension exactly \(C\) that is
    tangent to the orbits, and that the fiber \(P_{s}^{-1}(\vf^{\star})
    \cap \cD_{s}\) is the orbit of \(\vtheta_{s}^{\star}\). For the fiber,
    \Cref{lem:block}(ii) applied to the components \((c, e, c)\), where
    \(\vW_{c}^{\star}\) is invertible and \(\vG_{ce}^{\star} \neq
    \vzero\), gives \(\vG_{ce} = \mu_{c e}^{-1} \vG_{ce}^{\star}\) and
    \(\vW_{cc} = \mu_{ce} \vW_{c}^{\star}\), where \(\mu_{ce}\) does not
    depend on \(e\) because the pattern is shared by the experts of the
    label \(c\); and for the flat predictor, \Cref{lem:block}(i) applied to
    the components \((c, e, c')\) with \(c' \neq c\), where \(\vf^{\star}\)
    vanishes and \(\vG_{ce} \neq \vzero\), gives \(\vW_{cc'} = \vzero\).
    For the kernel, the tangent space of the orbit through a point of
    \(\cD_{s}\) consists of the directions \(\dot{\vG}_{ce} = \mu_{c}
    \vG_{ce}\), \(\dot{\vW}_{cc'} = -\mu_{c} \vW_{cc'}\), and it has
    dimension \(C\) wherever the blocks of each label are not all zero,
    which is an open condition that holds at the target; the kernel
    contains it by invariance. At the target, the kernel is no larger:
    \Cref{lem:block}(ii) applied to the components \((c, e, c)\) gives
    \(\dot{\vW}_{cc} = -\mu_{c} \vW_{c}^{\star}\) and \(\dot{\vG}_{ce}
    = \mu_{c} \vG_{ce}^{\star}\) with \(\mu_{c}\) common to the experts of
    the label, and, for the flat predictor, the components \((c, e, c')\)
    with \(c' \neq c\) reduce to \(P(\vG_{ce}^{\star}, \dot{\vW}_{cc'}) =
    0\), so \(\dot{\vW}_{cc'} = \vzero\) by \Cref{lem:block}(i). Since the
    rank of the differential is lower semicontinuous, it is at least
    \(\dim \cD_{s} - C\) near the target, and it is at most this
    wherever the orbit has dimension \(C\). Hence \(P_{s}\) has the
    constant rank \(\dim \cD_{s} - C\) near \(\vtheta_{s}^{\star}\), and
    by the constant rank theorem there is a neighborhood \(U\) of
    \(\vtheta_{s}^{\star}\) in \(\cD_{s}\) such that \(P_{s}(U)\) is an
    embedded submanifold of \(\cH\) of dimension \(d_{s} = \dim \cD_{s} -
    C\).

    We show that \(\cM_{s}\) coincides with \(P_{s}(U)\) near
    \(\vf^{\star}\). Otherwise there are \(\vtheta_{k} \in \cD_{s}\) with
    \(P_{s}(\vtheta_{k}) \to \vf^{\star}\) and \(P_{s}(\vtheta_{k}) \notin
    P_{s}(U)\). The components \((c, e, c)\) of \(P_{s}(\vtheta_{k})\)
    converge to the nonzero components \(P(\vG_{ce}^{\star},
    \vW_{c}^{\star})\) of \(\vf^{\star}\), so by \Cref{lem:block}(i) the
    diagonal blocks of \(\vtheta_{k}\) are eventually nonzero, and by the
    invariance of \(P_{s}\) we may rescale so that they have the Frobenius
    norms of the \(\vW_{c}^{\star}\). The bound in \Cref{lem:block}(i) then
    shows that the \(\vG_{ce}\) of \(\vtheta_{k}\) are bounded and,
    since the components \((c, e, c)\) are bounded away from zero, also
    bounded away from zero, so that all blocks of \(\vtheta_{k}\) are
    bounded. A subsequence converges to some \(\tilde{\vtheta}\) in the
    closed set \(\cD_{s}\) with \(P_{s}(\tilde{\vtheta}) = \vf^{\star}\),
    which lies in the orbit of \(\vtheta_{s}^{\star}\). Applying the
    inverse group element to the subsequence gives points of \(U\) with the
    same images, a contradiction. Therefore \(\cM_{s}\) is a smooth manifold
    of dimension \(d_{s}\) near \(\vf^{\star}\), the limit above applies,
    and \eqref{eq:domain-dims} gives \eqref{eq:dims}. The difference
    \eqref{eq:dims-gap} is positive for \(C \geq 2\), and
    \begin{equation}
        C d_{\mathrm{N}} - d_{\mathrm{F}} = C \big[C r (D - r) + E r ((C -
        2) D + r) - (C - 1)\big] \geq C \big[C - (C - 1)\big] > 0,
    \end{equation}
    since \(1 \leq r < D\) gives \(r (D - r) \geq 1\).

    \textit{Part (c).} With \(T = F / F_{s}\) samples, \(F\,
        \cE(\hat{\vf}^{s}) = F_{s} \cdot T\, \cE(\hat{\vf}^{s})
    \rightsquigarrow \sigma^{2} F_{s} \chi^{2}_{d_{s}}\) by part (b). The
    means \(\sigma^{2} d_{s} F_{s} / F_{s}'\) at the budgets \(F_{s}'\) are
    equal when \(F_{\mathrm{F}}' / F_{\mathrm{N}}' = d_{\mathrm{F}}
    F_{\mathrm{F}} / (d_{\mathrm{N}} F_{\mathrm{N}}) = \kappa(S)\), and
    \(\kappa\) is increasing in \(S\) with the stated limits because \((S_{0}
    + S - 1) / (S_{0} + (S - 1) / C)\) is.
\end{proof}

\subsection{Proof of \Cref{prop:forgetting}}\label{app:theory-forgetting}

\begin{proof}[Proof of \Cref{prop:forgetting}]
    The summary \(\vm\) of \Cref{model:linear} has \(\vSigma_{ce} \succ
    \vzero\): for \(\vv \neq \vzero\), on the event \(\cI = \{1\}\) the
    scalar \(\vv^{\top} \vm = (\vz_{S}^{\top} \vW_{c}^{\star} \vz_{1})
    (\vv^{\top} \vz_{1})\) is a nonzero polynomial, because
    \(\vW_{c}^{\star}\) is invertible, so it is nonzero almost surely.
    The same argument gives \(\vSigma_{e}^{Q} \succ \vzero\).

    The loss \eqref{eq:new-domain-loss} depends on the weights only through
    the predictions \(\vf(c_{Q}, e, \vX; \vtheta)\), \(e \in [E]\). For
    the nested predictor these depend only on \(\vW_{c_{Q}}\),
    \(\vA_{c_{Q}}\), and \(\vB_{c_{Q} e}\), so the gradient with respect
    to \(\vW_{c}\), \(\vA_{c}\), and \(\vB_{ce}\) is zero for \(c \neq
    c_{Q}\), these weights stay at their initial values, and \(\vf(c, e,
    \vX; \vtheta(t)) = \vf^{\star}(c, e, \vX)\). This is part (a).

    For the flat predictor, \(\vf(c_{Q}, e, \vX; \vtheta) = \vW_{c_{Q}
        e}^{\mathrm{out}} \vW_{c_{Q} e}^{\mathrm{in}} \vV \sum_{i \in [S - 1]}
    (\vz_{S}^{\top} \vW_{c_{Q} c_{i}} \vz_{i}) \vz_{i}\) depends only on
    the blocks \(\vW_{c_{Q} c'}\), the experts of the label \(c_{Q}\), and
    \(\vV\). Hence the blocks \(\vW_{cc'}\) and the experts of a label \(c
    \neq c_{Q}\) stay at their initial values \(\vW_{cc} = \vW_{c}^{\star}\),
    \(\vW_{cc'} = \vzero\), \(\vW_{ce}^{\mathrm{in}} = \vA_{c}^{\star}\),
    and \(\vW_{ce}^{\mathrm{out}} = \vB_{ce}^{\star}\), so that \(\vf(c, e,
        \vX; \vtheta(t)) = \vB_{ce}^{\star} \vA_{c}^{\star} \vV(t) \vm =
    \vG_{ce}^{\star} \vV(t) \vm\), and
    \begin{equation}\label{eq:forgetting-flat-form}
        \Phi_{ce}(t) = \frac{1}{2} \Ex\big[\norm{\vG_{ce}^{\star} (\vV(t)
        - \vI) \vm}_{2}^{2} \,\big|\, c, e\big] = \frac{1}{2}
        \norm{\vG_{ce}^{\star} (\vV(t) - \vI) \vSigma_{ce}^{1/2}}_{F}^{2}.
    \end{equation}
    On the new domain the pattern is that of the label \(c_{Q}\), so at
    \(t = 0\) the predictor's summary \(\sum_{i} (\vz_{S}^{\top}
    \vW_{c_{Q} c_{i}} \vz_{i}) \vz_{i}\) equals the summary \(\vm\) of
    \(Q\). Since \(\veps\) has mean zero and is independent of \(\vX\), for
    any matrix \(\vG\) we have \(\Ex_{Q}[(\vG \vm - \vy) \vm^{\top} \mid e]
    = (\vG - \vG'_{e}) \vSigma_{e}^{Q}\), and therefore
    \begin{equation}
        \vnabla_{\vV} L_{Q}(\vtheta(0)) = \sum_{e \in [E]} \pi_{e}^{Q}
        \big(\vW_{c_{Q} e}^{\mathrm{out}} \vW_{c_{Q}
        e}^{\mathrm{in}}\big)^{\top} (\vG_{c_{Q} e}^{\star} - \vG'_{e})
        \vSigma_{e}^{Q} = \vA_{c_{Q}}^{\star\top} \vXi,
    \end{equation}
    using \(\vW_{c_{Q} e}^{\mathrm{out}} \vW_{c_{Q} e}^{\mathrm{in}} =
    \vB_{c_{Q} e}^{\star} \vA_{c_{Q}}^{\star}\) at \(t = 0\). Since
    \(\vA_{c_{Q}}^{\star\top}\) is injective, this gradient is zero if and
    only if \(\vXi = \vzero\). The vector field of the gradient flow is
    polynomial in \(\vtheta\), so \(\vtheta(t)\) is smooth and \(\vV(t) =
    \vI - t \vA_{c_{Q}}^{\star\top} \vXi + O(t^{2})\). Substituting into
    \eqref{eq:forgetting-flat-form} gives \eqref{eq:forgetting-rate}.
    As \(\vB_{ce}^{\star}\) is injective and \(\vSigma_{ce}^{1/2}\)
    invertible, the leading coefficient is zero if and only if
    \(\vA_{c}^{\star} \vA_{c_{Q}}^{\star\top} \vXi = \vzero\); this holds
    when \(\vA_{c}^{\star} \vA_{c_{Q}}^{\star\top} = \vzero\), and when
    \(\vA_{c}^{\star} \vA_{c_{Q}}^{\star\top}\) is invertible it
    holds if and only if \(\vXi = \vzero\).
\end{proof}

\subsection{Further Comparisons}\label{app:theory-comparisons}

We give two remarks on the scope of \Cref{thm:nested}: the first is on
other shapes of the flat predictor, and the second relates the product
features of the flat predictor to the full architectures.

\begin{remark}{}{flat-configurations}
    The comparison in \Cref{thm:nested} does not depend on the shape of the
    experts of the flat model. If its experts have rank \(k > r\), its
    class contains \(\cM_{\mathrm{F}}\), so its tangent cone \(\cK\) at
    \(\vf^{\star}\) contains the tangent space \(\cT\) of
    \(\cM_{\mathrm{F}}\). The class is the image of a polynomial map and
    therefore Chernoff regular \citep[Lemma 3.3]{drton2009likelihood}, the
    limit of \(T \cE\) is the squared norm of a projection of \(\vxi\) onto
    \(\cK\), and for a closed cone this is \(\norm{\vxi}_{\pi}^{2} -
        \operatorname{dist}(\vxi, \cK)^{2} \geq \norm{\vP_{\cT}
    \vxi}_{\pi}^{2}\). Wider experts therefore do not reduce the limit. If
    \(C r < D\) and the row spaces of the matrices \(\vA_{c}^{\star}\) are
    linearly independent, the flat model can also represent the target
    with a shared value map \(\vV \in \R^{C r \times D}\) that stacks the
    matrices \(\vA_{c}^{\star}\) and experts \(\vW_{ce} \in \R^{D \times C
    r}\) of rank \(r\), i.e., linear MLPs with \(r\) hidden units. The map
    \((\vV, (\vW_{ce})_{c, e}) \mapsto (\vW_{ce} \vV)_{c, e}\) is invariant
    to \(\vV \mapsto \vX \vV\), \(\vW_{ce} \mapsto \vW_{ce} \vX^{-1}\) for
    invertible \(\vX\), and the kernel of its differential at the target
    consists of these directions only, because \(\bigcap_{c, e} \ker
    \vW_{ce} = \{\vzero\}\) there. The map terms of the dimension of this
    class at the target are therefore \(C r D + C E r (C r + D - r) - (C
    r)^{2} = C r (D - C r) + C E r (C r + D - r)\), which exceed the map
    terms \(C r (E D + D - r)\) of \(d_{\mathrm{N}}\) by \(C (C - 1) (E -
    1) r^{2}\), while the pattern terms are those of \(d_{\mathrm{F}}\).
    The extra terms count the expert parameters needed to pick out
    the \(r\) coordinates of a column from the \(C r\) outputs of \(\vV\).
\end{remark}

\begin{remark}{}{simulation}
    An MoE transformer can simulate a column of \ours{} with a residual
    stream that is wider by a constant factor. Consider one head of
    column \(c\),
    with logits \(\ip{\vQ_{c} \vx_{j}}{\vK_{c} \vx_{i}}\) and values
    \(\vV_{c} \vx_{i}\) for \(i \in \cU_{c}\). In an MoE transformer, let the
    experts of the preceding layer be routed by the column labels, and let
    the expert for the label \(c\) output \(\vQ_{c} \vx_{i}\), \(\vK_{c}
    \vx_{i}\), and \(\vV_{c} \vx_{i}\) in separate coordinates of the
    residual stream. A shared head whose queries, keys, and values are
    these coordinates has the same logits
    and values for all pairs of tokens in the same column, and the following
    experts can apply the output map of the column. The pairs in
    different columns are removed by adding \(\lambda (\indvar\{c_{i} = c_{j}\}
    - 1)\) to the logits, which needs \(C\) more coordinates and is exact as
    \(\lambda \to \infty\). This simulation has the same number of
    stored parameters as the columns. It needs a residual stream that
    is wider by
    the total width of the queries, keys, and values of the active columns.
    This width is a constant multiple of \(D\), and it increases the size of
    every other matrix. We
    therefore expect the two architectures to differ by constants at
    equal resources. The product features of the flat predictor in
    \eqref{eq:nested-predictor} are this construction for one layer, and
    \Cref{thm:nested} gives the resulting constant; \Cref{prop:scaling}
    in \Cref{app:scaling-laws} shows that a constant in the coefficients
    is a constant compute multiplier.
\end{remark}

\end{document}